\documentclass[10pt,twocolumn,letterpaper]{article}

\usepackage{iccv}
\usepackage{times}
\usepackage[utf8]{inputenc}
\usepackage{amsmath}
\usepackage{amssymb}
\usepackage{graphicx}
\usepackage[font=small]{caption}
\usepackage{tabularx}
\usepackage{booktabs}
\usepackage{multirow}
\usepackage{subfig}
\usepackage{setspace}
\usepackage{color}			
\usepackage{microtype}
\usepackage{pifont}
\newcommand{\xmark}{\ding{55}}%
\usepackage{pgfplots}
\usepackage{tikz}
\usepackage{pstricks}
\usetikzlibrary{decorations.pathreplacing}
\usetikzlibrary{calc,3d}
\usetikzlibrary{positioning,shapes,decorations,shadows,arrows,plotmarks,fit}
\usetikzlibrary{spy}
\newlength \figureheight 
\newlength \figurewidth

\renewcommand{\vec}[1]{\boldsymbol{#1}}
\newcommand{\fref}[1]{Fig.\,\ref{#1}}
\newcommand{\Fref}[1]{Figure\,\ref{#1}}
\newcommand{\tref}[1]{Tab.\,\ref{#1}}
\newcommand{\Tref}[1]{Table\,\ref{#1}}
\newcommand{\sref}[1]{Sect.\,\ref{#1}}

\usepackage[pagebackref=true,breaklinks=true,letterpaper=true,colorlinks,bookmarks=false]{hyperref}

\iccvfinalcopy % *** Uncomment this line for the final submission

\def\iccvPaperID{2403} % *** Enter the ICCV Paper ID here

\ificcvfinal\fi
\begin{document}

%%%%%%%%% TITLE
\title{Benchmarking Super-Resolution Algorithms on Real Data}

\author{Thomas K\"ohler$^1$, Michel B\"atz$^2$, Farzad Naderi$^1$, Andr\'{e} Kaup$^2$, Andreas K. Maier$^1$, and Christian Riess$^{1,3}$
\vspace{0.2em}  \and $^1$ Pattern Recognition Lab\\
Dept. of Computer Science
\and $^2$ Multimedia Communications and Signal Processing\\
Dept. of Electrical, Electronic and Communication Engineering\\
\and $^3$ IT Infrastructures Lab\\
Dept. of Computer Science
\and Friedrich-Alexander-Universit\"at (FAU) Erlangen-N\"urnberg, Erlangen, Germany\\
\url{http://www.superresolution.tf.fau.de/}
}
\maketitle

%%%%%%%%% ABSTRACT
\begin{abstract}
Over the past decades, various super-resolution (SR) techniques have been developed to enhance the spatial resolution of digital images. Despite the great number of methodical contributions, there is still a lack of comparative validations of SR under practical conditions, as capturing real ground truth data is a challenging task. Therefore, current studies are either evaluated 1) on simulated data or 2) on real data without a pixel-wise ground truth.

To facilitate comprehensive studies, this paper introduces the publicly available Super-Resolution Erlangen (SupER) database that includes real low-resolution images along with high-resolution ground truth data. Our database comprises image sequences with more than 20k images captured from 14 scenes under various types of motions and photometric conditions. The datasets cover four spatial resolution levels using camera hardware binning. With this database, we benchmark 15 single-image and multi-frame SR algorithms. Our experiments quantitatively analyze SR accuracy and robustness under realistic conditions including independent object and camera motion or photometric variations.
\end{abstract}

%%%%%%%%% BODY TEXT
\section{Introduction}
\label{sec:Introduction}

Super-resolution (SR) \cite{Milanfar2010} is a class of algorithms that aims at enhancing the spatial resolution of digital images. SR features retrospective resolution enhancement -- without modifying the detectors or optics. This facilitates cost-effective high-resolution imagery to break limitations dictated by the sampling theorem and holds the potential to improve various vision tasks \cite{Dai2016}, including surveillance \cite{Zhang2010a}, remote sensing \cite{Zhang2012b}, 3-D imaging \cite{Schuon2009}, and healthcare \cite{Kohler2014,Kohler2015a}. SR is an ill-posed problem and applicable to single or multiple images. Single-image SR (SISR) exploits the information within a single low-resolution (LR) image to infer high-resolution (HR) details. Multi-frame SR (MFSR) uses multiple LR frames with relative motion or blur among them to reconstruct HR images \cite{Elad1997}. 

SR is a well-researched problem and several seminal works on inherent limitations appeared. This includes algebraic or numerical studies regarding the maximum resolution gain \cite{Baker2002,Lin2004} and statistical performance studies \cite{Robinson2006}. Despite these insights and the wide deployment of existing algorithms, there is still a lack of comparative validations of SR under practical conditions. On the one hand, recent theoretical studies are based on certain approximations and simplifications, \eg simplified motion models or linearity and shift invariance of the imaging system. Hence, they can only roughly predict upper or lower performance bounds. On the other hand, experimental studies have only partially addressed practical constraints such as real non-Gaussian noise, low-light exposures, or photometric variations. This can be attributed to the fact that prior work either employs simulated data generated under somewhat simplifying environmental conditions, or real acquisitions without ground truth.
Compared to other areas in computer vision, \eg motion analysis \cite{Geiger2012} or deblurring \cite{Lai2016}, there is still a lack of quantitative and real SR benchmarks. This limits the significance and reproducibility of experimental studies.

\begin{figure*}[!t]
	\centering
		\includegraphics[width=1.00\textwidth]{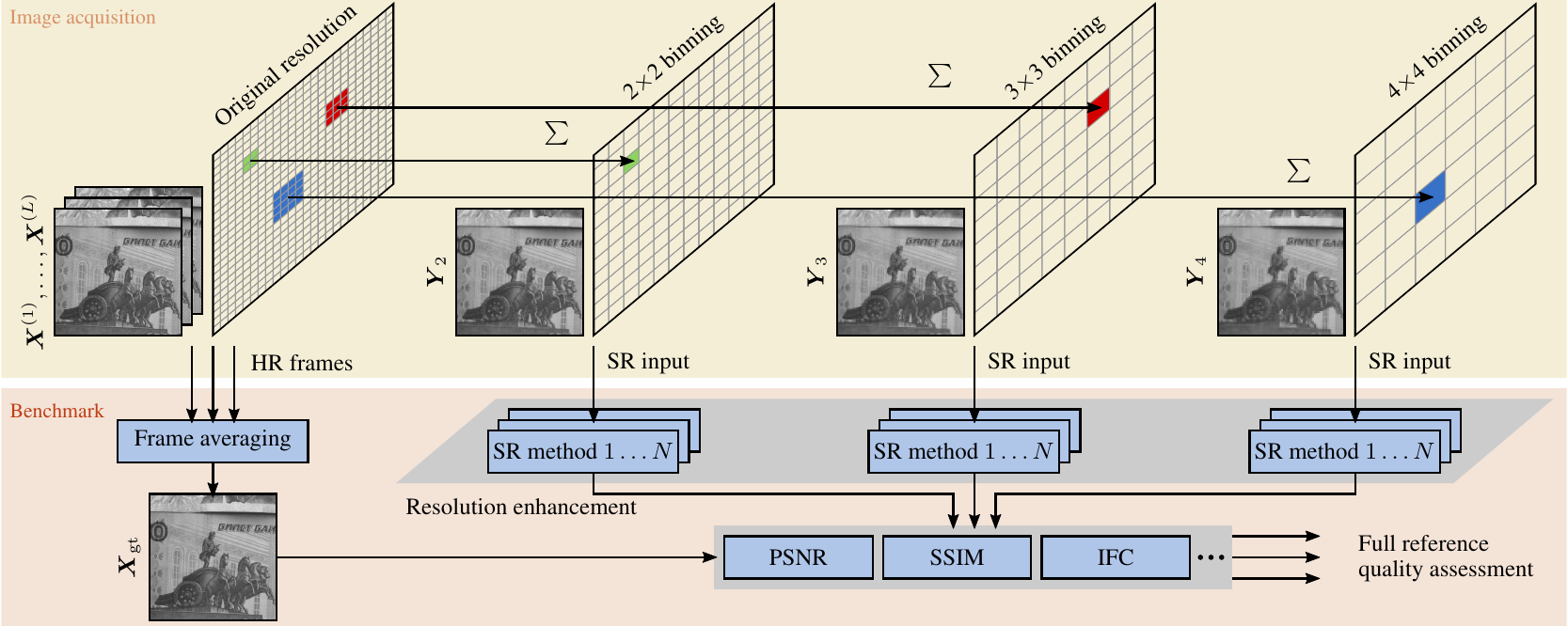}
	\caption{Overview of the proposed image acquisition and benchmark setup. In our image acquisition scheme, we capture multiple frames at the actual pixel resolution of our camera to obtain a ground truth high-resolution image via frame averaging. In contrast to prior work, we use hardware binning on the sensor to gain real low-resolution images without software-based simulations (see \sref{sec:SuperResolutionBenchmarkDataset}). In the benchmark, we employ full-reference quality measures to assess the fidelity of super-resolved data relative to the ground truth (see Sects.~\ref{sec:BenchmarkSetup} and \ref{sec:ExperimentsAndResults}).}
	\label{fig:flowchart}
\end{figure*}

This paper complements theoretical works \cite{Baker2002,Lin2004,Robinson2006}
by a comparative benchmark of SR methods. To this end, we set up a novel image
database to enable quantitative evaluations, see \fref{fig:flowchart}. Our
contributions are two-fold: 1) We collected a large database termed \textit{Super-Resolution Erlangen} (SupER) database
including LR images at \textit{multiple levels of spatial resolution} and ground
truth HR data, see \fref{fig:flowchart} (top). In contrast to prior work, our LR data are not simulated.
They consist of real acquisitions that are obtained via
\textit{hardware binning}, and covers difficult real-world conditions like local
object motion or
photometric variations. The database comprises more than 20k images of 14
scenes at 4 resolution levels. 2) We present a comprehensive benchmark of
state-of-the-art SR algorithms. Our study is based on four full-reference
quality measures that uses our ground truth data to quantitatively assess
SR, see \fref{fig:flowchart} (bottom). In total, we validated six SISR and
nine MFSR algorithms.

The proposed database and benchmark can serve as common base in the community
to understand and evaluate SR techniques. We provide all data and all source
code implementing the evaluation protocols on our webpage, to foster
quantitative SR evaluation on real images.

\section{Related Work}
\label{sec:RelatedWork}

\noindent
\textbf{Super-resolution algorithms.}
Current SISR algorithms either use the information within a single image or
external data. \textit{Internal} methods exploit prior knowledge on HR images,
\eg edge statistics \cite{Fattal2007} or self-similarities
\cite{Glasner2009,Huang2015a}. \textit{External} methods use training data to
learn mappings from source LR images to target HR images. Recent
approaches include sparse coding of image patches for dictionary learning
\cite{Yang2010}, example-based kernel ridge regression \cite{Kim2010}, tree-based 
methods \cite{Salvador2015}, or random forest regression \cite{Schulter2015}. Another 
approach is to infer end-to-end mappings via deep learning, \eg using CNNs 
\cite{Dong2014,Kim2016}.

Current MFSR algorithms can be divided into three classes.
\textit{Interpolation} schemes fuse multiple LR frames into a HR image by
motion compensation followed by non-uniform interpolation using
kernel regression~\cite{Takeda2007}, Voronoi tessellation~\cite{Batz2016}, or
hybrid example-based interpolation \cite{Batz2015}.
\textit{Reconstruction} methods are based on iterative energy
minimization. This includes maximum a-posteriori (MAP)
\cite{Bercea2016,Elad1997,Farsiu2004a,Kohler2015c} and variational Bayesian
inference \cite{Babacan2011} using statistical priors to alleviate the
ill-posedness of SR. Several algorithms estimate HR images along with optical
blur~\cite{Liu2014} or incorporate motion deblurring \cite{Ma2015,Zhang2014a}.
\textit{Deep learning} methods learn end-to-end mappings between
motion-compensated LR frames and the HR
image \cite{Dirks2016,Kappeler2016a,Kappeler2016,Liao2015}.
\\[0.8ex]
\textbf{Datasets and evaluation strategies.}
Compared to the great number of algorithmic contributions, there is only few
prior work on their comparative evaluations.

Yang \etal \cite{Yang2014a} and Timofte \etal \cite{Timofte2016} have reported benchmarks and improvements of various SISR techniques. However, besides visual inspection, their quantitative evaluations are entirely based on simulated images. This facilitates comparisons to a ground truth by full-reference quality measures but limits the significance to evaluate SR under realistic constraints. For instance, the study in \cite{Yang2014a} considered simplified artificial noise, \eg Gaussian noise, and does not cover challenging environmental conditions, such as low-light exposures or photometric variations. Liu and Sun \cite{Liu2014} benchmarked MFSR on video datasets but LR images are obtained by artificial sampling and noise. Other studies \cite{Dai2016,Raghavendra2013} validated SR for specific vision tasks under more realistic constraints but have limited informative value for general benchmarks on natural images. Our work aims at broadly benchmarking SR on real captured images. 

Existing real-world image databases \cite{Farsiu2014,Vandewalle2016} are designed for evaluations by visual inspection due to the lack of ground truth data. Thus, quantitative evaluations need to use no-reference quality measures as for example done in \cite{Yeganeh2012,Yuan2012}. However, finding appropriate no-reference measures to assess SR on general scenes is a controversial issue. Another strategy are large-scale human subject studies to assess image quality as previously conducted for deblurring \cite{Lai2016} or SISR \cite{Yang2014a}. This ensures high agreement to human visual perception but is cumbersome and difficult to reproduce. Our work aims at constructing a database of real LR images with ground truth HR data for quantitative studies.

Qu \etal \cite{Qu2016} have constructed a database of real face images with ground truth data. Their setup utilizes two cameras that are combined with a beam splitter to simultaneously capture LR and HR images. However, the alignment of LR and HR data is potentially affected by error-prone system calibrations and image registrations. This makes the use of full-reference quality measures for pixel-wise comparisons between super-resolved and ground truth data less reliable. Furthermore, their database covers only single images, which precludes studies of MFSR algorithms. We propose a single-camera setup that avoids these limitations and acquires aligned images at multiple resolution levels.

\begin{figure}[!t]
	\centering
	\includegraphics[width=0.19\linewidth]{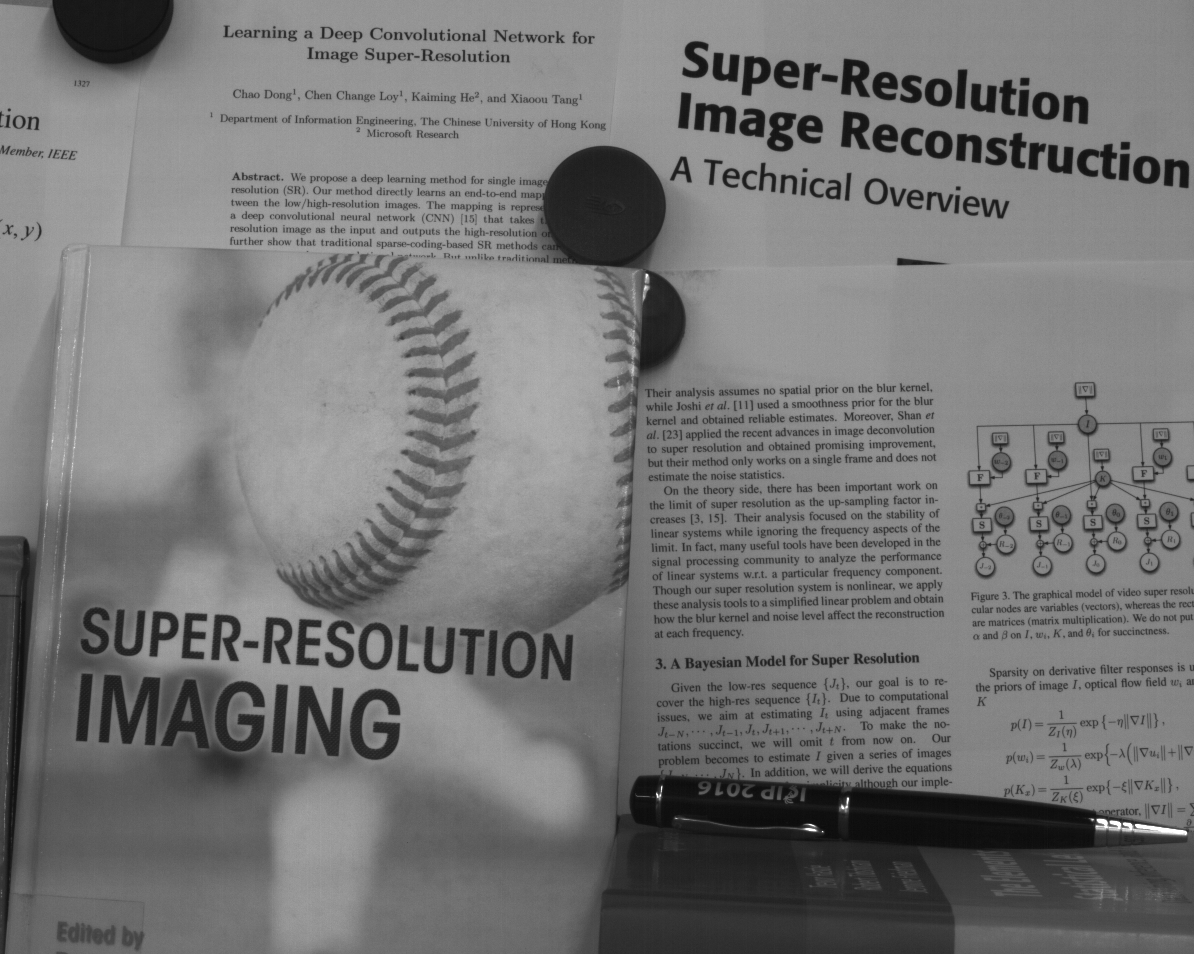}
	\includegraphics[width=0.19\linewidth]{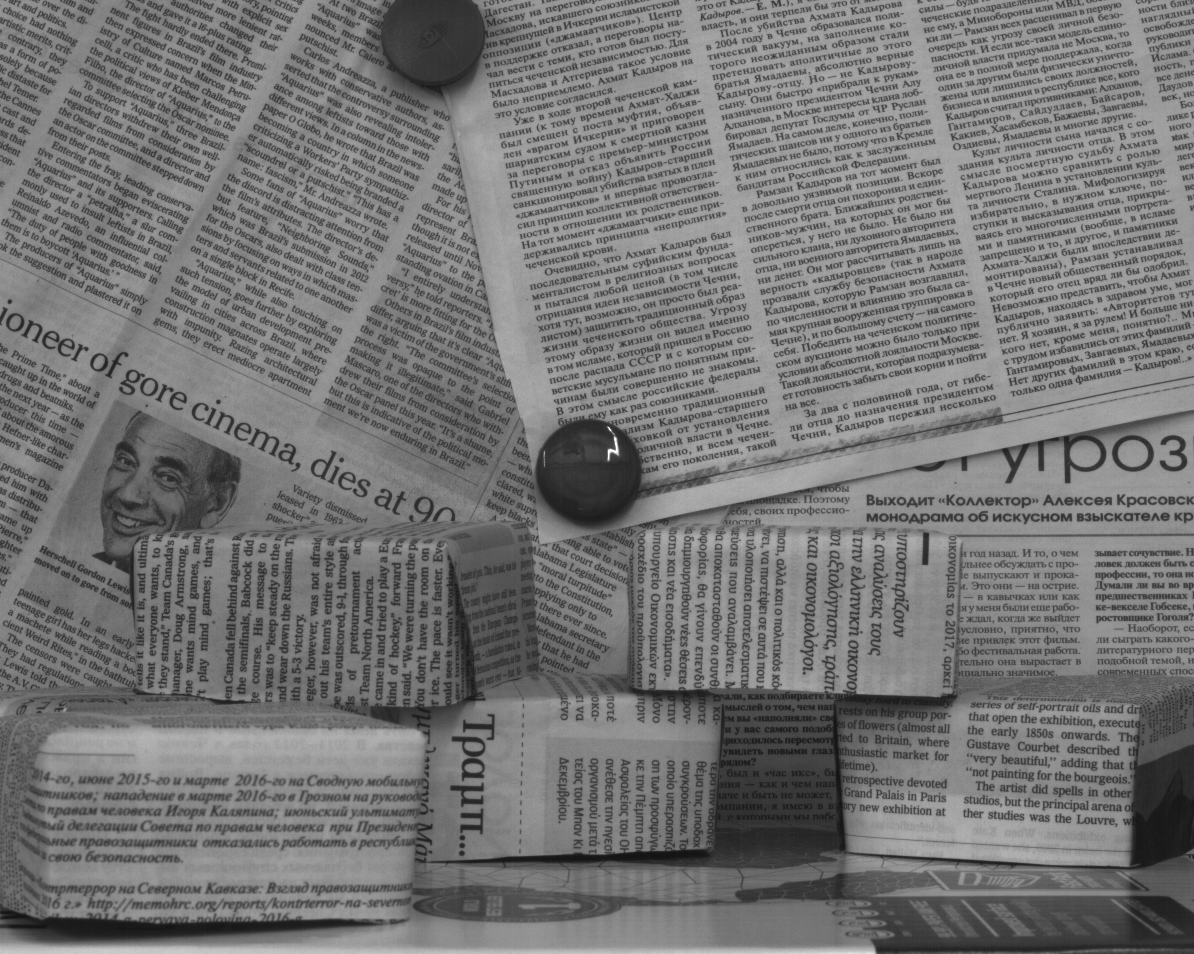}
	\includegraphics[width=0.19\linewidth]{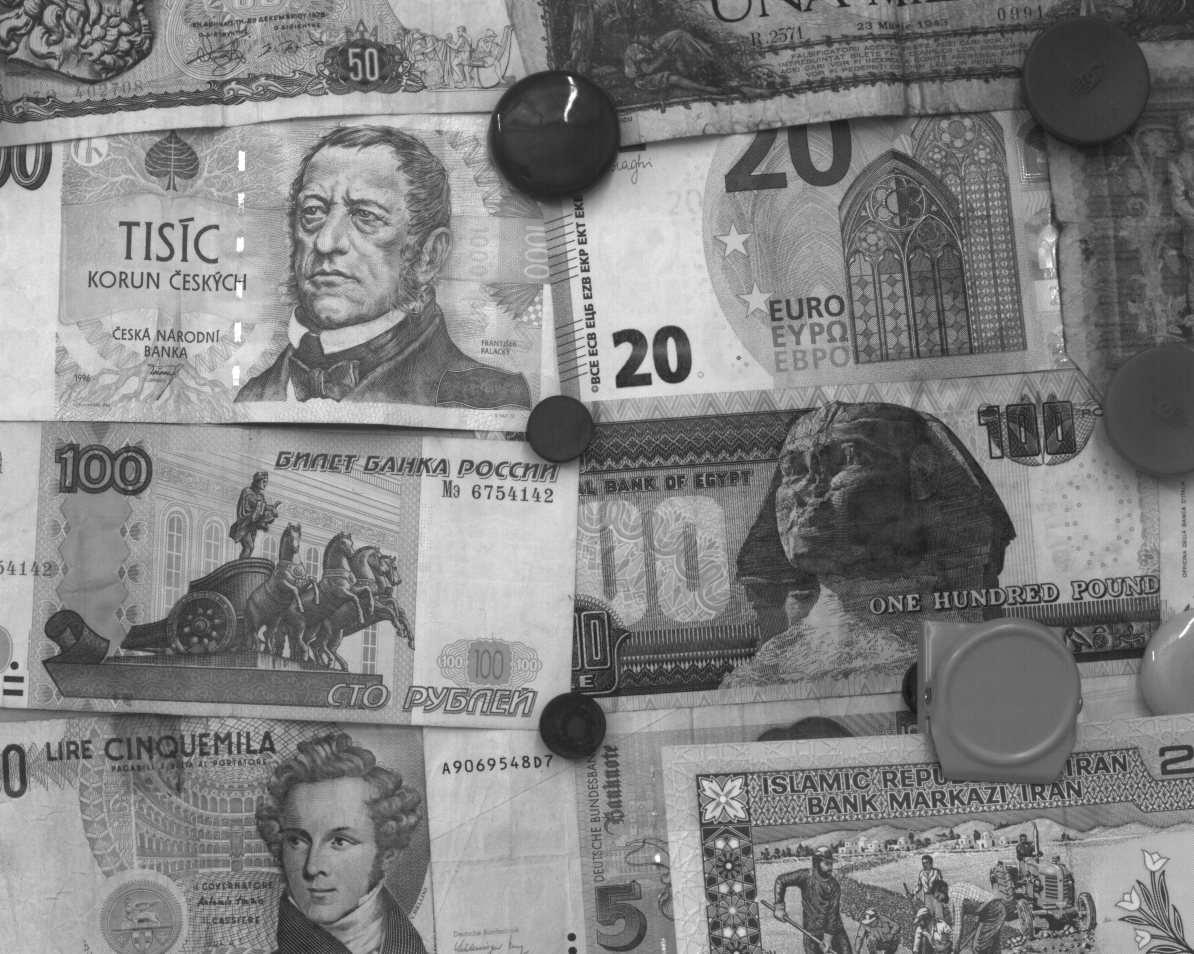}
	\includegraphics[width=0.19\linewidth]{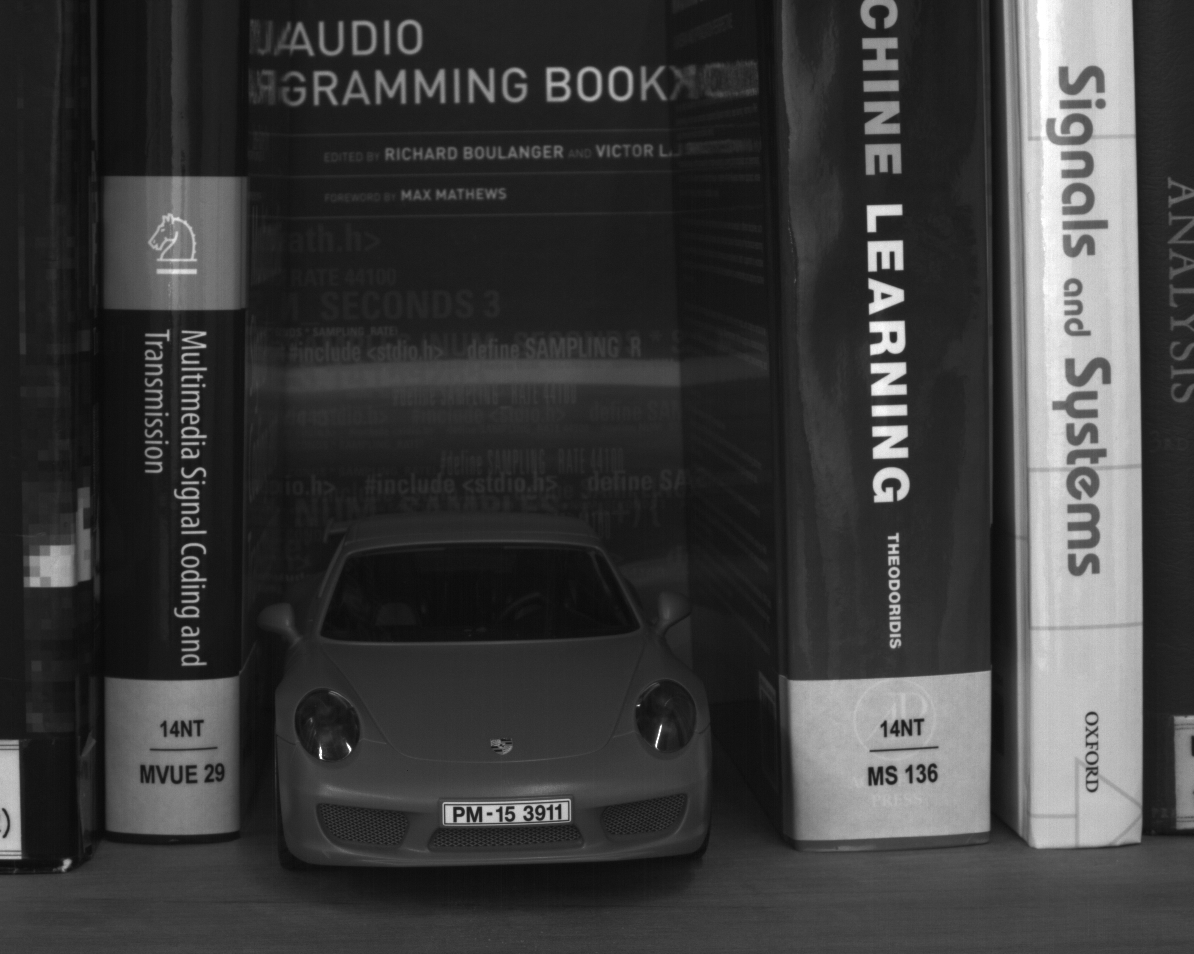}
	\includegraphics[width=0.19\linewidth]{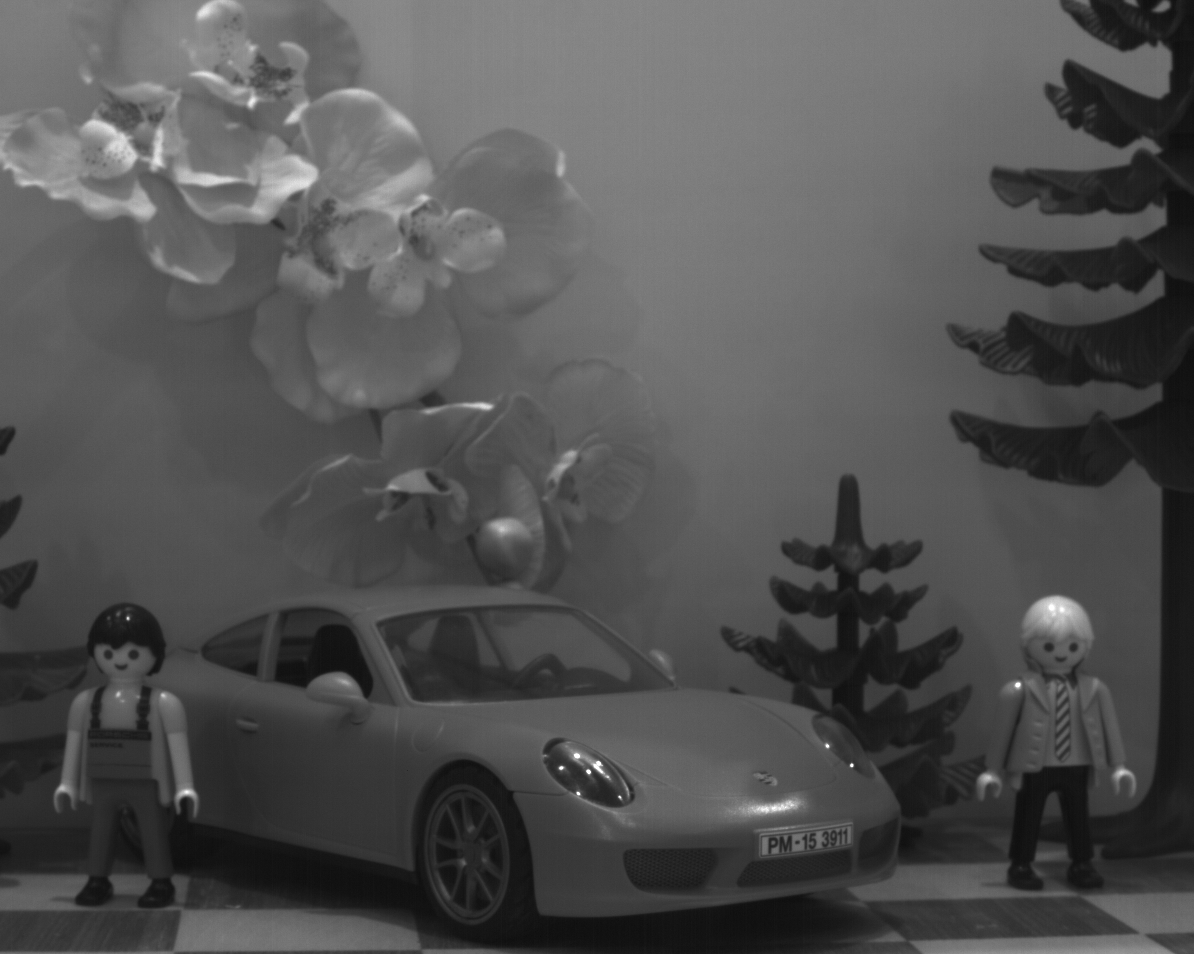}\\[0.2ex] 
	\includegraphics[width=0.19\linewidth]{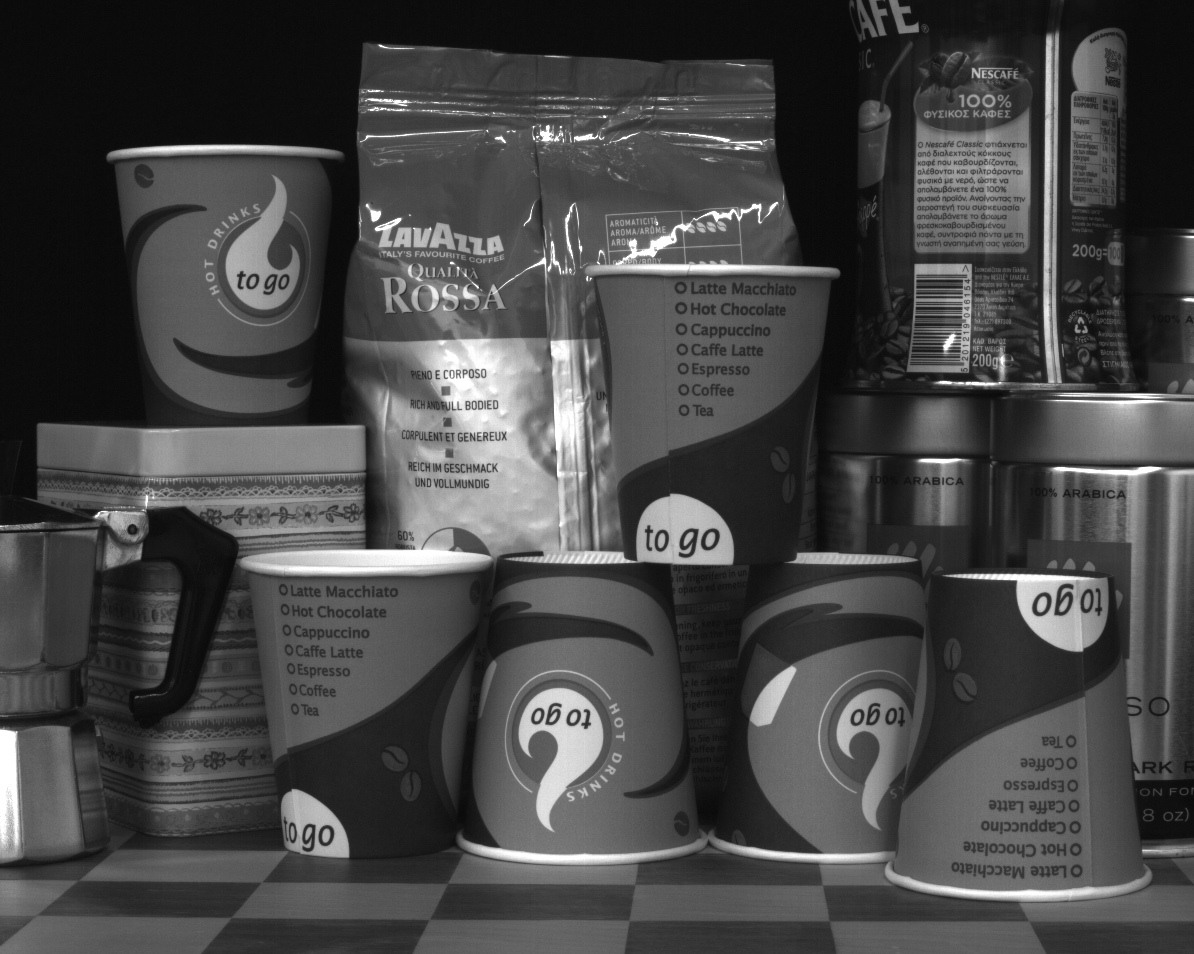}
	\includegraphics[width=0.19\linewidth]{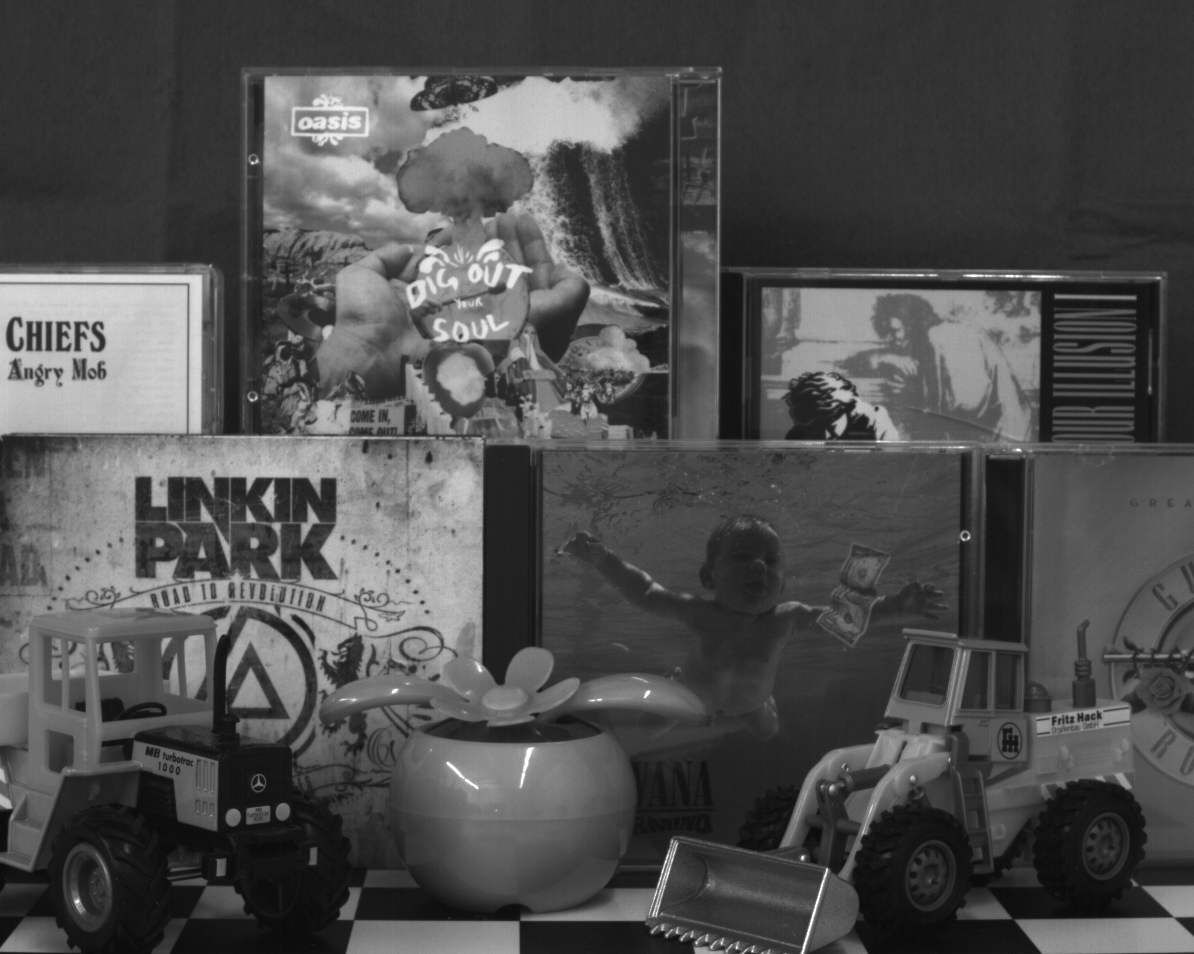}
	\includegraphics[width=0.19\linewidth]{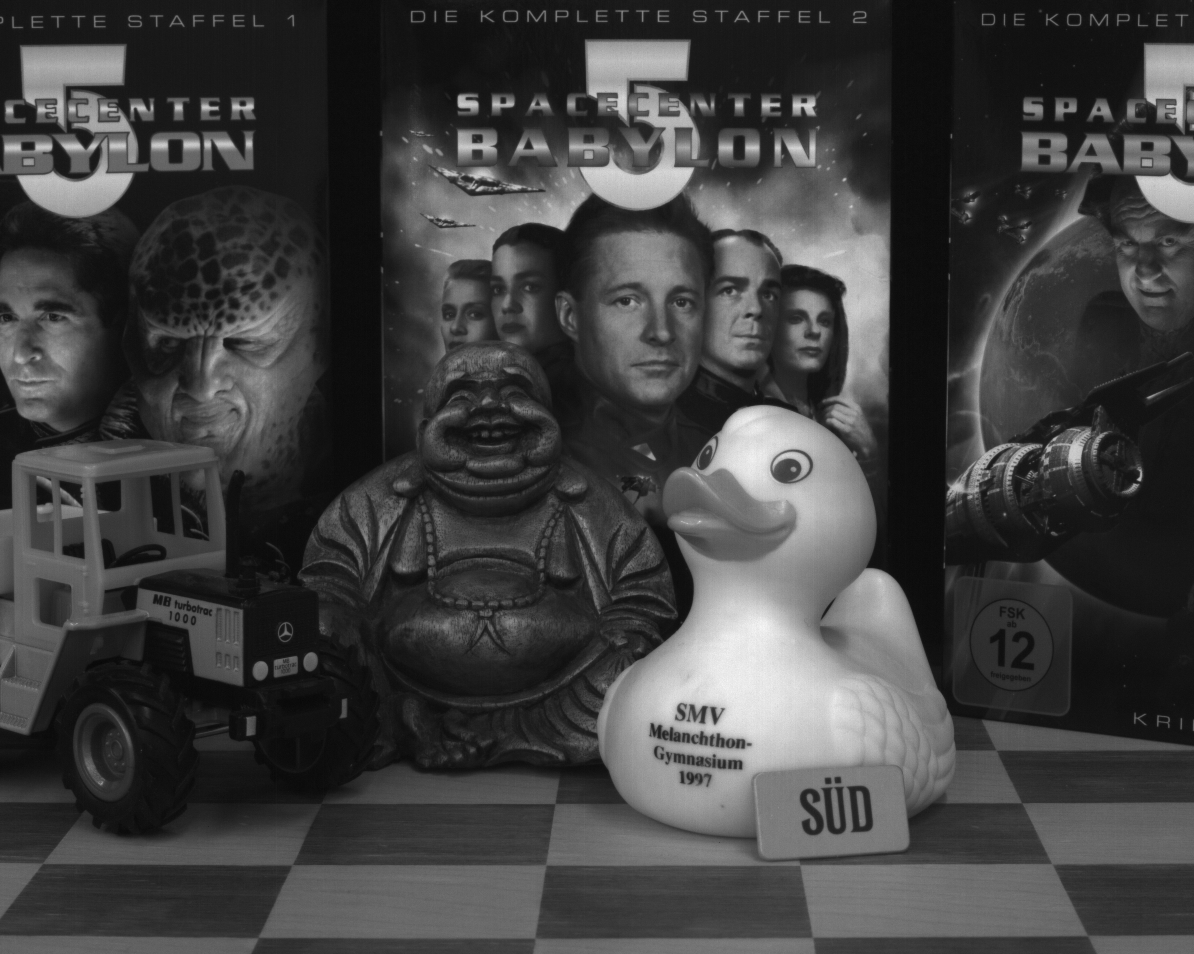}
	\includegraphics[width=0.19\linewidth]{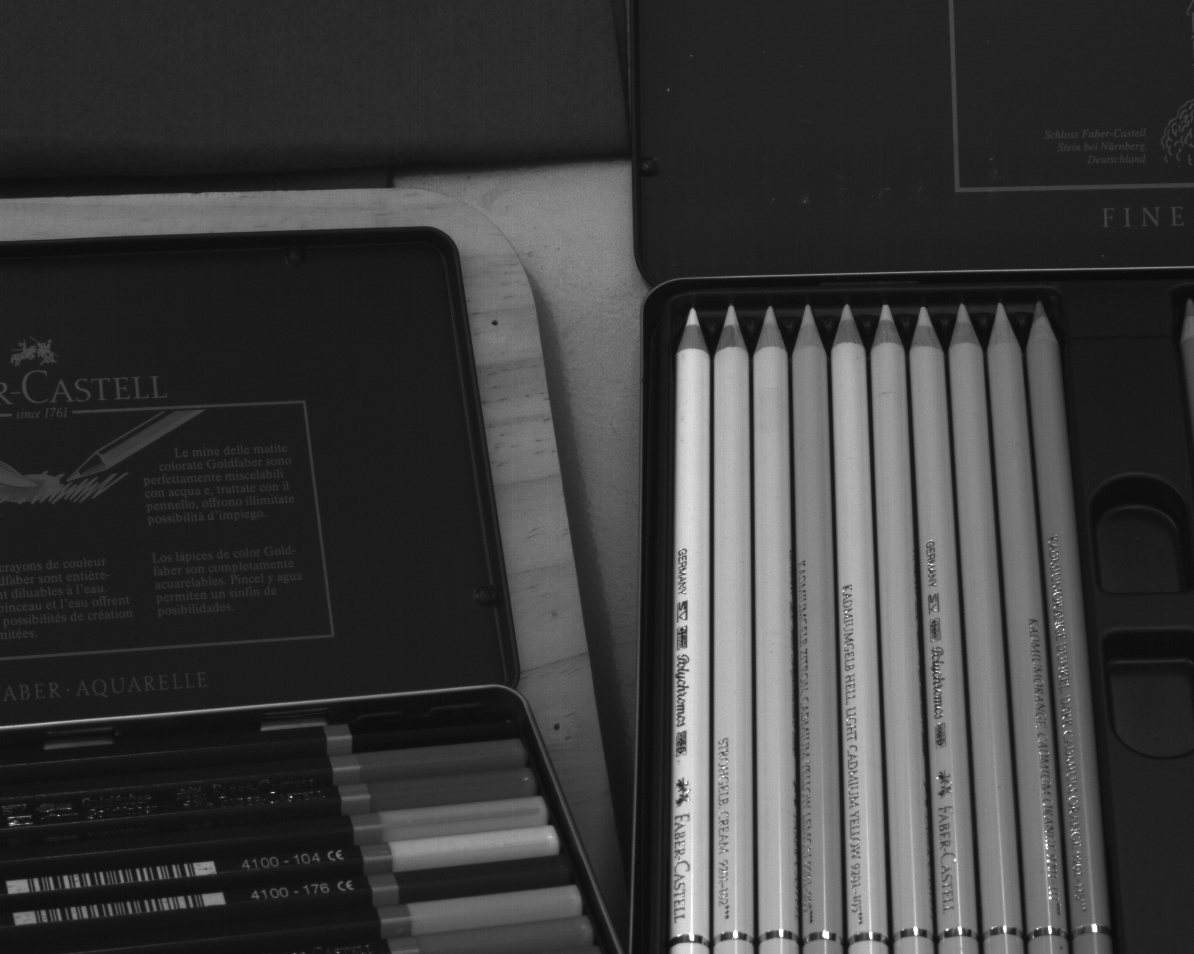}
	\includegraphics[width=0.19\linewidth]{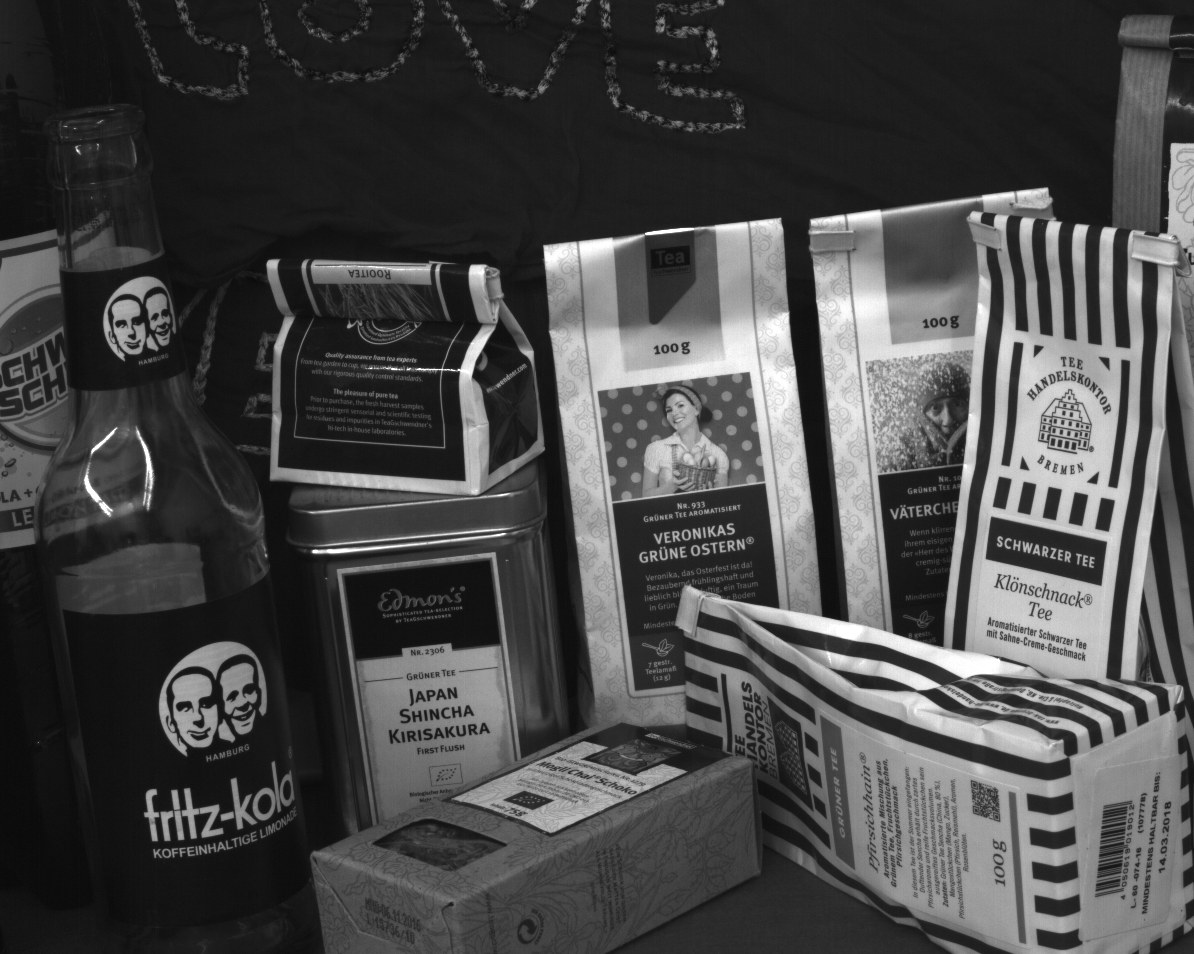}\\[0.2ex]
	\includegraphics[width=0.19\linewidth]{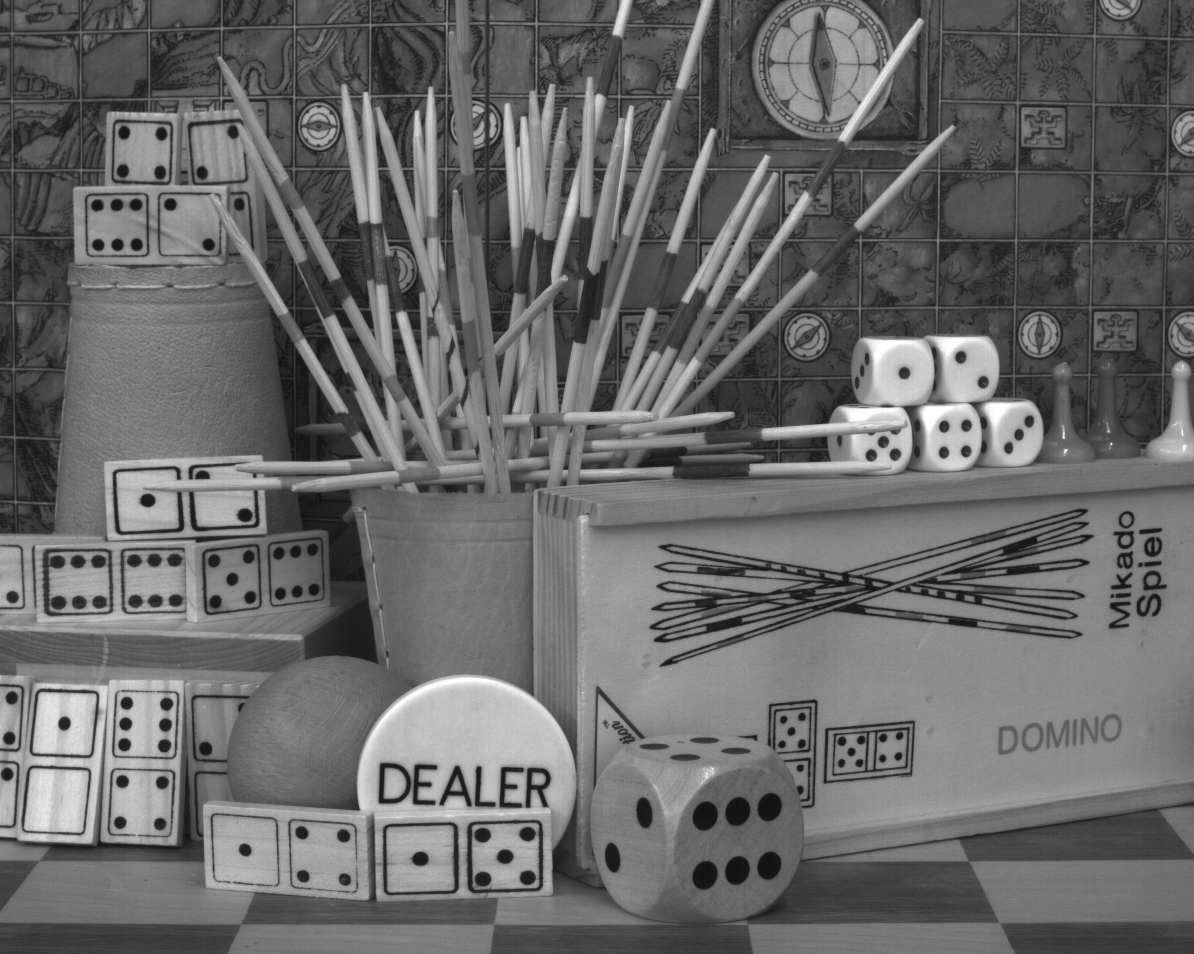}
	\includegraphics[width=0.19\linewidth]{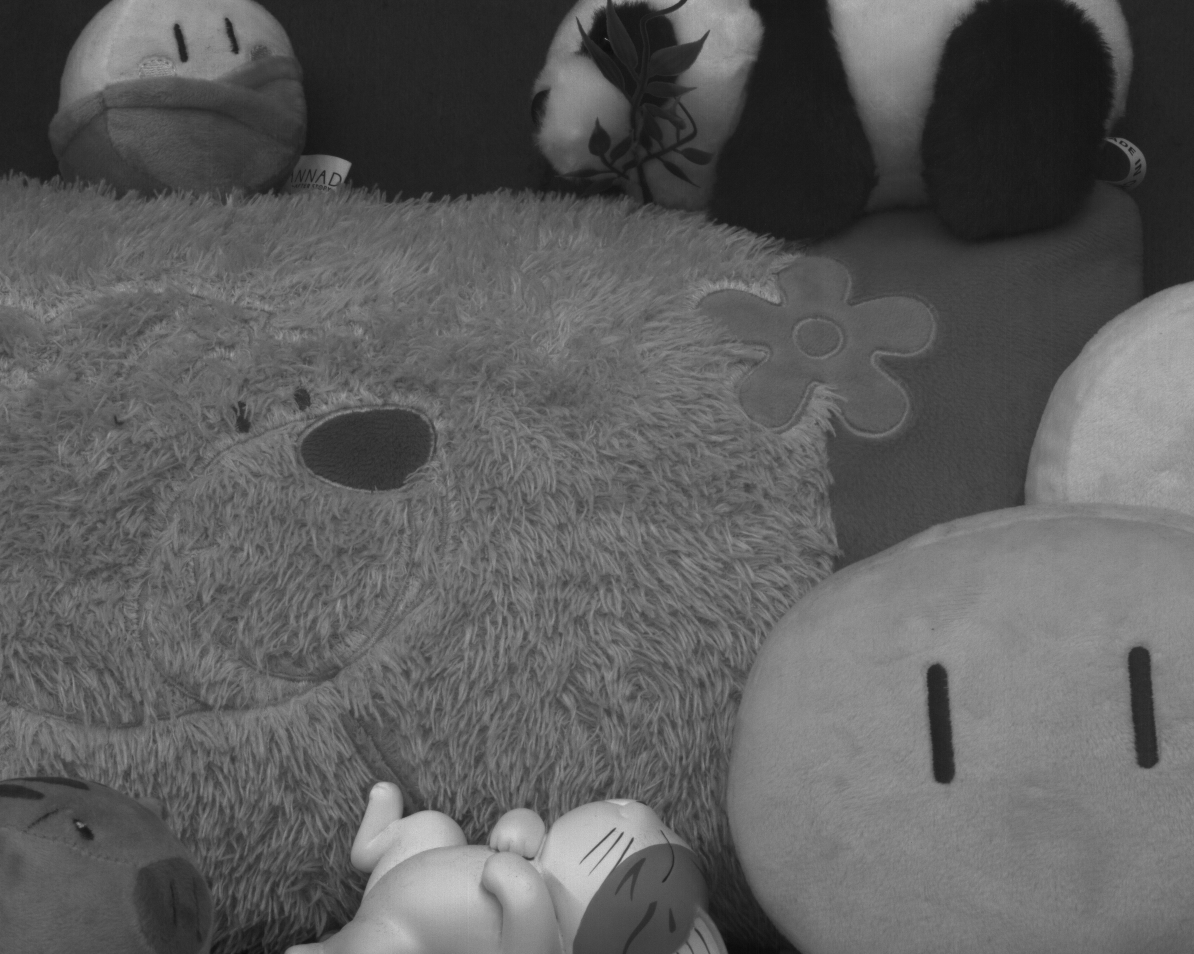}
	\includegraphics[width=0.19\linewidth]{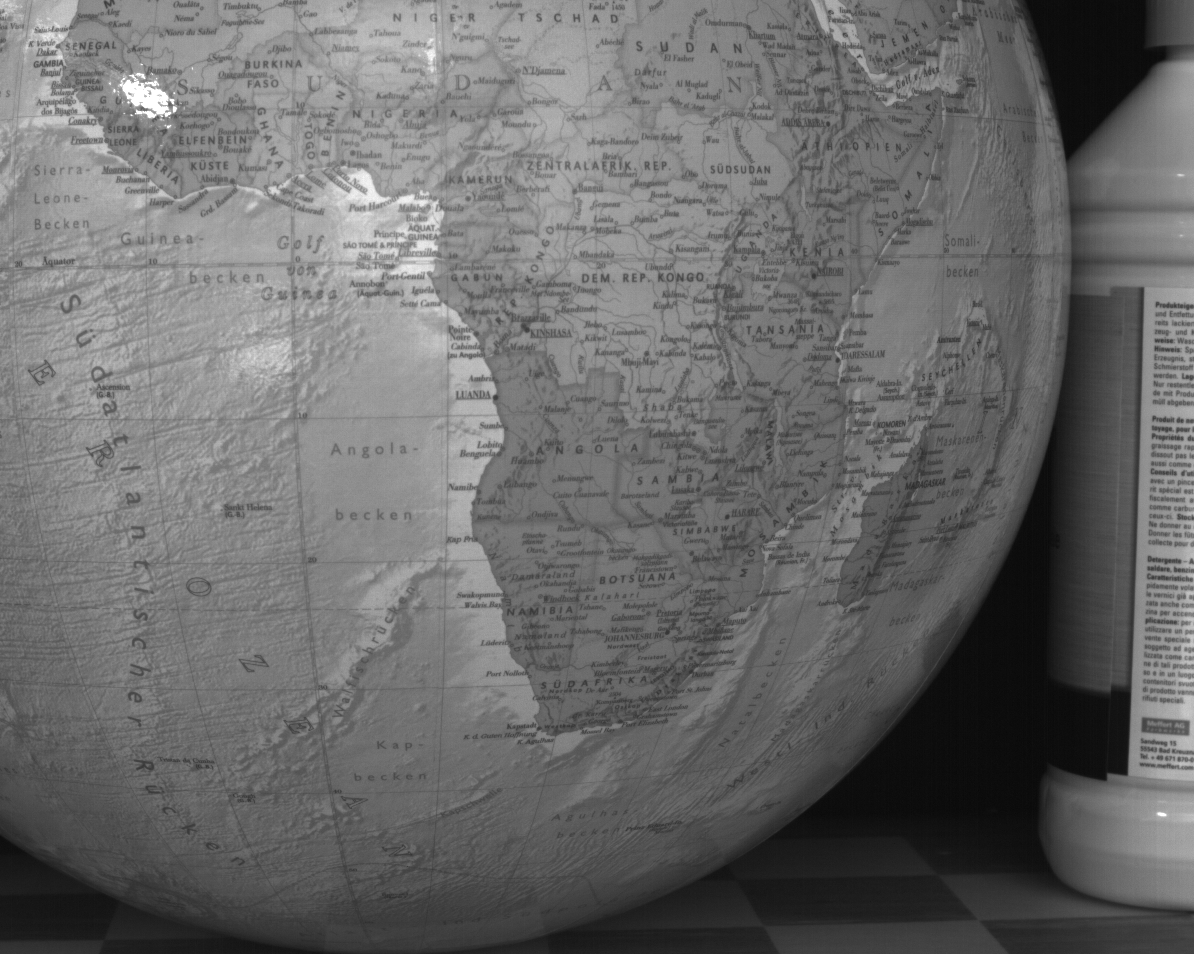}
	\includegraphics[width=0.19\linewidth]{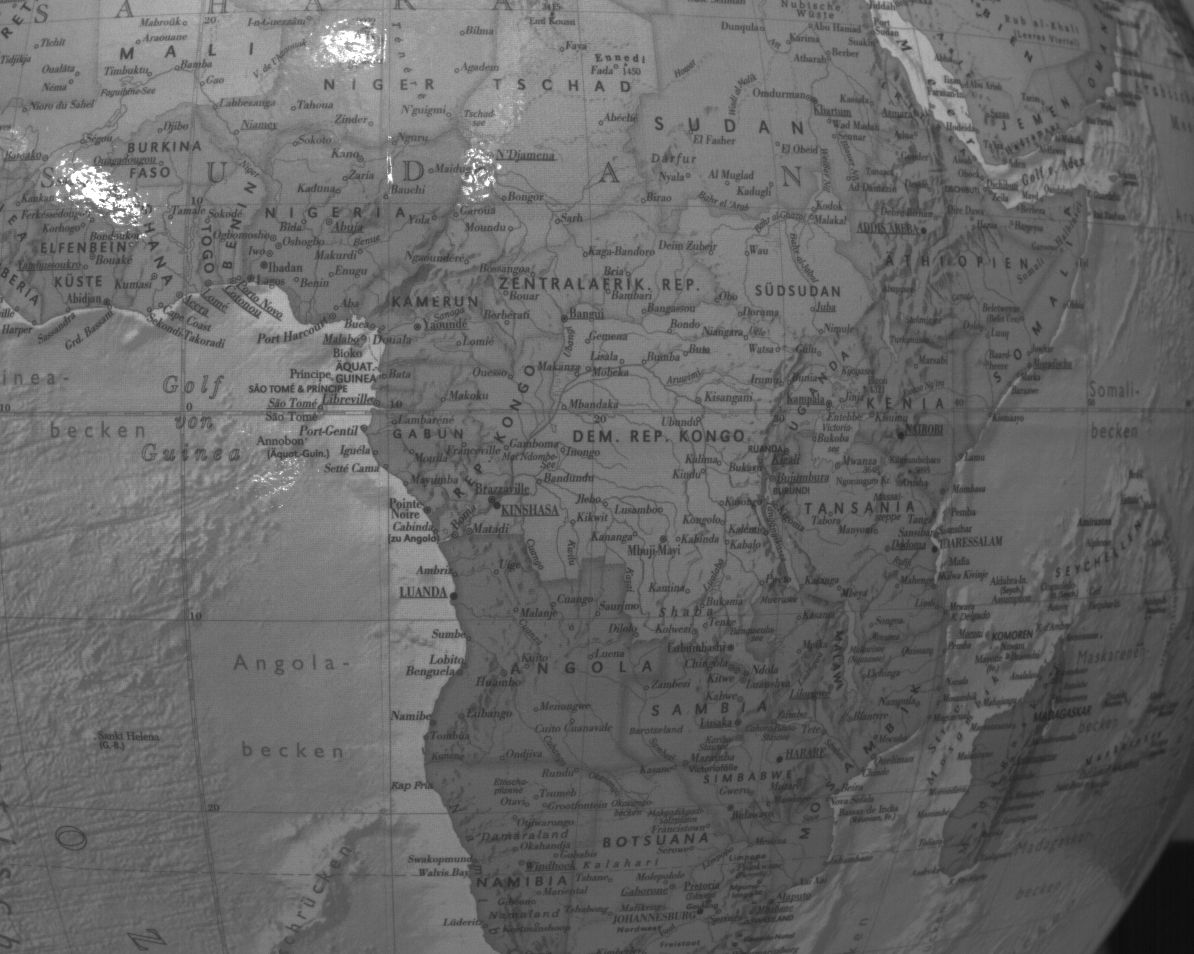}
	\includegraphics[width=0.19\linewidth]{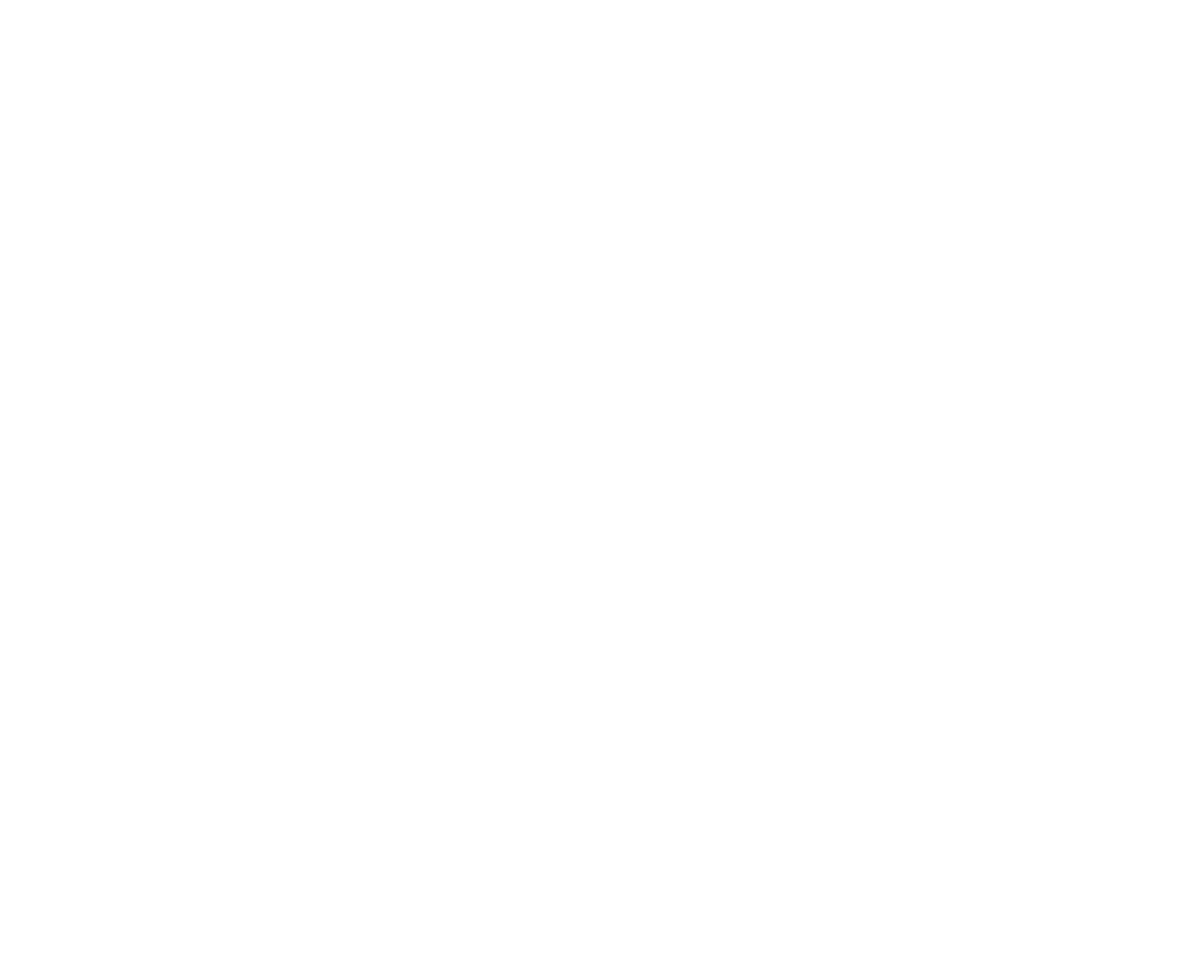}
	\caption{Overview of the scenes covered by our database.}
	\label{fig:databaseOverview}
\end{figure}

\begin{table}[!t]
	\footnotesize
	\caption{Motion types and photometric conditions in our datasets.}
	\centering
	\begin{tabular}{clll}
		\toprule
		\multicolumn{2}{l}{\textbf{Motion type}}	& \textbf{Camera trajectory} 	& \textbf{Photom. cond.}\\
		\midrule
		\multirow{4}{*}{\rotatebox[origin=c]{90}{\textbf{Global}}} 
		& translation $z$											& linear 											& day + night \\
		& translation $x$,$z$									& sinusoidal 									& day + night \\
		& panning															& circular 										& day + night \\
		& translation $x$,$y$,$z$, pan 				& sinusoidal + circular 			& day + night \\
		\midrule
		\multirow{5}{*}{\rotatebox[origin=c]{90}{\textbf{Mixed}}} 
		& static background										& static 											& day + night \\
		& translation $z$											& linear 											& day + night \\
		& translation $x$,$z$									& sinusoidal 									& day + night \\
		& panning															& circular 										& day + night \\
		& translation $x$,$y$,$z$, pan 				& sinusoidal + circular 			& day + night \\
		\bottomrule
	\end{tabular}
	\label{tab:motionAndPhotometricTypes}
\end{table}

\section{SupER Benchmark Database}
\label{sec:SuperResolutionBenchmarkDataset}

We acquire LR and HR images in a multi-resolution scheme with a single camera by capturing \textit{stop-motion} videos. At each time step of a stop-motion video, the underlying scene, the environmental conditions, as well as the camera pose are kept static. For consecutive time steps, the scene undergoes changes related to camera and/or object movements or environmental variations. One time step is represented by the $(n+1)$-tuple $(\vec{X}_{\mathrm{gt}}, \vec{Y}_{b_1}, \ldots, \vec{Y}_{b_n})$, where $\vec{X}_{\mathrm{gt}}$ denotes a ground truth HR image of size $N_u \times N_v$ and $\vec{Y}_{b_i}$, $i = 1, \ldots, n$ are LR frames of size $N_u/b_i \times N_v/b_i$ that are captured with $n$ different hardware binning factors $b_i$.

\subsection{Image Formation and Data Acquisition Scheme}
\label{sec:ImageAcquisition}

\begin{figure}[!t]
	\centering
	\subfloat[Basler acA2000-50gm camera and an example for a scene]{
		\includegraphics[width=0.236\textwidth]{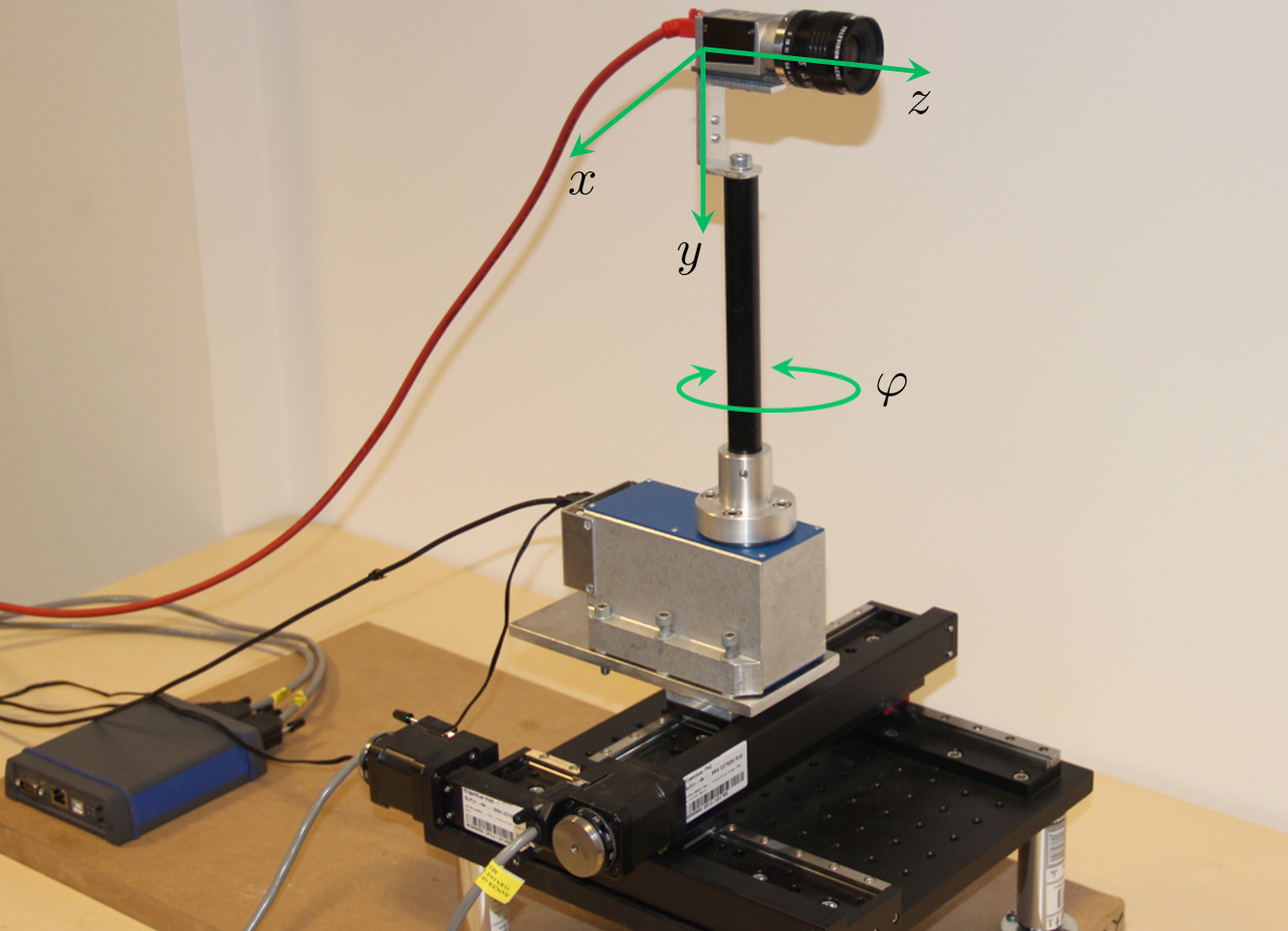}~
		\includegraphics[width=0.236\textwidth]{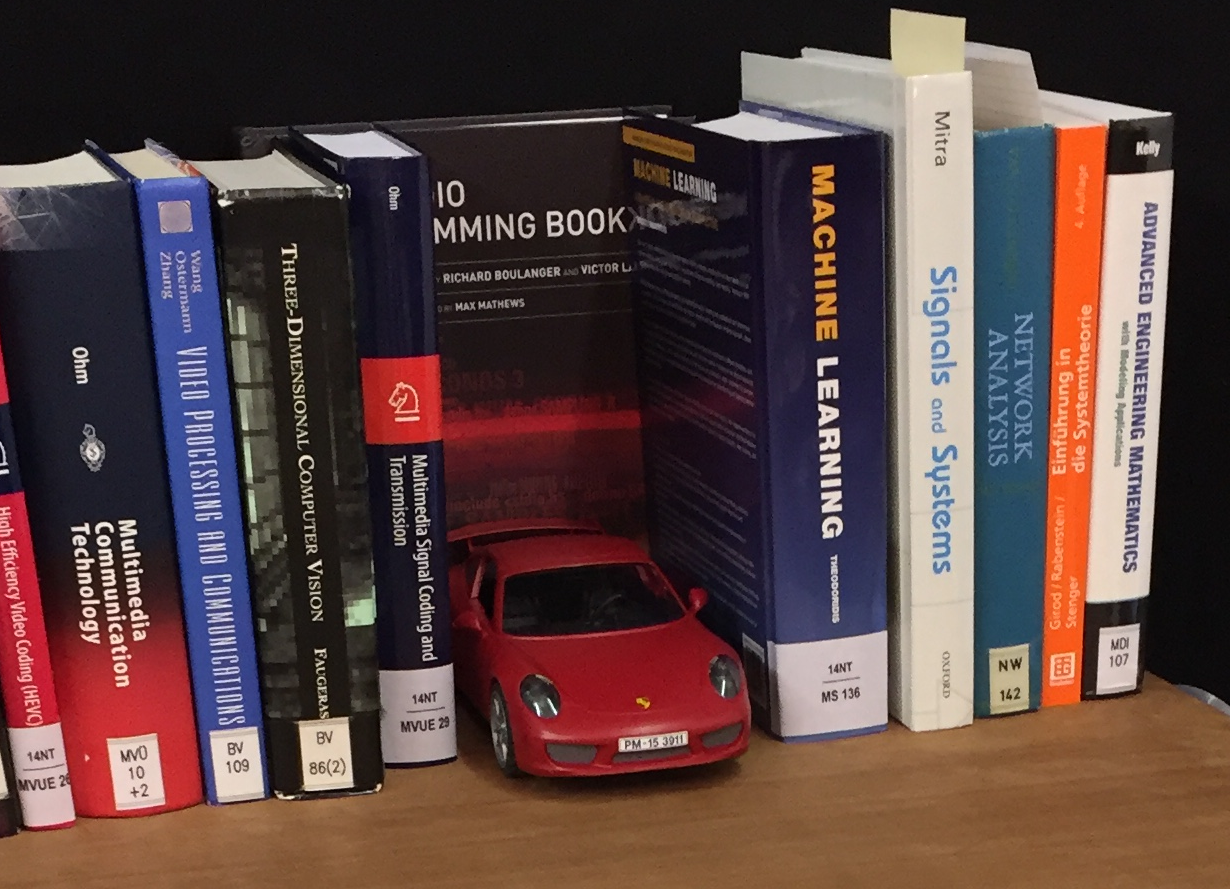}\label{fig:hardwareSetup:camera}}\\
	\subfloat[Local object motion for the \textit{pencils} scene]{
		\includegraphics[width=0.236\textwidth]{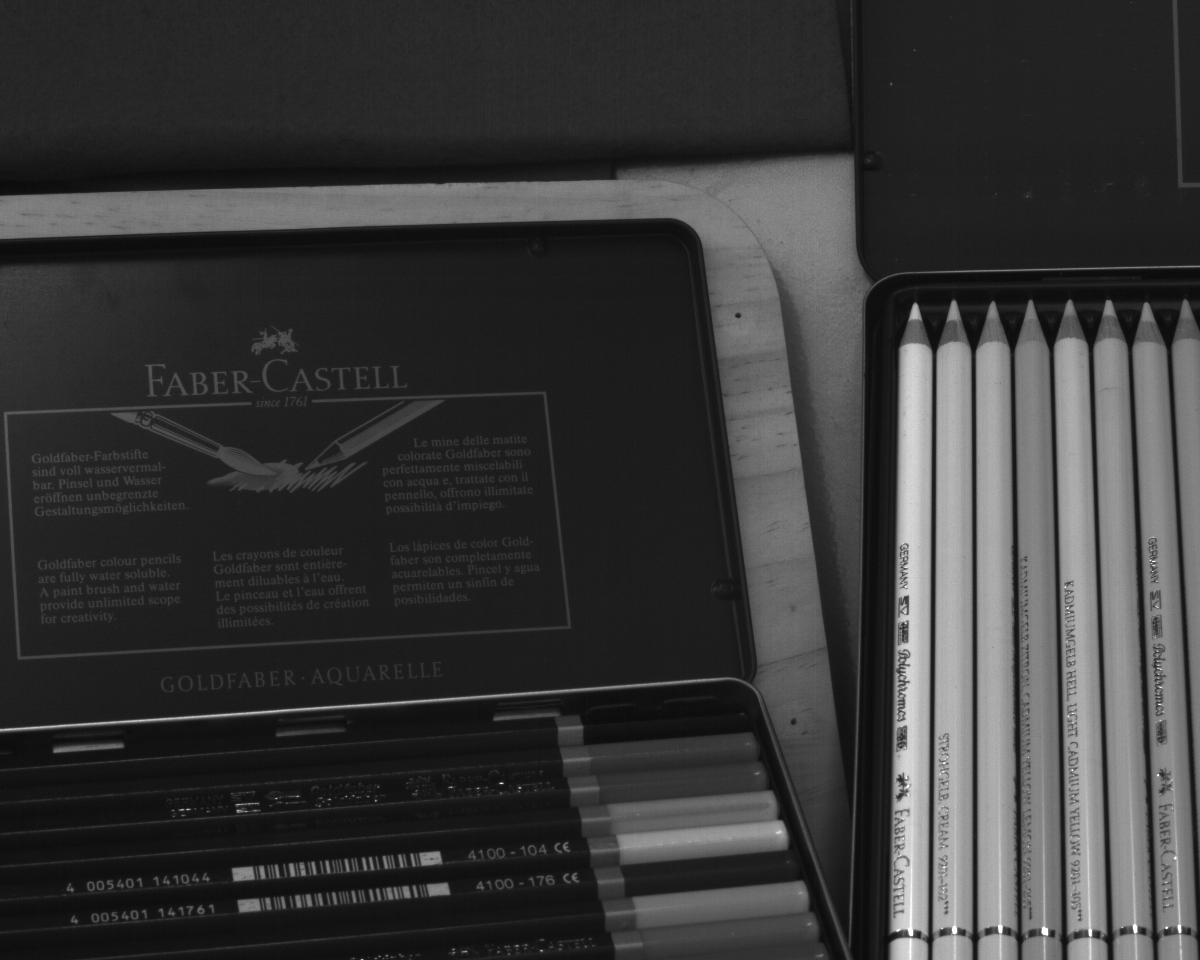}~
		\includegraphics[width=0.236\textwidth]{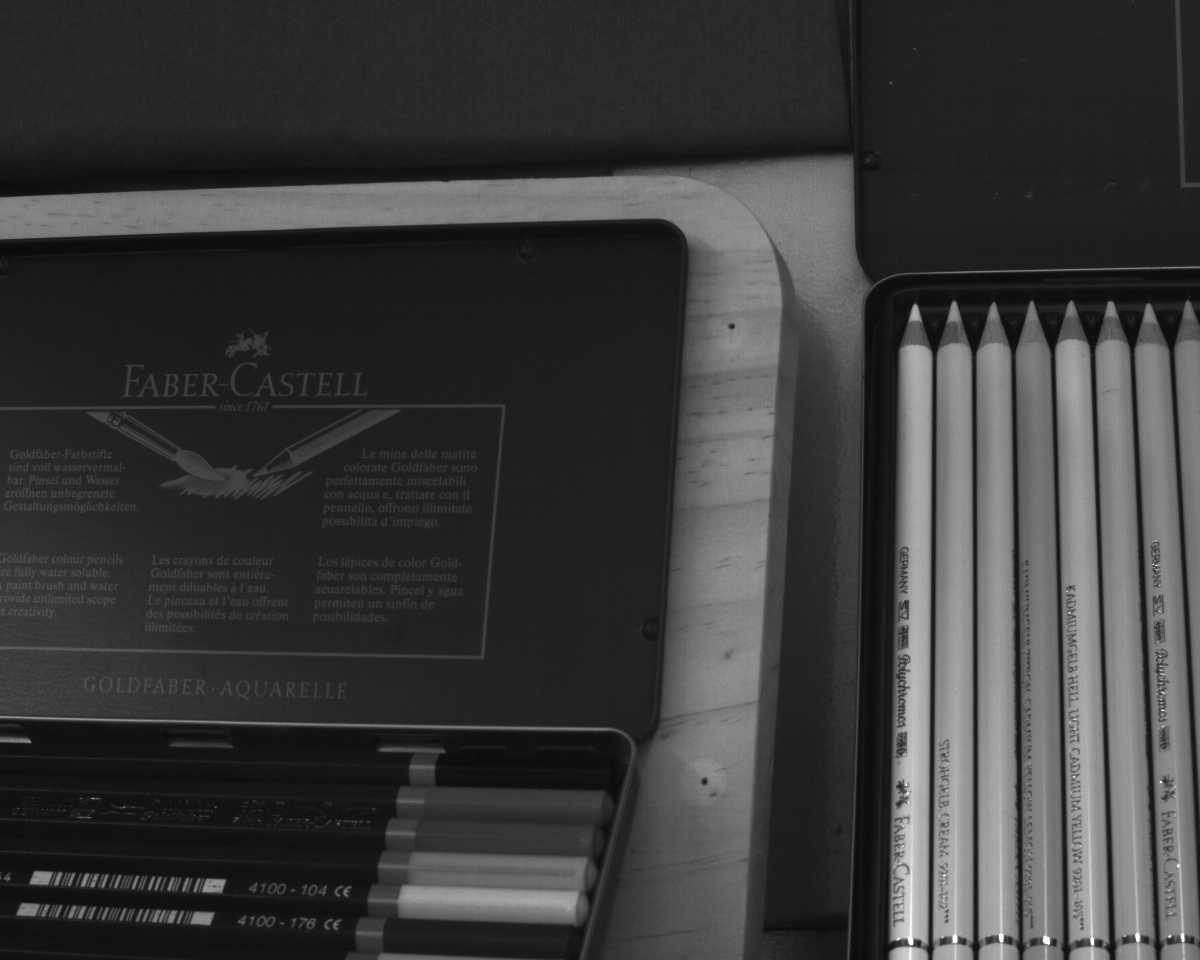}\label{fig:hardwareSetup:localMotion}}\\
	\subfloat[Photometric conditions (daylight and nightlight) for the \textit{games} scene]{
		\includegraphics[width=0.236\textwidth]{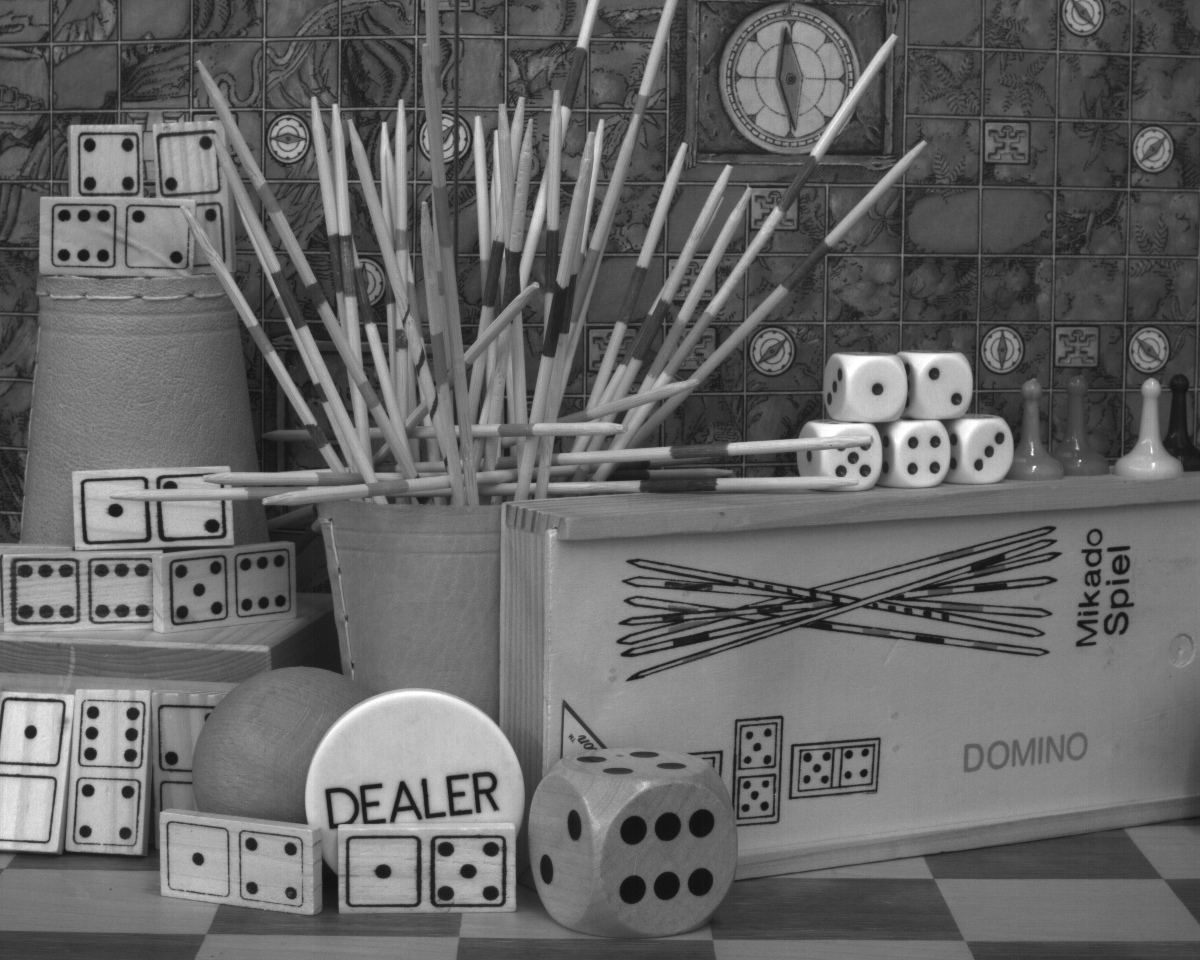}~
		\includegraphics[width=0.236\textwidth]{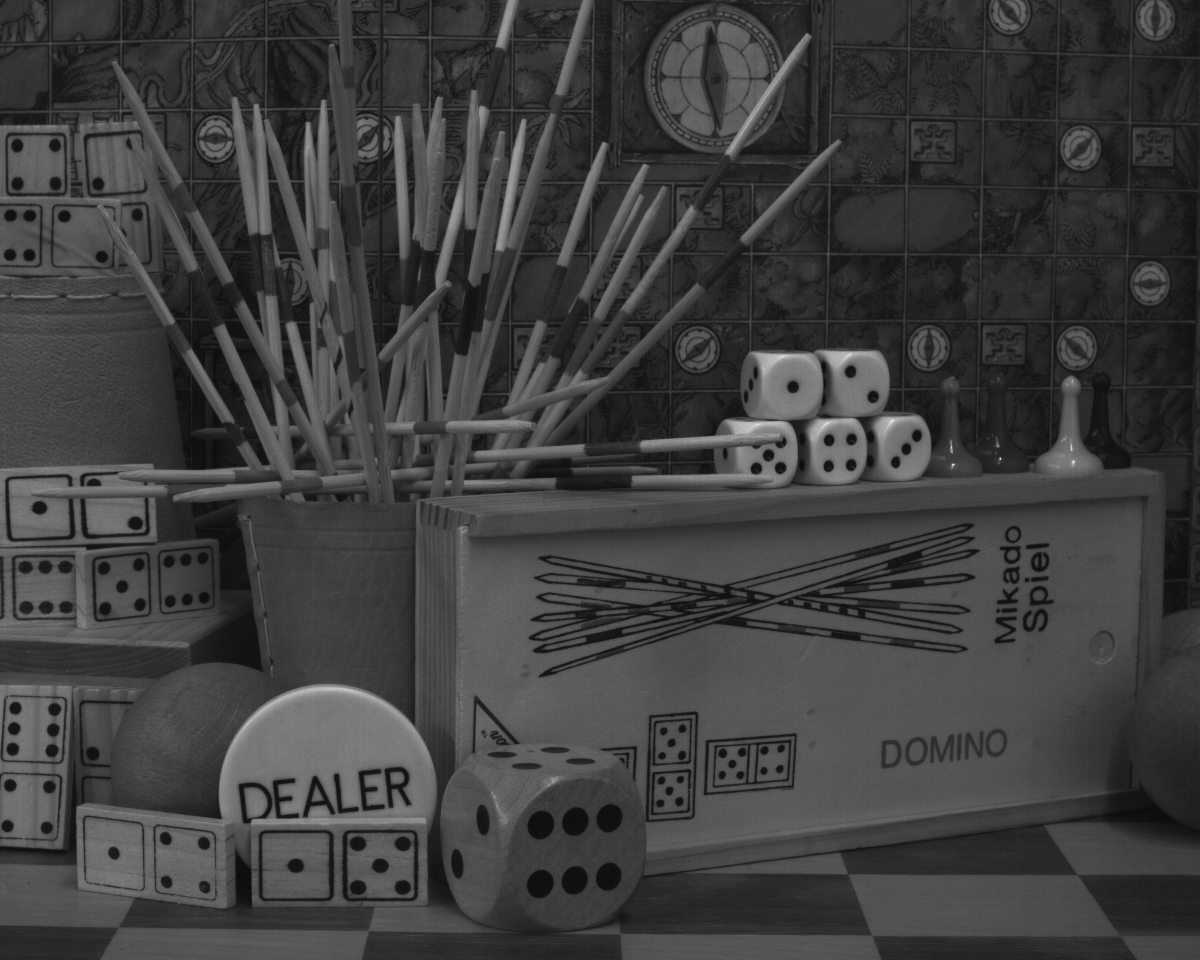}\label{fig:hardwareSetup:photo}}\\
	\caption{Hardware setup, and examples for motion types and photometric variations acquisitions (second and third row).}
	\label{fig:hardwareSetup}
\end{figure}

\newcolumntype{P}[1]{>{\centering\arraybackslash}p{#1}}
\begin{table*}[!t]
	\footnotesize
	\caption{Comparison of our SupER database to other publicly available benchmark datasets (excluding datasets without separate LR data). Unlike existing datasets, we provide captured image sequences at multiple spatial resolution levels including ground truth HR images. All quantitative properties (number of sequences, images, and resolution levels) refer to the original versions of the datasets.}
	\centering
	\begin{tabular}{l P{7.4em} P{7.5em} P{7.5em} P{7.5em} P{7.5em} P{7.5em}}
		\toprule
		\textbf{Dataset}	& \textbf{Content} & \textbf{Real/Simulated} & \textbf{\# Sequences}	& \textbf{\# LR images}	& \textbf{\# Ground truth}	& \textbf{\# Res. levels} \\
		\midrule
		MDSP \cite{Farsiu2014}						& Mixed scenes		& Real						& 21								 	& 915				& \xmark		& 1 \\
		Vandewalle \cite{Vandewalle2016}	& Mixed scenes		& Real						& 3										& 12				& \xmark		& 1 \\
 		Liu and Sun \cite{Liu2014}				& Natural scenes	& Simulated				& 4										&	171				&	171				& 2 \\
		Yang \etal \cite{Yang2014a}				& Natural scenes	& Simulated				& Single images only	& 2,061			& 229				& 4 \\
		Qu \etal \cite{Qu2016}						& Faces 					& Real						& Single images only	& 93				& 93				& 2 \\
		\midrule
		\textbf{SupER (ours)}							& Mixed scenes		& Real						& 254									& 17,145		& 5,715			& 4 \\
		\bottomrule
	\end{tabular}
	\label{tab:databaseOverview}
\end{table*}

To gain the ground truth $\vec{X}_{\mathrm{gt}}$, we capture $L$ frames $\vec{X}^{(l)}$, $l = 1, \ldots, L$ at each time step of a stop-motion video using the actual pixel resolution of the camera. The ground truth is computed by averaging over $L$ ($L = 10$) consecutive frames according to $\vec{X}_{\mathrm{gt}} = \frac{1}{L} \sum_{l = 1}^L \vec{X}^{(l)}$ to alleviate sensor noise. 

To obtain the LR data $\vec{Y}_{b_i}$, we use camera hardware binning. This reduces pixel resolution by aggregating adjacent pixels on the sensor array, see \fref{fig:flowchart} (top). Let $x(\vec{u})$, $\vec{u} \in \mathbb{R}^2$ be an irradiance light field \cite{Lin2004}. Then, hardware binning links $x(\vec{u})$ to a discrete image $\vec{Y}_{b}$ by
\begin{equation}
	\label{eqn:hardwareBinning}
	\vec{Y}_{b} = \mathcal{Q} \left\{ \mathcal{D}_b \left\{ x(\vec{u}) \right\} + \vec{\epsilon} \right\}\enspace,
\end{equation}
where $\mathcal{D}_b\{ \cdot \}$ denotes sampling according to the binning factor $b$, $\mathcal{Q}\{ \cdot \}$ denotes quantization to capture image intensities, and $\vec{\epsilon}$ is additive noise. The sampling is modeled by:
\begin{equation}
	\label{eqn:hardwareBinningSamplingOp}
	\mathcal{D}_b \left\{ x(\vec{u}) \right\} = \left( \vec{H}_{\mathrm{sensor}, b} \star \vec{H}_{\mathrm{optics}} \star x \right) (\vec{u})\enspace,
\end{equation}
where $\vec{H}_{\mathrm{optics}}$ denotes the optical point spread function
(PSF), $\vec{H}_{\mathrm{sensor}, b}$ models the spatial integration over $b
\times b$ pixels on the sensor array, and $\star$ is the convolution operator
\cite{Lin2004}. As we use a single optical system to capture HR and LR data,
$\vec{H}_{\mathrm{sensor}, b}$ is determined by the binning factor $b$ while
$\vec{H}_{\mathrm{optics}}$ is constant for different binning factors. We used
high-quality optical equipment for the acquisition, such that
$\vec{H}_{\mathrm{sensor}, b}$ is the main limiting factor for resolution and
signal degradations. In \sref{sec:ComparisonToExistingDatasets}, we compare
the proposed hardware binning to the closely related software binning.

We capture raw LR and ground truth data in the proposed multi-resolution scheme while camera internal processing is avoided. This enables to explicitly investigate SR under different types of postprocessing, \eg white balancing or compression. As the majority of SR algorithms deal either with grayscale or a single luminance channel while the chrominance is simply interpolated, we limited ourselves to monochromatic acquisitions. To study color SR algorithms that super-resolve only the luminance layer of color images \cite{Dirks2016,Dong2014,Kappeler2016,Timofte2016}, our monochromatic data can serve as a luminance channel\footnote{To study full color SR, our setup can be generalized to provide multiple channels, \eg using color filters or a full RGB camera.}. Our database covers 14 scenes including text, emulated surveillance scenes, and various objects with $n = 3$ binning factors $b \in \{ 2, 3, 4 \}$, see \fref{fig:databaseOverview}.

\subsection{Motion Types and Environmental Conditions}
\label{sec:MotionAndEnvironmentConditions}

We used the setup depicted in \fref{fig:hardwareSetup:camera} to capture datasets from mixed scenes. To this end, a Basler acA2000-50gm CMOS camera \cite{Basler2016} was mounted on a positioning stage. The camera pose was controlled by a stepper motor and a height-adjustable table. This enables camera panning in one dimension and translations in three dimensions. We considered nine motion types that were described by camera trajectories and translational and rotational object movements. The photometric conditions were controlled by artificial lighting and we considered bright illumination (\textit{daylight}) and low-light illumination (\textit{nightlight}). \Tref{tab:motionAndPhotometricTypes} summarizes motion and photometric conditions, which forms four dataset categories.
	\\[0.8ex]
	\textbf{Global motion.}
This category consists of data of static scenes under constant daylight conditions. All inter-frame motion was related to global camera motion. We captured translations in $z$-direction, translations in $x$- and $z$-direction, camera pan, and joint translation and pan, which followed linear, circular, and sinusoidal trajectories. The camera positions were uniformly distributed over the trajectories.
	\\[0.8ex]
	\textbf{Local motion.}
	This category consists of dynamic scenes captured under daylight conditions with a static background but moving objects, see \fref{fig:hardwareSetup:localMotion}. We considered translational and/or rotational object motion in the foreground. This necessitates the use of non-rigid models, \eg optical flow, to describe inter-frame motion. 
	\\[0.8ex]
	\textbf{Mixed motion.}
	This category combines the global and local motion datasets. Thus, each camera trajectory was combined with translational and/or rotational object motion.
	\\[0.8ex]
	\textbf{Photometric variation.}
	This category augments each of the aforementioned datasets by photometric variations. The datasets comprise sequences of $K$ frames, where the first $K - K_{\mathrm{day}}$ frames were taken from the global, local, and mixed motion datasets and the remaining $K_{\mathrm{night}}$ frames were obtained under nightlight conditions, see \fref{fig:hardwareSetup:photo}. We consider the nightlight images as photometric \textit{outliers} while daylight images are \textit{inliers}.

Our database comprises 56 global, 56 mixed, and 14 local motion image sequences with $K = 40$ frames each captured from 14 scenes. The photometric variation datasets augment each sequence by $K_{\mathrm{night}} = 5$ nightlight images. 

\subsection{Comparison to Existing Datasets}
\label{sec:ComparisonToExistingDatasets}

A comparison of our database to existing SR datasets is shown in \tref{tab:databaseOverview}. Regarding the analysis of SR algorithms, our image acquisition scheme features several favorable advantages over existing strategies. Most importantly, it goes beyond existing real-world databases \cite{Farsiu2014,Vandewalle2016} by providing 1) real LR acquisitions, and 2) a corresponding HR ground truth. This enables quantitative benchmarks as opposed to subjective evaluations by visual inspection. The existence of ground truth data also circumvents the use of no-reference quality measures for quantitative studies.

In contrast to simulated datasets \cite{Liu2014,Yang2014a}, our image formation
model is based on hardware binning. It is worth noting that hardware binning
according to \eqref{eqn:hardwareBinning} is different to \textit{software
binning} that uses the image formation model
\begin{equation}
	\label{eqn:softwareBinning}
	\vec{Y}_{b} = \mathcal{Q} \left\{ \mathcal{D}_b \left\{ \vec{X} \right\} + \vec{\eta} \right\}\enspace,
\end{equation}
where $\vec{X}$ is a discretized version of $x(\vec{u})$ and $\vec{\eta}$ is
simulated noise. In most evaluations of current SR algorithms, $\vec{X}$ is
chosen as a reference image from an existing database,
\eg
LIVE \cite{Sheikh2016}, Set5, Set14, B100, or L20 \cite{Timofte2016}, or from
HR videos \cite{Liu2014}.
The LR image $\vec{Y}_{b}$ is simulated according to
\eqref{eqn:softwareBinning} and $\vec{X}$ can be considered as a ground truth
for SR.
Note that a simulation cannot model the true physics of image formation since
it does not have access to the original irradiance light field $x(\vec{u})$ as
used in \eqref{eqn:hardwareBinning}. More specifically, simulated datasets are
usually based on simplified models for $\vec{\eta}$, \eg zero-mean Gaussian
noise \cite{Yang2014a}, while LR images in our database are degraded by real
non-Gaussian noise. Moreover, our hardware setup is flexible to consider
different environmental conditions, \eg photometric variations or local object
motion, that are by design physically correct. Some prior works also use the
same models for image simulations and reconstruction. This can be seen as
\textit{inverse crime} \cite{Wirgin2004} and limits the significance of experimental evaluations.

One key advantage of our setup  is that it guarantees by design a perfect
alignment between LR and ground truth data.  This allows pixel-wise comparisons
among super-resolved images and the ground truth by full-reference quality
measures. In contrast to \cite{Qu2016}, our ground truth is not the outcome of
a potentially error-prone registration procedure. Our corresponding LR images
also cover multiple resolution levels.  Additionally, our database consists of
image sequences instead of single images.  This makes the data usable for both,
SISR and MFSR, and enable studies of model parameters, \eg the magnification
factor or the number of input frames.

\section{Benchmark Setup}
\label{sec:BenchmarkSetup}

\noindent
\textbf{Evaluation protocol.}
We perform SR on $K = 2L + 1$ consecutive LR frames $\vec{Y}^{(-L)}, \ldots, \vec{Y}^{(0)}, \ldots, \vec{Y}^{(L)}$. $\vec{Y}^{(0)}$ is referred to as the reference frame. For SISR, $\vec{Y}^{(0)}$ serves as input to determine the corresponding HR image $\vec{X}_{\mathrm{sr}}$. In case of MFSR, $\vec{Y}^{(-L)}, \ldots, \vec{Y}^{(L)}$ is exploited to obtain $\vec{X}_{\mathrm{sr}}$ using variational optical flow \cite{Liu2009} to estimate subpixel motion towards $\vec{Y}^{(0)}$. For MFSR with customized motion compensation (\eg \cite{Kappeler2016} or \cite{Ma2015}), we employ the optical flow estimation used in the original versions. We study the magnification factors 2, 3, and 4 to super-resolve LR images at the respective binning factors to the resolution of the ground truth. The number of input frames for MFSR is chosen according to the desired magnification such that the underlying image reconstruction problem is not underdetermined. We use 5, 11, and 17 frames for the magnification factors 2, 3, and 4, respectively. Experimental results for other sequence lengths are shown in our supplementary material.

\begin{table}[!t]
	\footnotesize
	\caption{Categorization of the SR algorithms in our benchmark.}
	\centering
	\begin{tabular}{|l|l|l|l|}
		\hline %\toprule
		\textbf{Category}			& \textbf{Single-image}			& \textbf{Multi-frame} 				&	\textbf{Hybrid}			 	\\
		\hline %\midrule
		Internal methods			& SESR \cite{Huang2015a}		&															&												\\
		(self-exemplars)			&														&															&												\\
		\hline %\midrule
		External methods			& SCSR \cite{Yang2010}			&	VSRNET \cite{Kappeler2016}	&	HYSR \cite{Batz2015}	\\
		(dictionary /	& EBSR \cite{Kim2010}				&															&												\\
		deep learning)				& NBSRF \cite{Salvador2015}	&															&												\\
													& SRCNN \cite{Dong2014}			&															&												\\
		\cline{1-3} %\midrule		
		Interpolation-based		&	BICUBIC										&	NUISR	\cite{Park2003}				&												\\
													& 													&	WNUISR \cite{Batz2016}			&												\\
													& 													& DBRSR \cite{Batz2016b}			&												\\
		\hline %\midrule	
		Reconstruction-based	&														&	L1BTV \cite{Farsiu2004a}		&												\\
		(non-blind)						& 													&	BEPSR \cite{Zeng2013}				&												\\
													& 													& IRWSR \cite{Kohler2015c}		&												\\
		\hline
		Reconstruction-based	&														& SRB \cite{Ma2015}						&												\\
		(blind)								&														&															&												\\
		\hline %\bottomrule
	\end{tabular}
	\label{tab:algorithmOverview}
\end{table}

We use four full-reference quality measures $Q(\vec{X}_{\mathrm{sr}})$ to assess the fidelity of super-resolved data $\vec{X}_{\mathrm{sr}}$ \wrt the ground truth $\vec{X}_{\mathrm{gt}}$ assuming that $\vec{X}_{\mathrm{sr}}$ and $\vec{X}_{\mathrm{gt}}$ are aligned\footnote{If a super-resolved image is not aligned to the ground truth, we compensate for this misalignment and assess the fidelity on the overlap region.}. The peak-signal-to-noise ratio (PSNR) is used to determine the fidelity on a pixel level. As structural measures, we use the structural similarity index (SSIM) \cite{Wang2004} and multi-scale SSIM (MS-SSIM) \cite{Wang2003}. We use the wavelet-based information fidelity criterion (IFC)~\cite{Sheikh2005} to consider natural scene statistics on an information theoretic level. Higher PSNR, SSIM, MS-SSIM, and IFC scores express higher similarity with the ground truth. Recent SISR studies \cite{Yang2014a} showed that human visual perception better correlates with IFC and MS-SSIM than with PSNR and SSIM.

Note that scene content can considerably influences the absolute values of these measures \cite{Yang2014a}. To reduce dependency from scene content and to study the improvement of SR over the input data, 
we additionally evaluate normalized versions $\tilde{Q}(\vec{X}_{\mathrm{sr}})$ of each quality measure, defined as
\begin{equation}
	\tilde{Q}(\vec{X}_{\mathrm{sr}}) = \big( Q(\vec{X}_{\mathrm{sr}}) - Q(\tilde{\vec{Y}}^{(0)}) \big) / Q(\tilde{\vec{Y}}^{(0)})\enspace,
\end{equation}
where $\tilde{\vec{Y}}^{(0)}$ denotes the nearest-neighbor interpolation of the
reference frame $\vec{Y}^{(0)}$ on the target HR grid. Absolute performances and
further results are in the supplemental material.
\\[0.8ex]
\textbf{Evaluated algorithms.}
Besides bicubic interpolation, we study 14 classical and state-of-the-art methods as categorized in \tref{tab:algorithmOverview}. Interpolation-based MFSR comprises conventional non-uniform interpolation (NUISR) \cite{Park2003}, NUISR with outlier weighting (WNUISR) \cite{Batz2016} and denoising-based refinement (DBRSR) \cite{Batz2016b}. The reconstruction-based methods include non-blind $L_1$ norm reconstruction with bilateral total variation prior (L1BTV) \cite{Farsiu2004a}, adaptive bilateral edge preserving prior (BEPSR), and iteratively re-weighted minimization (IRWSR) \cite{Kohler2015c} as well as blind SR with motion blur handling (SRB) \cite{Ma2015}. As a representative of deep learning, we use the video SR neural network (VSRNET) \cite{Kappeler2016}. In terms of SISR using external data, we study example-based ridge regression (EBSR) \cite{Kim2010}, dictionary sparse coding (SCSR) \cite{Yang2010}, the Naive Bayes SR forest (NBSRF) \cite{Salvador2015}, and convolutional neural networks (SRCNN) \cite{Dong2014}. As an internal method, we studied transformed self-exemplars (SESR) \cite{Huang2015a}. Furthermore, we use the hybrid approach (HYSR) proposed in \cite{Batz2015} that adaptively combines EBSR with NUISR.

\begin{figure*}[!t]
	%\centering
	\subfloat{\includegraphics[width=1.00\textwidth]{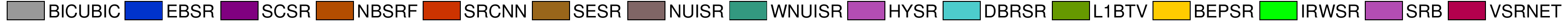}}\\[-1.0ex]
	\rotatebox{90}{\scriptsize Norm. PSNR\\}~
	\subfloat{\includegraphics[width=0.315\textwidth]{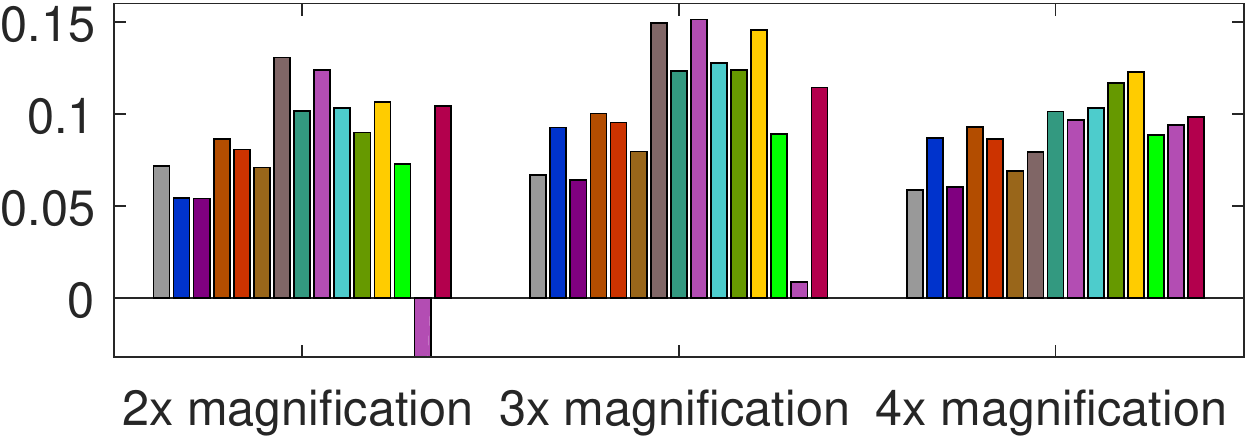}}~
	\subfloat{\includegraphics[width=0.315\textwidth]{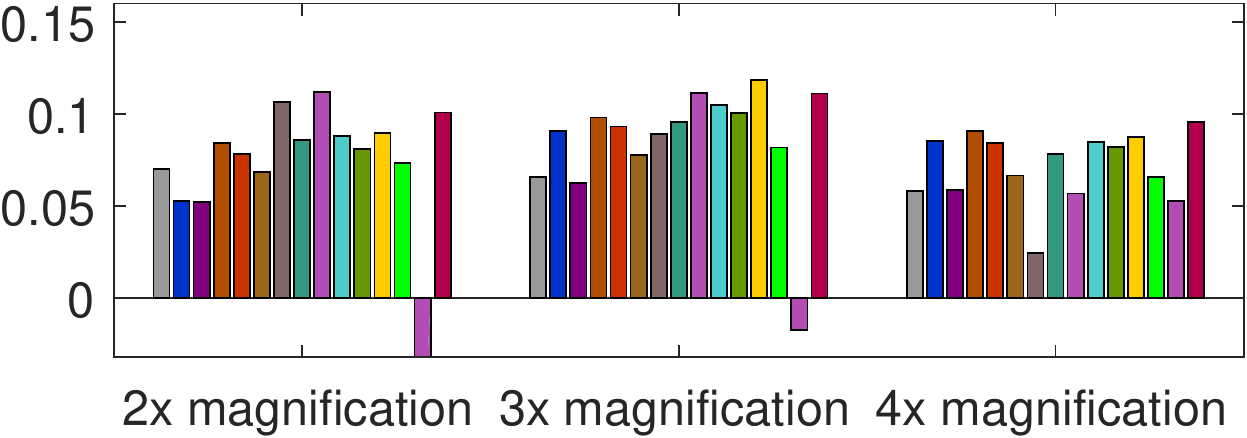}}~
	\subfloat{\includegraphics[width=0.315\textwidth]{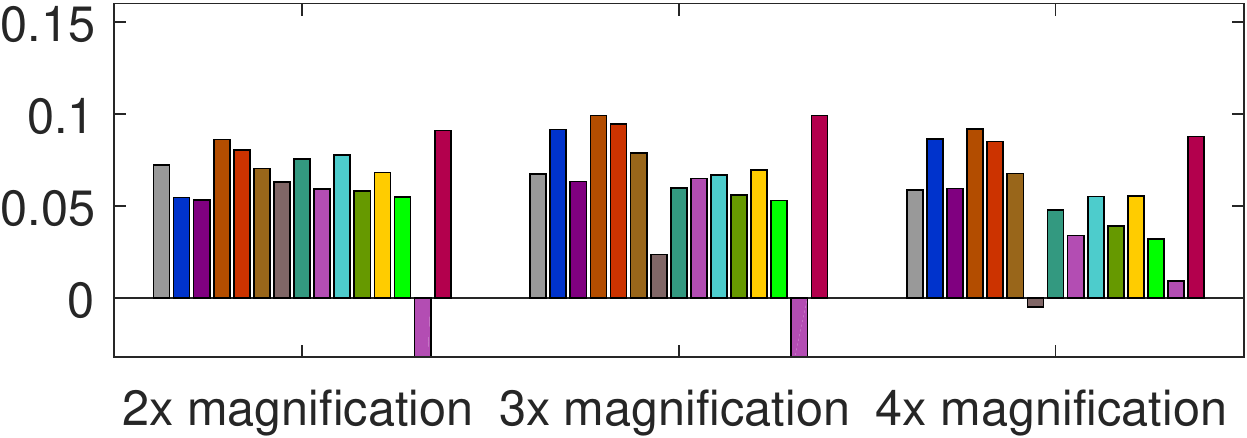}}\\[-0.8ex]
	\rotatebox{90}{\scriptsize Norm. SSIM}~
	\subfloat{\includegraphics[width=0.315\textwidth]{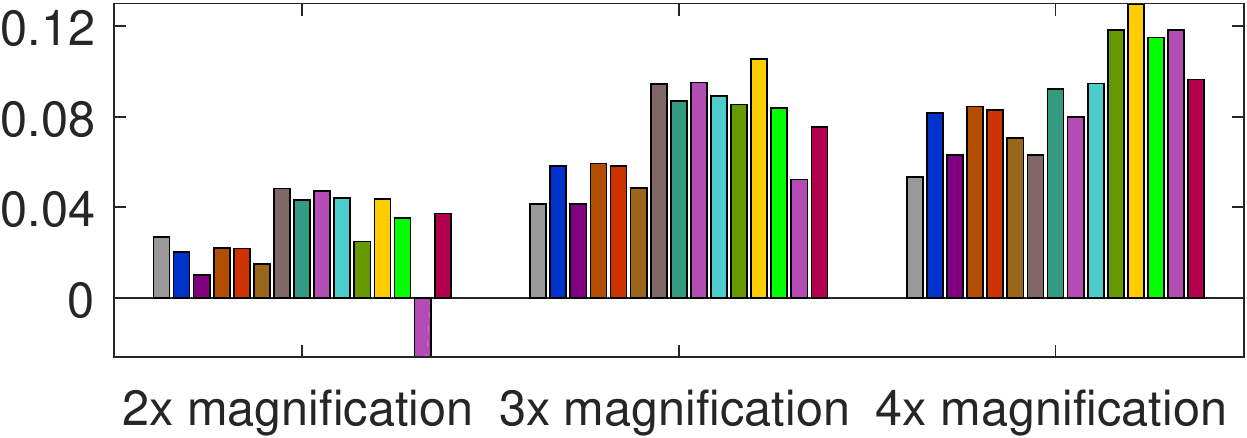}}~
	\subfloat{\includegraphics[width=0.315\textwidth]{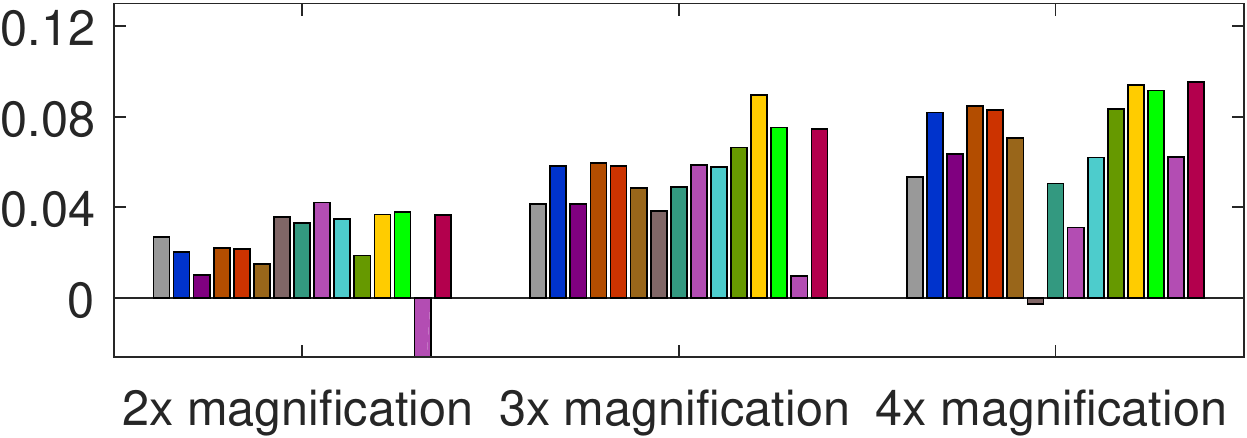}}~
	\subfloat{\includegraphics[width=0.315\textwidth]{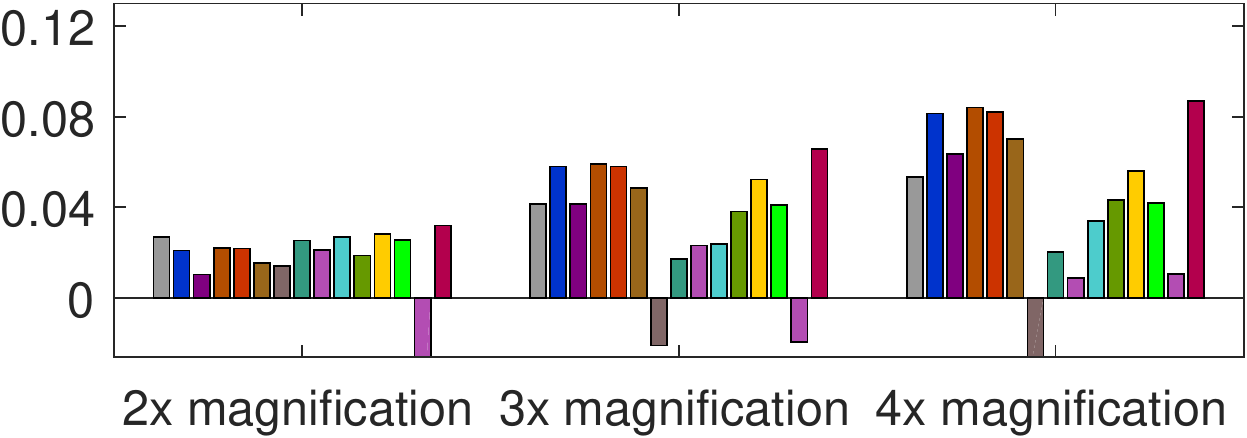}}\\[-0.8ex]
	\rotatebox{90}{\scriptsize Norm. MS-SSIM}~
	\subfloat{\includegraphics[width=0.315\textwidth]{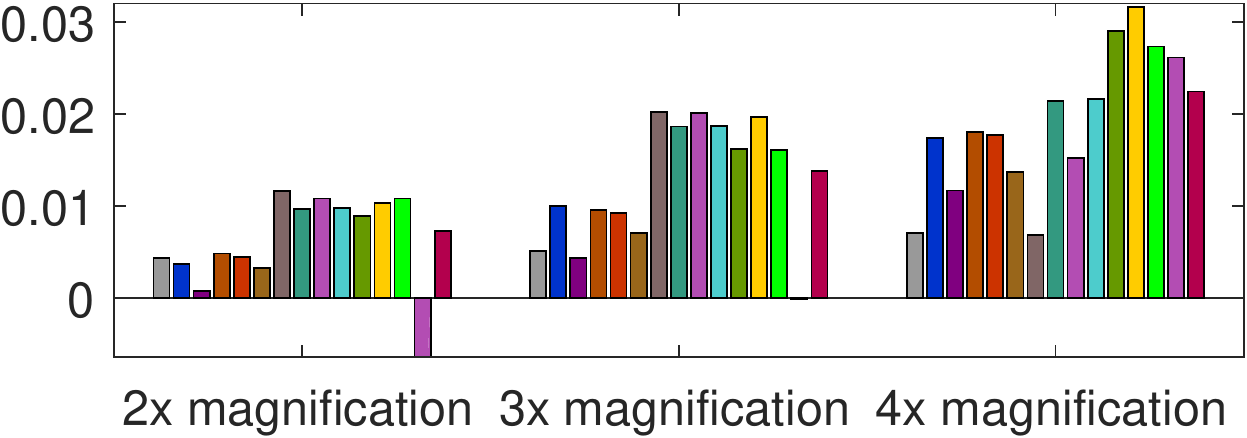}}~
	\subfloat{\includegraphics[width=0.315\textwidth]{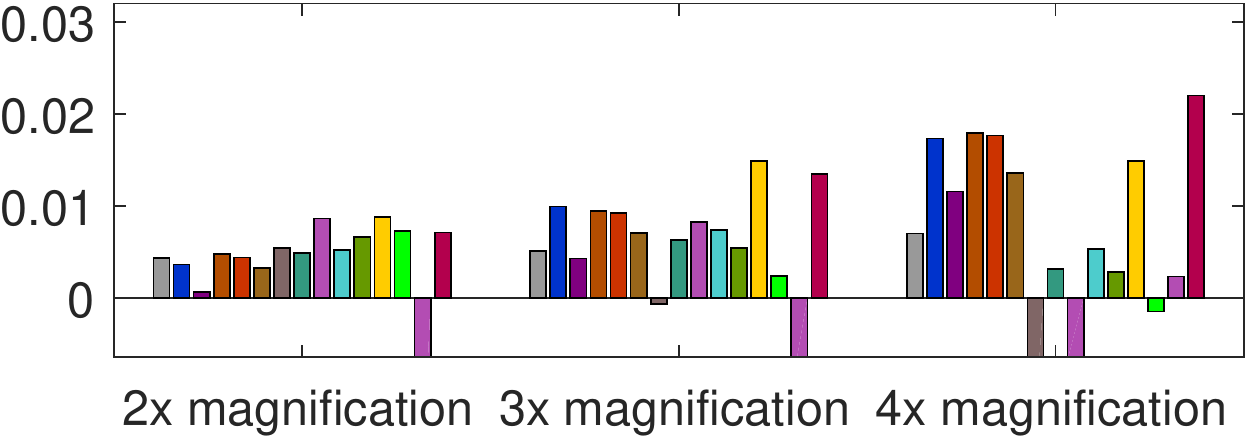}}~
	\subfloat{\includegraphics[width=0.315\textwidth]{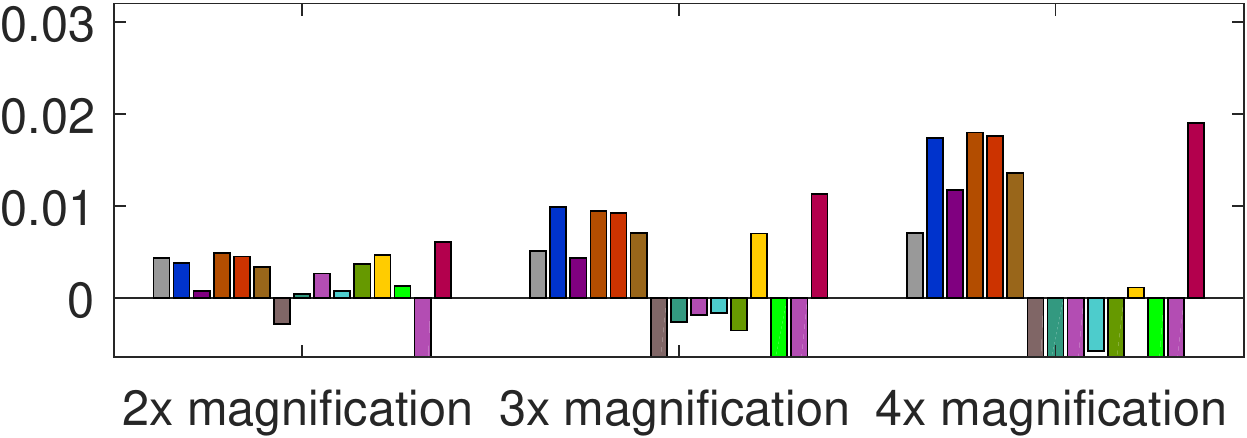}}\\[-0.8ex]
	\setcounter{subfigure}{0}
	\rotatebox{90}{\scriptsize Norm. IFC}~
	\subfloat[Global motion datasets]{\includegraphics[width=0.315\textwidth]{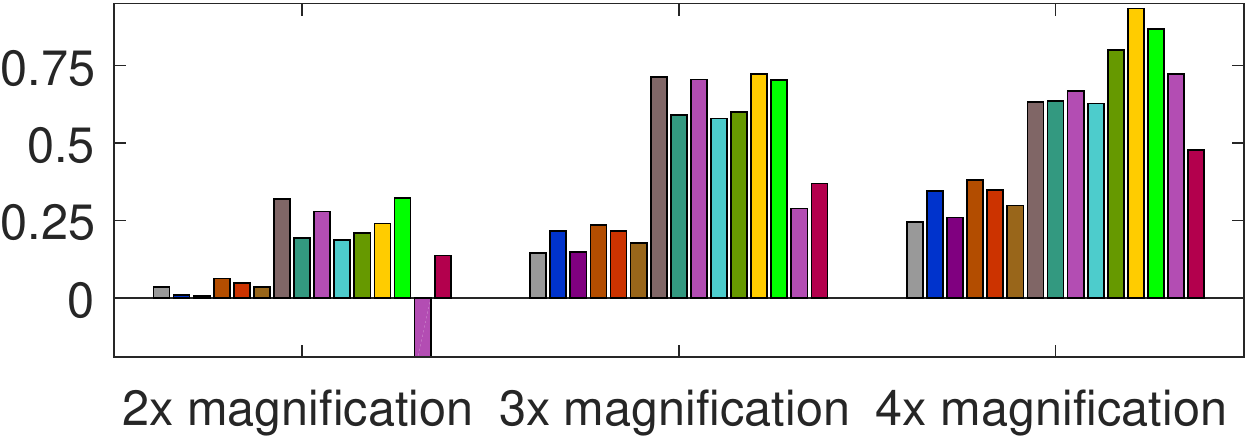}
		\label{fig:srBenchmarkMotionTypes:global}}~
	\subfloat[Mixed motion datasets]{\includegraphics[width=0.315\textwidth]{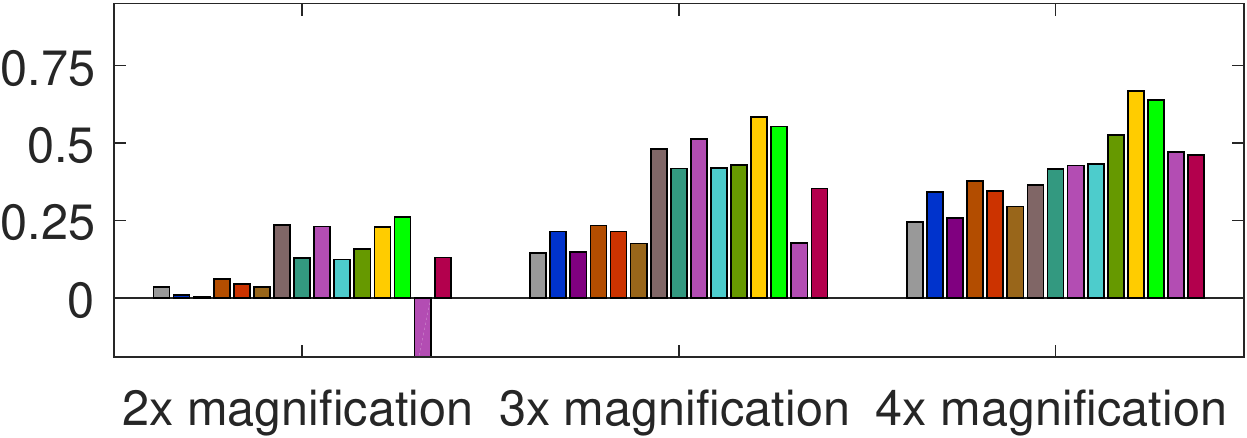}
		\label{fig:srBenchmarkMotionTypes:local}}~
	\subfloat[Local motion datasets]{\includegraphics[width=0.315\textwidth]{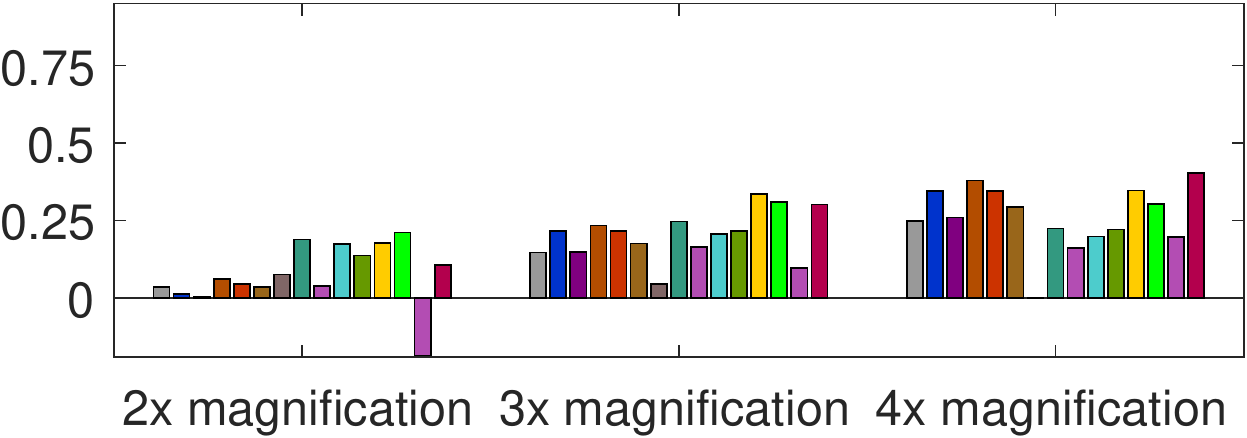}
		\label{fig:srBenchmarkMotionTypes:staticBackground}}
	\caption{Benchmark of the SR algorithms on our global \protect\subref{fig:srBenchmarkMotionTypes:global}, mixed \protect\subref{fig:srBenchmarkMotionTypes:local}, and local motion datasets \protect\subref{fig:srBenchmarkMotionTypes:staticBackground}. From top to bottom: average normalized PSNR, SSIM, MS-SSIM, and IFC relative to low-resolution data for $2\times$, $3\times$ and $4\times$ magnification. Note that a negative normalized measure indicates that super-resolved data is worse than low-resolution input data (figure best viewed in color).}
	\label{fig:srBenchmarkMotionTypes}
\end{figure*}

\begin{figure*}[!t]
	\centering
	\mbox{
	\centering
	\subfloat{
		\begin{tikzpicture}[spy using outlines={rectangle,red,magnification=4.0,height=1.5cm, width=2.8cm, connect spies, every spy on node/.append style={thick}}] 
			\node {\pgfimage[width=0.16\linewidth]{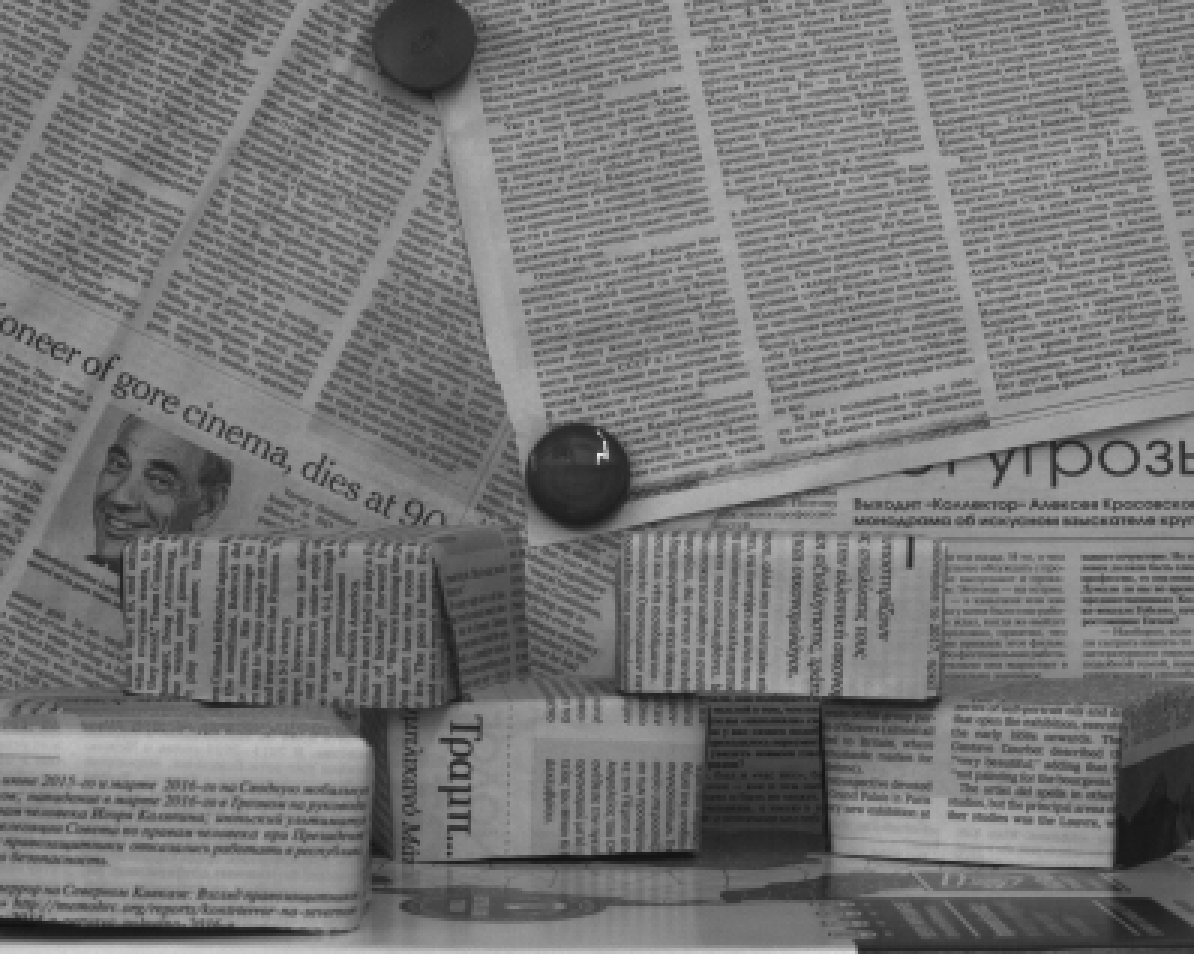}}; 
      \spy on (0.75, -0.1) in node [left] at (1.4, -1.9); 
    \end{tikzpicture}
	}\hspace{-1.2em}
	\subfloat{
		\begin{tikzpicture}[spy using outlines={rectangle,red,magnification=4.0,height=1.5cm, width=2.8cm, connect spies, every spy on node/.append style={thick}}] 
			\node {\pgfimage[width=0.16\linewidth]{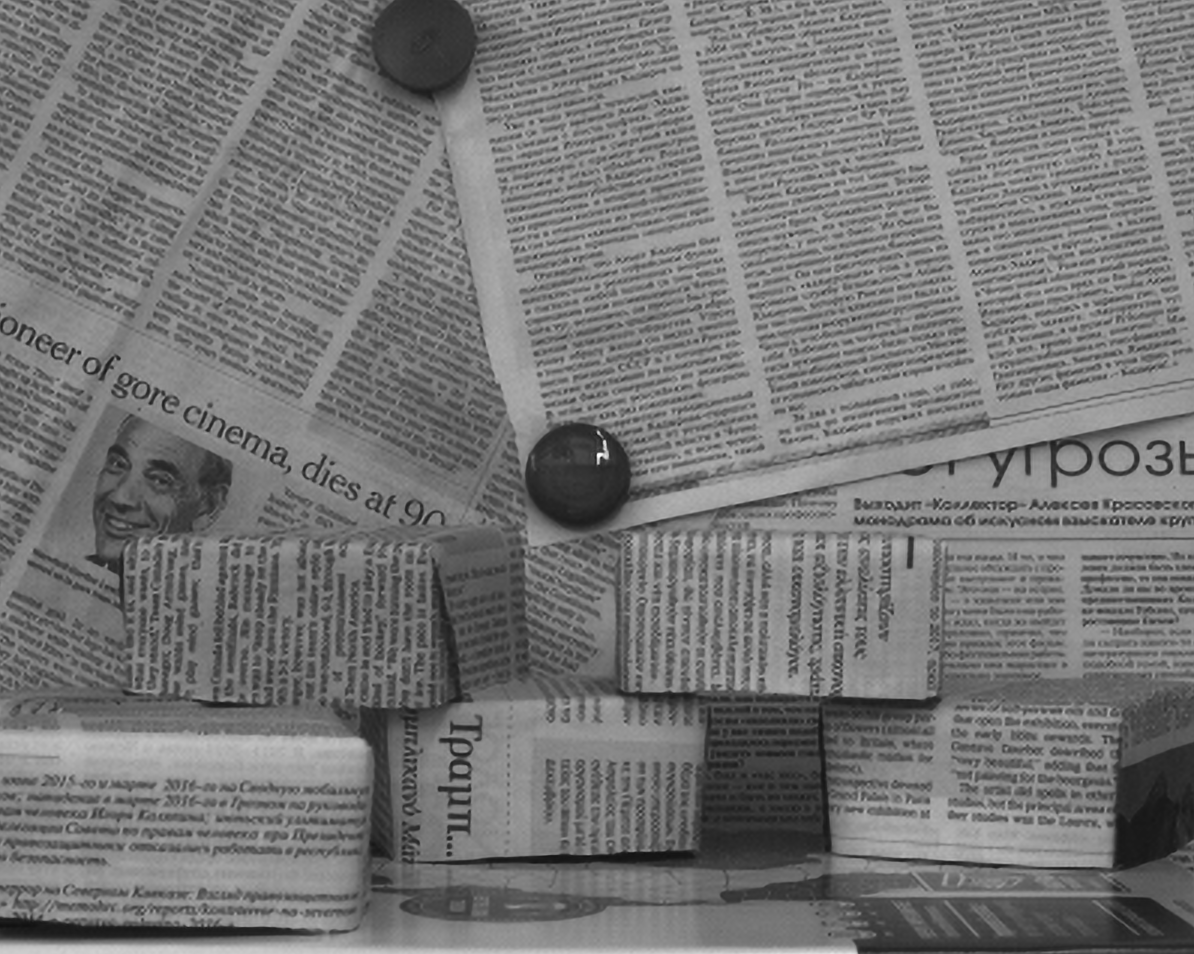}}; 
      \spy on (0.75, -0.1) in node [left] at (1.4, -1.9); 
    \end{tikzpicture}
	}\hspace{-1.2em}
	\subfloat{
		\begin{tikzpicture}[spy using outlines={rectangle,red,magnification=4.0,height=1.5cm, width=2.8cm, connect spies, every spy on node/.append style={thick}}] 
			\node {\pgfimage[width=0.16\linewidth]{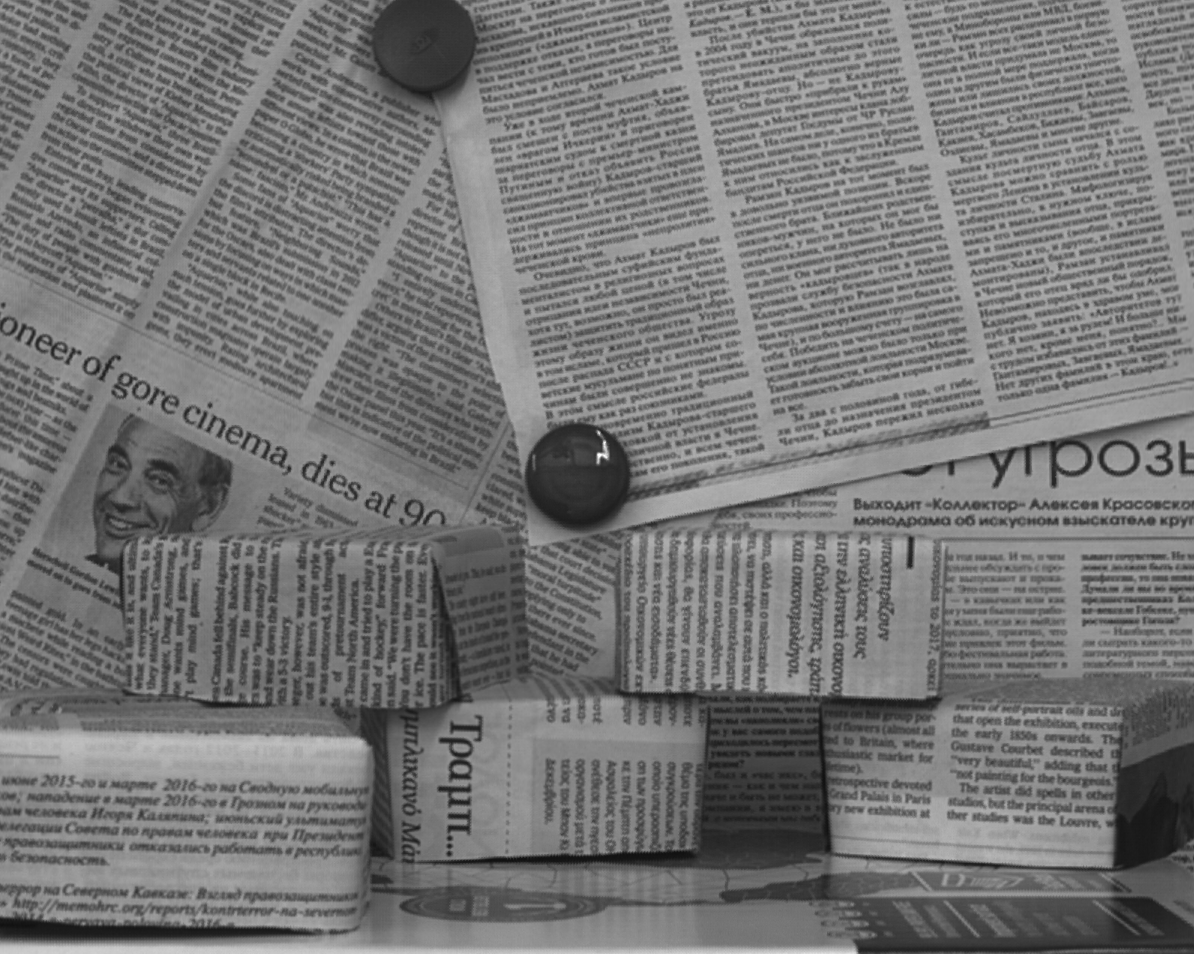}}; 
      \spy on (0.75, -0.1) in node [left] at (1.4, -1.9); 
    \end{tikzpicture}
	}\hspace{-1.2em}
	\subfloat{
		\begin{tikzpicture}[spy using outlines={rectangle,red,magnification=4.0,height=1.5cm, width=2.8cm, connect spies, every spy on node/.append style={thick}}] 
			\node {\pgfimage[width=0.16\linewidth]{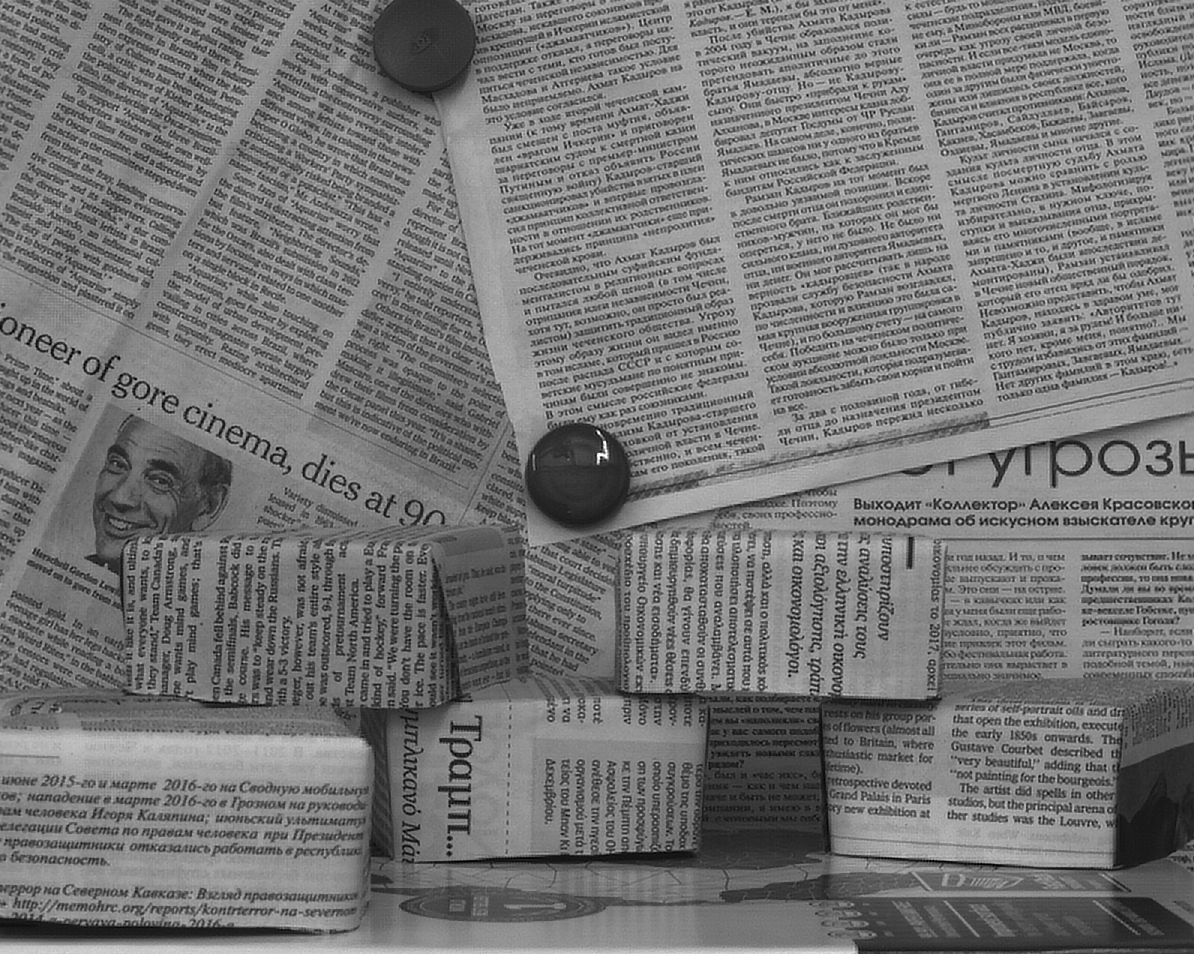}}; 
			\spy on (0.75, -0.1) in node [left] at (1.4, -1.9);  
    \end{tikzpicture}
	}\hspace{-1.2em}
	\subfloat{
		\begin{tikzpicture}[spy using outlines={rectangle,red,magnification=4.0,height=1.5cm, width=2.8cm, connect spies, every spy on node/.append style={thick}}] 
			\node {\pgfimage[width=0.16\linewidth]{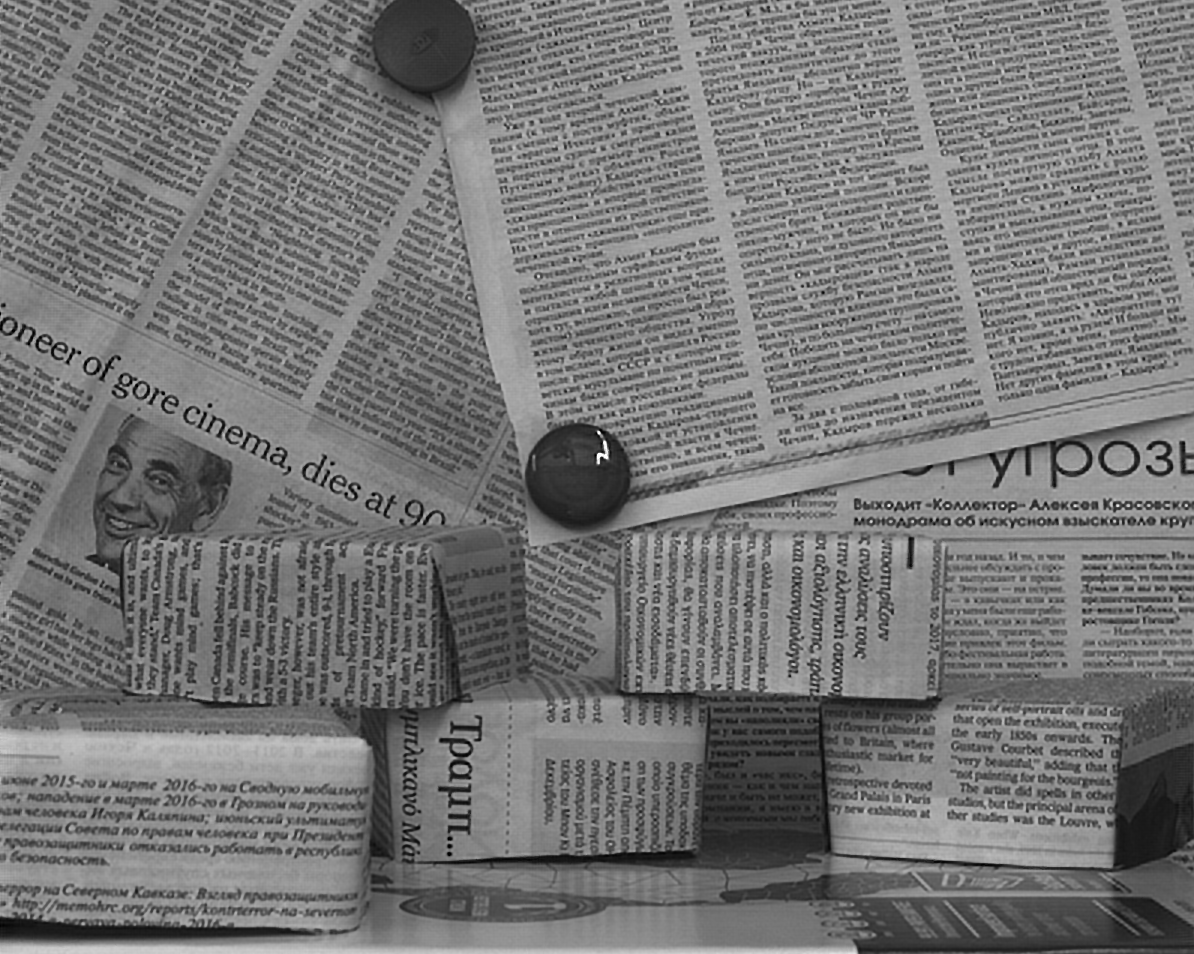}}; 
      \spy on (0.75, -0.1) in node [left] at (1.4, -1.9);  
    \end{tikzpicture}
	}\hspace{-1.2em}
	\subfloat{
		\begin{tikzpicture}[spy using outlines={rectangle,red,magnification=4.0,height=1.5cm, width=2.8cm, connect spies, every spy on node/.append style={thick}}] 
			\node {\pgfimage[width=0.16\linewidth]{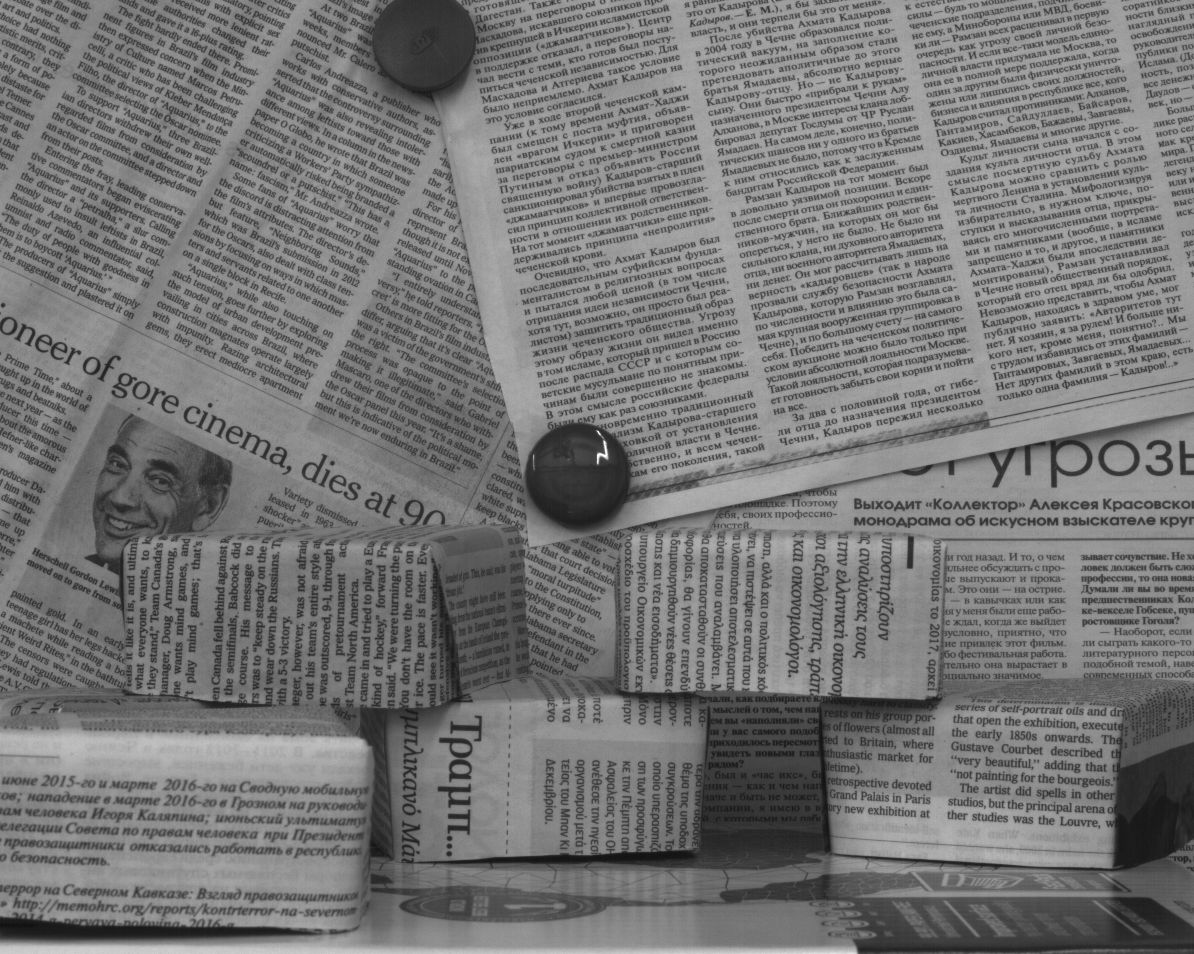}}; 
      \spy on (0.75, -0.1) in node [left] at (1.4, -1.9); 
    \end{tikzpicture}
	}}\\[-2.0ex]
	\setcounter{subfigure}{0}
	\mbox{
	\centering
	\subfloat[LR input]{
		\begin{tikzpicture}[spy using outlines={rectangle,red,magnification=4.0,height=1.5cm, width=2.8cm, connect spies, every spy on node/.append style={thick}}] 
			\node {\pgfimage[width=0.16\linewidth]{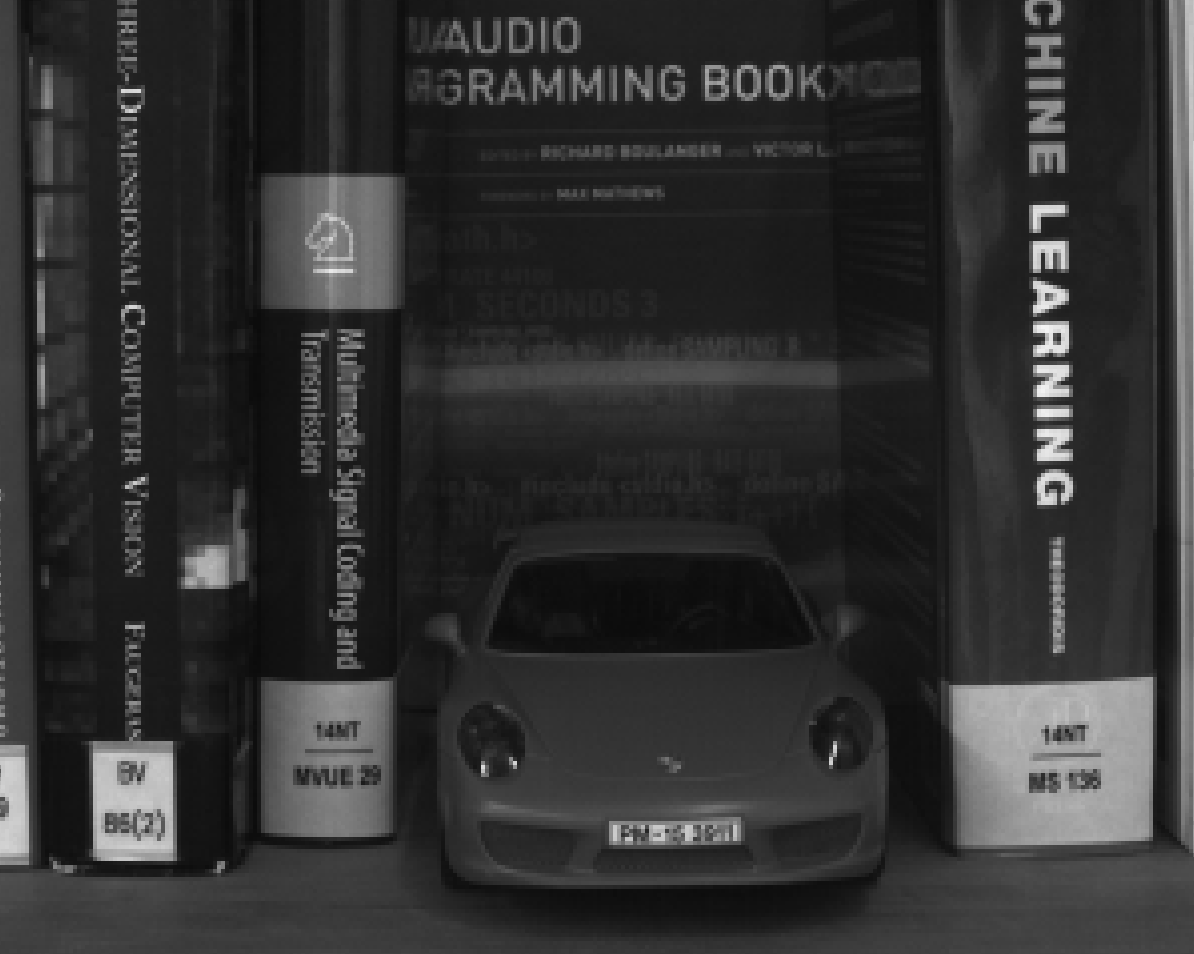}}; 
      \spy on (0.1, -0.75) in node [left] at (1.4, -1.9); 
    \end{tikzpicture}
	}\hspace{-1.2em}
	\subfloat[SRCNN \cite{Dong2014}]{
		\begin{tikzpicture}[spy using outlines={rectangle,red,magnification=4.0,height=1.5cm, width=2.8cm, connect spies, every spy on node/.append style={thick}}] 
			\node {\pgfimage[width=0.16\linewidth]{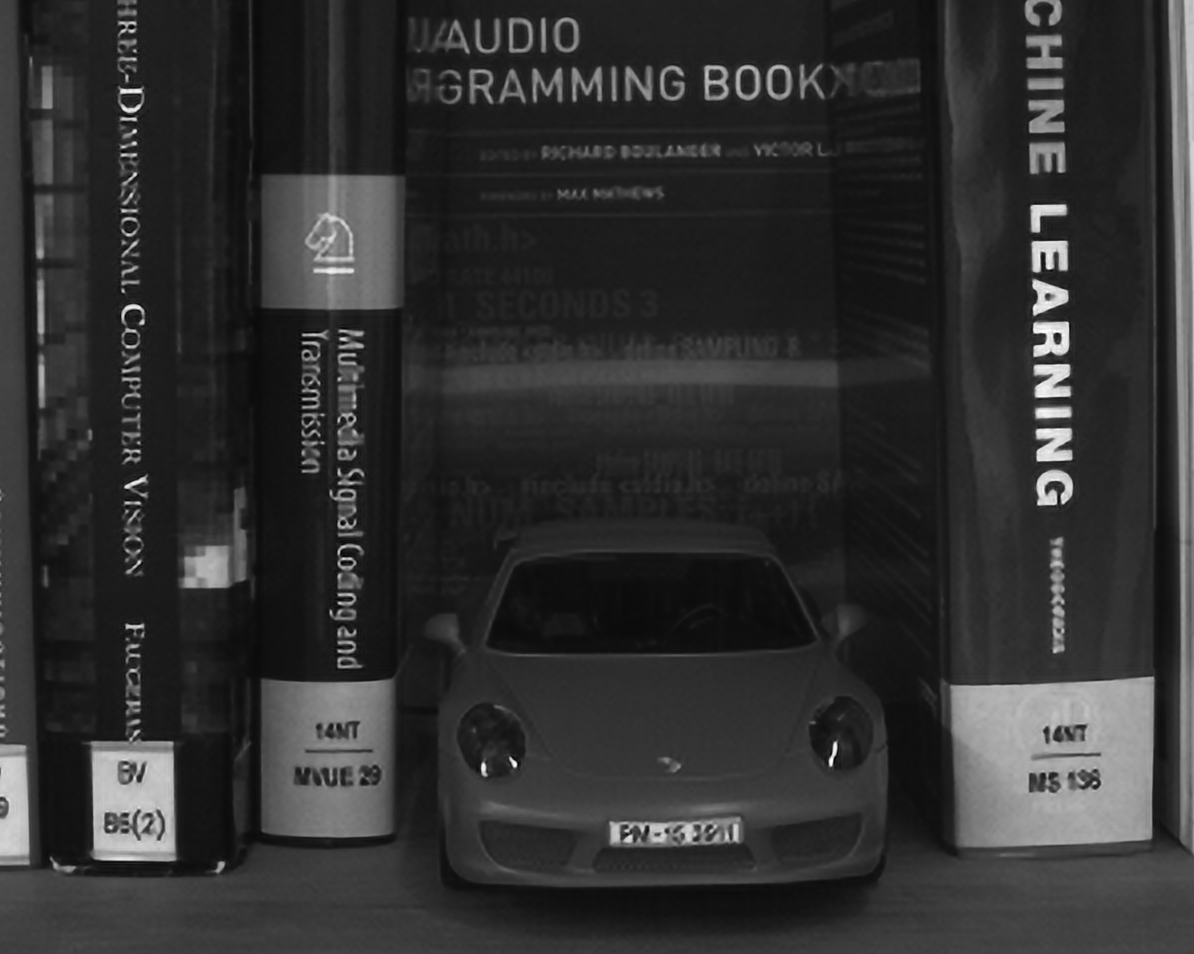}}; 
      \spy on (0.1, -0.75) in node [left] at (1.4, -1.9); 
    \end{tikzpicture}
	}\hspace{-1.2em}
	\subfloat[WNUISR \cite{Batz2016}]{
		\begin{tikzpicture}[spy using outlines={rectangle,red,magnification=4.0,height=1.5cm, width=2.8cm, connect spies, every spy on node/.append style={thick}}] 
			\node {\pgfimage[width=0.16\linewidth]{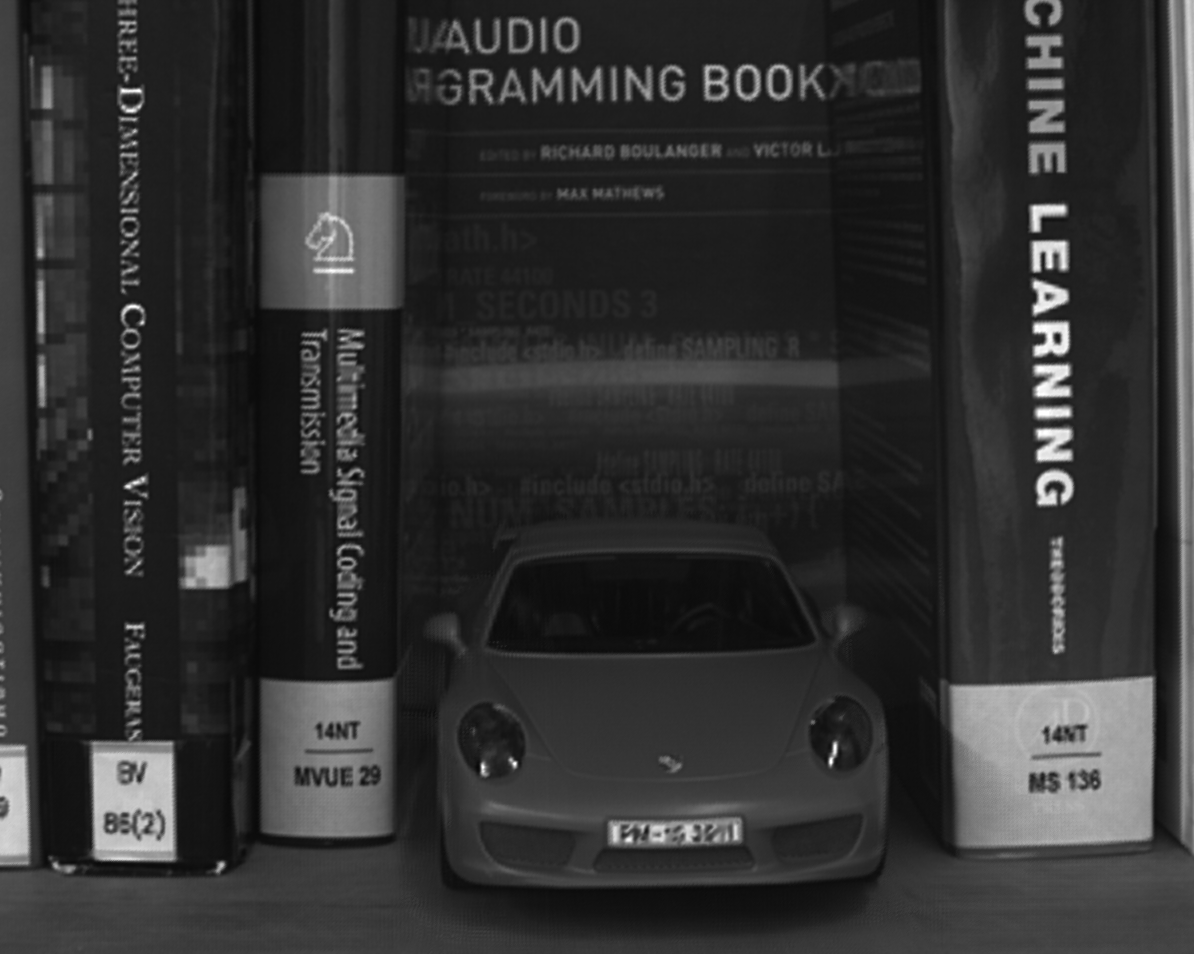}}; 
      \spy on (0.1, -0.75) in node [left] at (1.4, -1.9); 
    \end{tikzpicture}
	}\hspace{-1.2em}
	\subfloat[IRWSR \cite{Kohler2015c}]{
		\begin{tikzpicture}[spy using outlines={rectangle,red,magnification=4.0,height=1.5cm, width=2.8cm, connect spies, every spy on node/.append style={thick}}] 
			\node {\pgfimage[width=0.16\linewidth]{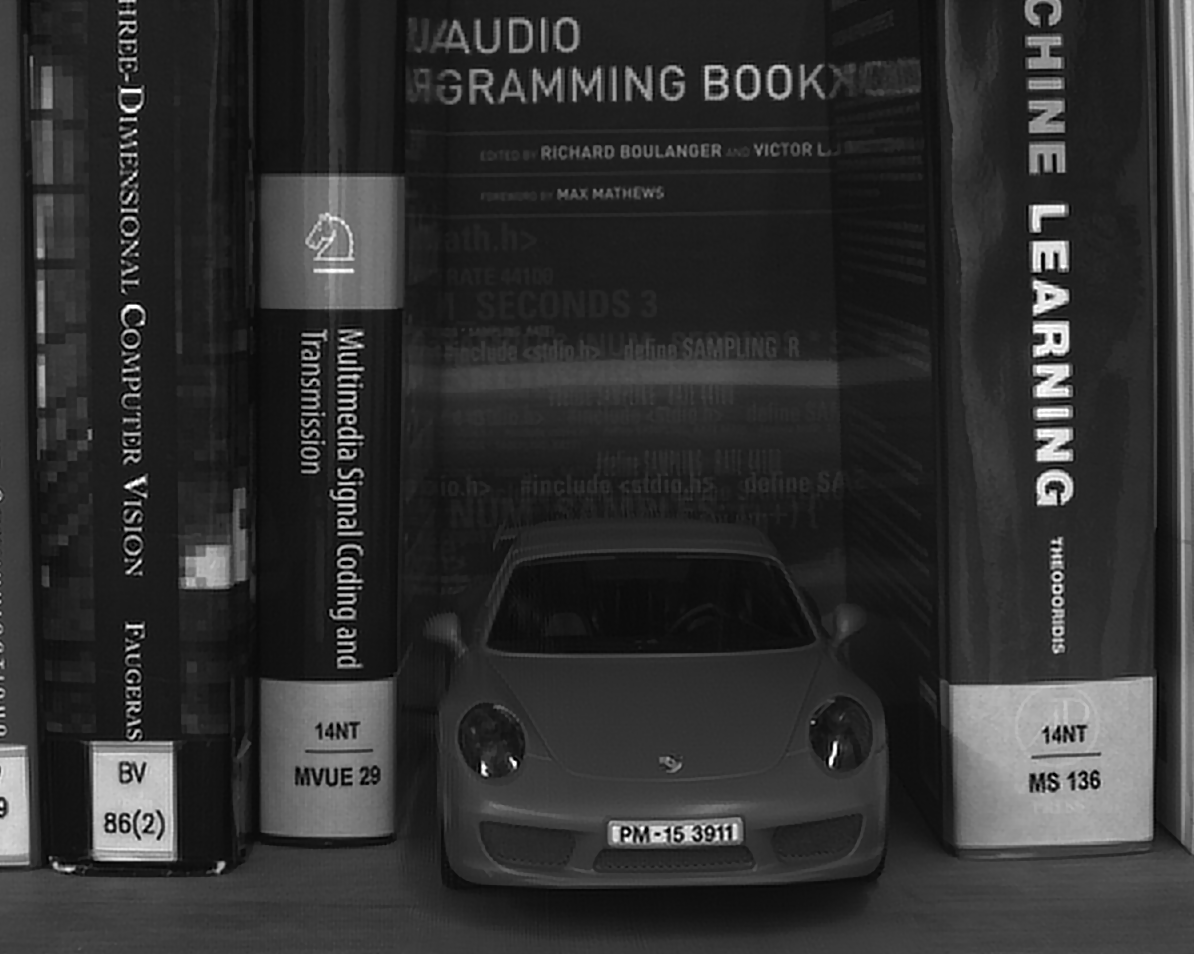}}; 
			\spy on (0.1, -0.75) in node [left] at (1.4, -1.9); 
    \end{tikzpicture}
	}\hspace{-1.2em}
	\subfloat[SRB \cite{Ma2015}]{
		\begin{tikzpicture}[spy using outlines={rectangle,red,magnification=4.0,height=1.5cm, width=2.8cm, connect spies, every spy on node/.append style={thick}}] 
			\node {\pgfimage[width=0.16\linewidth]{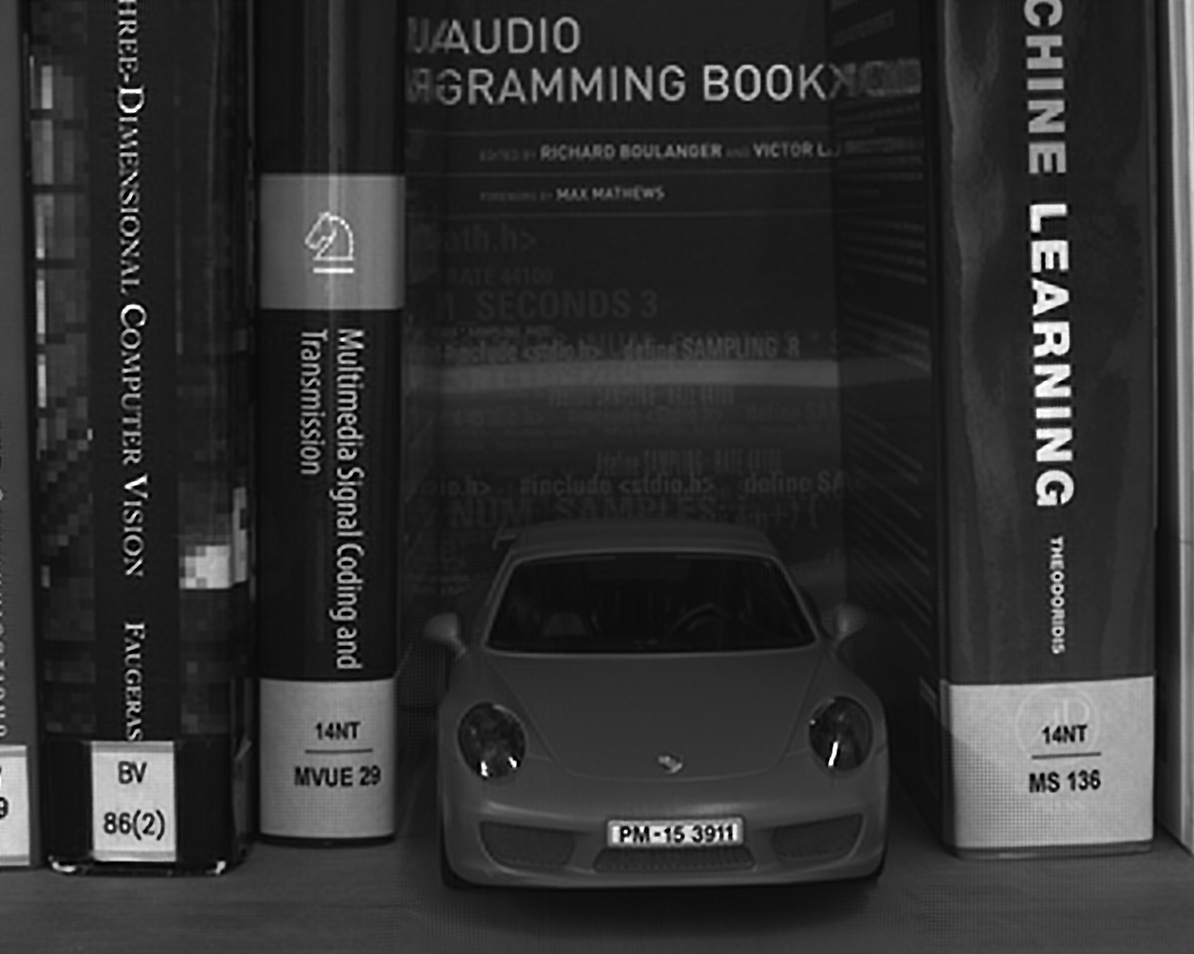}}; 
      \spy on (0.1, -0.75) in node [left] at (1.4, -1.9); 
    \end{tikzpicture}
	}\hspace{-1.2em}
	\subfloat[Ground truth]{
		\begin{tikzpicture}[spy using outlines={rectangle,red,magnification=4.0,height=1.5cm, width=2.8cm, connect spies, every spy on node/.append style={thick}}] 
			\node {\pgfimage[width=0.16\linewidth]{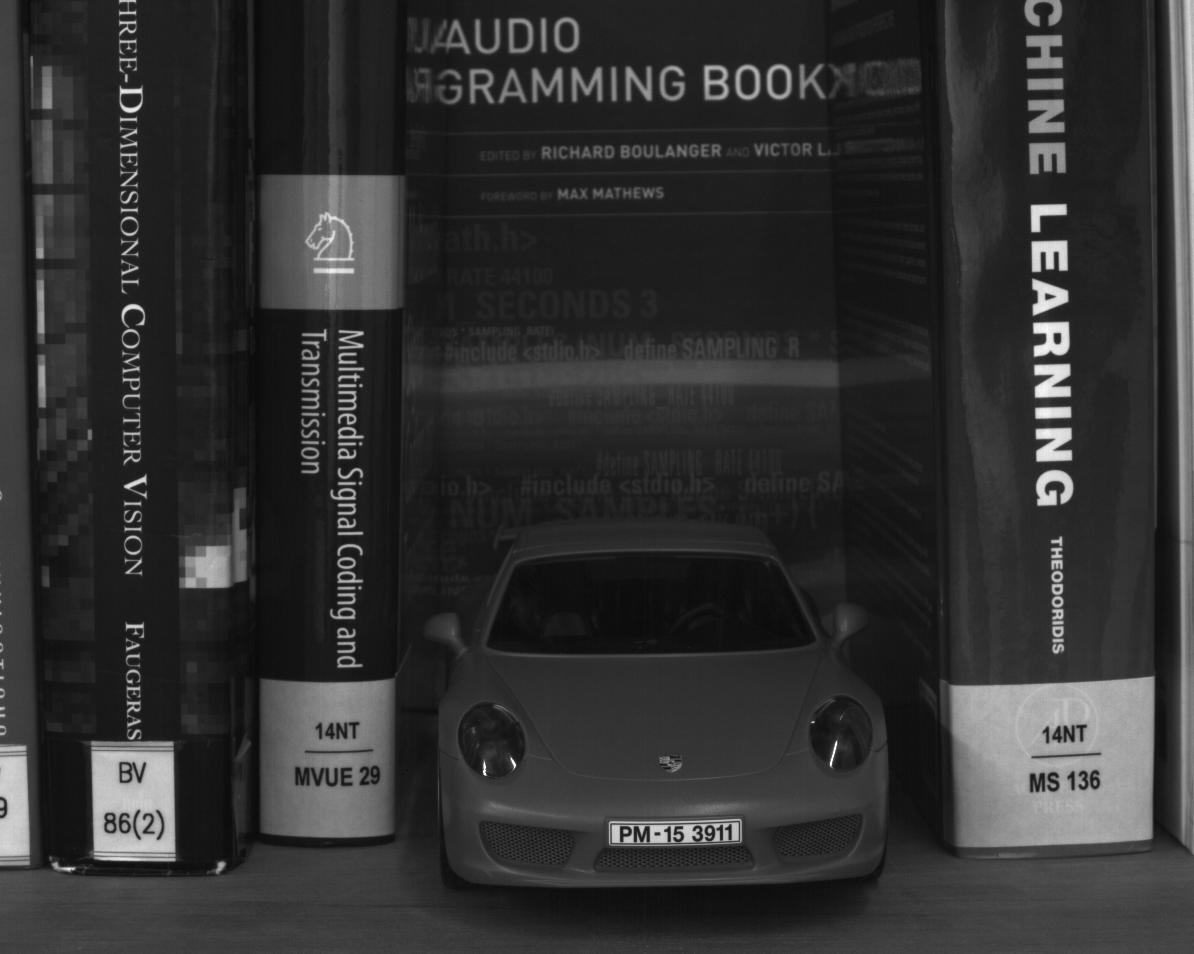}}; 
      \spy on (0.1, -0.75) in node [left] at (1.4, -1.9); 
    \end{tikzpicture}
	}
	}
	\caption{SR methods under global and mixed motion. Top row: \textit{newspapers} dataset with global motion ($3 \times$ magnification). Multi-frame SR (\eg, WNUISR, IRWSR, SRB) outperforms single-image SR (\eg, SRCNN) w.r.t. the recovery of fine structures like text. Bottom row: \textit{bookshelf} dataset with local motion of the vehicle car movements ($4 \times$ magnification). The interpolation-based approach (WNUISR) is prone to inaccurate motion estimation due to local motion while the single-image method (SRCNN) is robust to local motion.}
	\label{fig:globalAndLocalMotionExample}
\end{figure*}

We used the reference implementations provided by the corresponding authors if available. For L1BTV and BEPSR, we used the publicly available MATLAB SR toolbox \cite{Kohler2015c}. For NUISR, we adopted the method in \cite{Batz2015}. To evaluate the learning-based methods, we used their original pretrained models wherever possible. In case of NBSRF and SCSR, the models were retrained for $3\times$ and $4\times$ magnification on the original training data as pretrained models were unavailable. For VSRNET, we used the network that was trained by the authors for $K = 5$ frames for all magnifications. We selected free parameters following the guidelines in the cited papers or the available source codes. For all methods that require prior knowledge on the camera PSF, consistent parametrizations were used. We used an isotropic Gaussian kernel of size $\lceil 6 \sigma_{\text{PSF}} \rceil \times \lceil 6 \sigma_{\text{PSF}} \rceil$ pixels to model the PSF, where $\sigma_{\text{PSF}} = b \sigma_0$ denotes the standard deviation on the HR grid, $b$ is the desired magnification, and $\sigma_0 = 0.4$ is the standard deviation on the LR grid.

\section{Experiments and Results}
\label{sec:ExperimentsAndResults}

\noindent
\textbf{Static scenes.}
\Fref{fig:srBenchmarkMotionTypes:global} benchmarks the SR methods with different magnification factors on our global motion datasets. Note that the normalized quality measures relative to LR data tend to increase with the magnification, while PSNR has a maximum for medium factors (3). The performance of the algorithms relative to each other depends on the magnification as well as the utilized quality measure.

Regarding the performance for a fixed magnification factor, the different
measures are inconsistent. Except for large magnifications, the
interpolation-based methods (NUISR, HYSR) performed best in terms of the PSNR.
In case of the IFC, reconstruction-based SR (BEPSR, IRWSR) achieved better
results, especially for large magnifications. This can be explained by the
characteristics of the measures as well as
algorithm-specific properties. There are two main observations. 1) The PSNR
weighs deviations to the ground truth in homogeneous and textured regions
uniformly. We observed that the PSNR tends to prefer slightly oversmoothed
images, which is consistent with evaluations of full-reference quality
assessment \cite{Sheikh2006}. As interpolation-based SR tends to introduce
blur, especially for large magnifications, these methods are ranked higher by
the PSNR. 2) IFC puts the emphasis on high-frequency components
\cite{Yang2014a}. Reconstruction-based SR use statistical priors on
natural images, \eg, sparsity~\cite{Kohler2015c, Farsiu2004a}, which leads to a
better recovery of high frequencies and thus a higher IFC score.
Interestingly, blind SR (SRB) did not perform better than the computationally
more efficient non-blind methods. For short sequences at small magnifications,
SRB was prone to ringing artifacts, resulting in negative normalized measures.
\Fref{fig:globalAndLocalMotionExample} shows a comparison among different
methodologies. In the top, the sparsity priors contributed to the recovery of
the printed text.

\begin{figure}[!t]
	\centering
	\mbox{
	\hspace{-1.0em}
	\subfloat[SRB \cite{Ma2015}]{
		\begin{tikzpicture}[spy using outlines={rectangle,red,magnification=4.0,height=1.5cm, width=2.7cm, connect spies, every spy on node/.append style={thick}}] 
			\node {\pgfimage[width=0.329\linewidth]{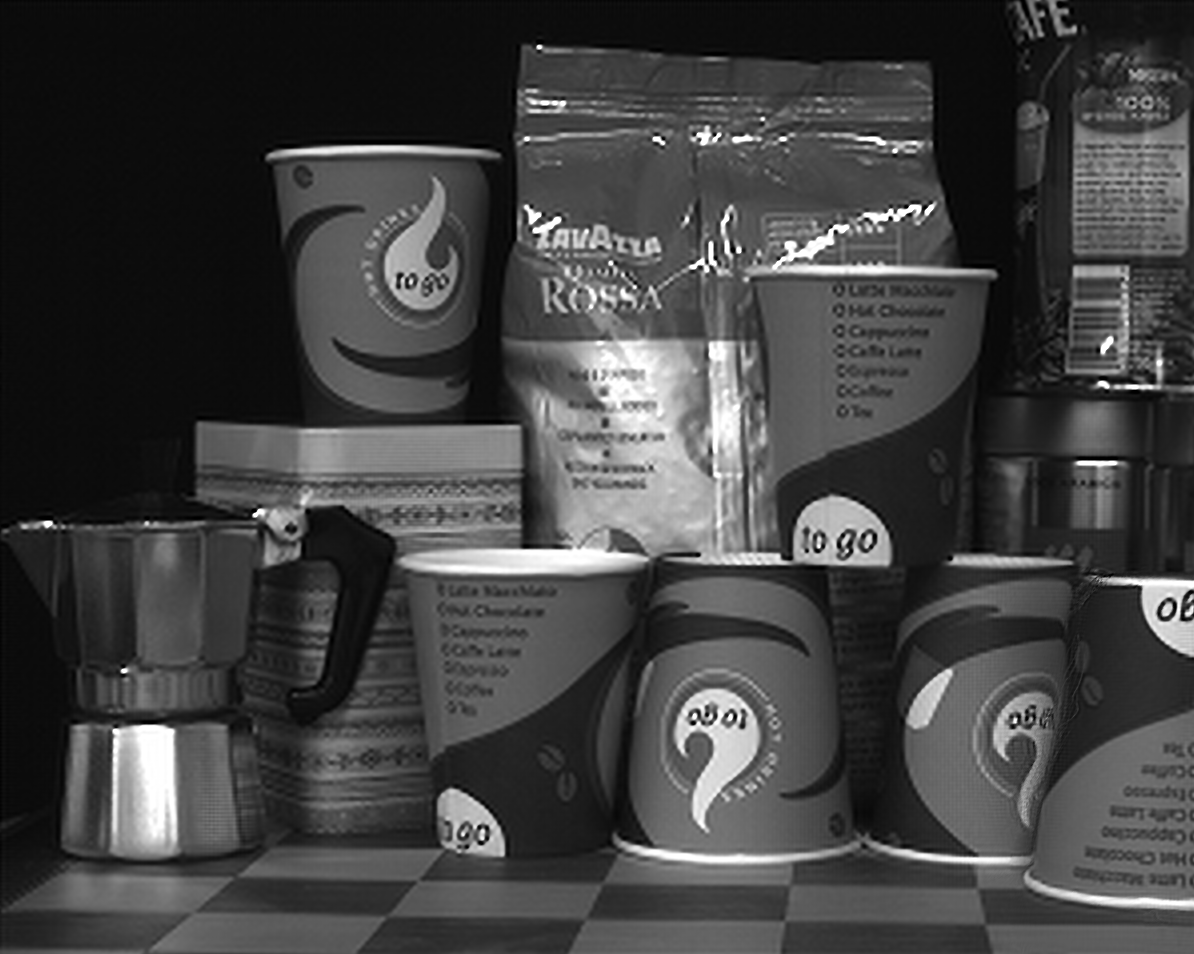}}; 
      \spy on (0.75, -0.1) in node [left] at (1.35, -1.88); 
    \end{tikzpicture}
	}\hspace{-1.2em}
	\subfloat[VSRNET \cite{Kappeler2016}]{
		\begin{tikzpicture}[spy using outlines={rectangle,red,magnification=4.0,height=1.5cm, width=2.7cm, connect spies, every spy on node/.append style={thick}}] 
			\node {\pgfimage[width=0.329\linewidth]{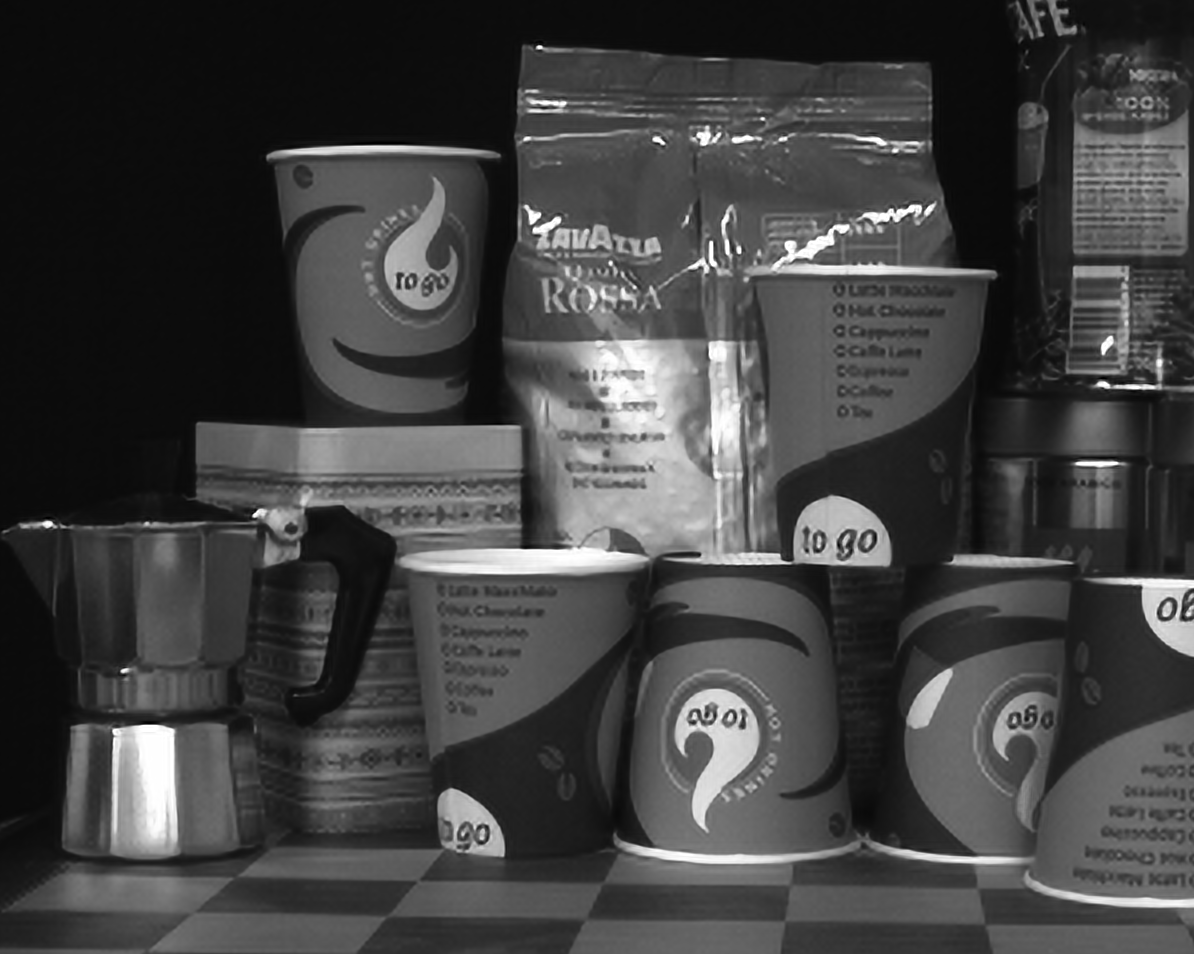}}; 
      \spy on (0.75, -0.1) in node [left] at (1.35, -1.88); 
    \end{tikzpicture}
	}\hspace{-1.2em}
	\subfloat[Ground truth]{
		\begin{tikzpicture}[spy using outlines={rectangle,red,magnification=4.0,height=1.5cm, width=2.7cm, connect spies, every spy on node/.append style={thick}}] 
			\node {\pgfimage[width=0.329\linewidth]{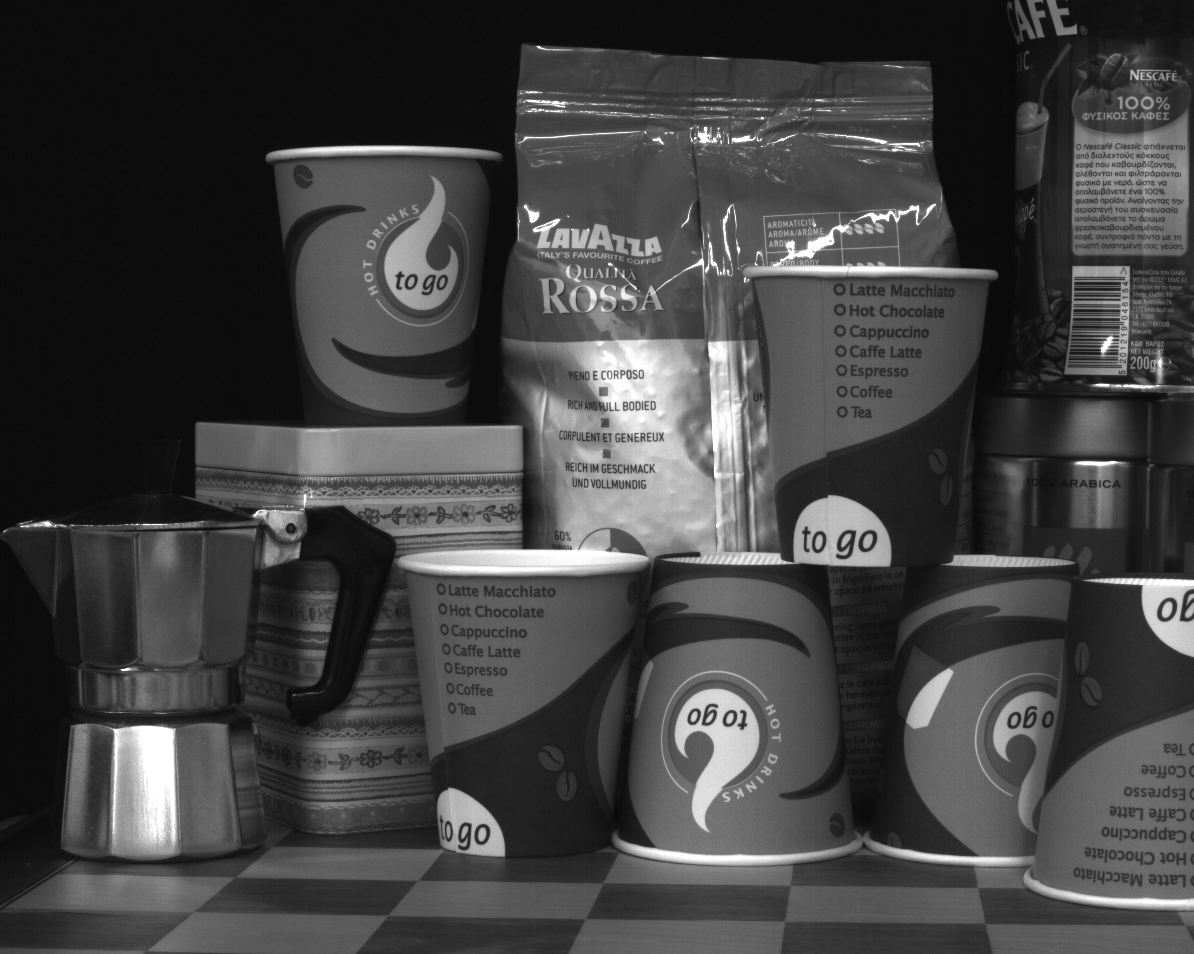}}; 
      \spy on (0.75, -0.1) in node [left] at (1.35, -1.88); 
    \end{tikzpicture}
	}}
	\caption{Multi-frame SR under local motion and the absence of global camera motion on the \textit{coffee} dataset ($4\times$ magnification).}
	\label{fig:staticBackgroundExample}
\end{figure}

Regarding the behavior across different magnifications, MFSR tends to outperform SISR,
especially in terms of the IFC. This is because MFSR exploits complementary
information across multiple images to recover HR details, while SISR can only
"hallucinate" such details. In SISR, it is worth noting that methods that use
external data (NBSRF, SRCNN) quantitatively outperformed the self-exemplar
approach (SESR). In the field of MFSR, interpolation-based algorithms were
suitable for small magnification (2) while reconstruction and deep
learning approaches (VSRNET) performed better for larger factors ($\geq 3$). We
explain this behavior by the use of statistical priors in reconstruction-based
SR, which guides the recovery of fine structures. Similarly, VSRNET learns such
a prior implicitly from examples.
\\[0.8ex]
\textbf{Dynamic scenes.}
\Fref{fig:srBenchmarkMotionTypes:local} benchmarks the competing methods on our
mixed motion data. This shows that the performance of most MFSR
algorithms considerably deteriorated compared to static scenes, which partly
resulted in negative normalized quality measures. Unlike MFSR, SISR algorithms
were obviously unaffected. In general, the impact of local motion was more
significant for more input frames at larger magnification factors. That is
because motion estimation (typically done via optical flow) becomes more difficult for large displacements
related to local motion over longer input sequences.
We found that algorithms
building on simple interpolation (NUISR, HYSR) were most sensitive.
Interpolation-based SR with proper outlier weighting (WNUISR) or refinement
(DBRSR) as well as reconstruction-based SR with outlier-insensitive models
showed higher robustness. Interestingly, VSRNET was only slightly affected by
local motion. We explain this observation by the neural network architecture
that was trained for a fixed number of input frames and the underlying adaptive motion
compensation scheme. \Fref{fig:globalAndLocalMotionExample} (bottom) depicts
some representative methods on an emulated surveillance scene, where local
motion is related to movements of a car.

\Fref{fig:srBenchmarkMotionTypes:staticBackground} depicts our benchmark under pure local motion. Note that the absence of global motion inherently affected MFSR as complementary information across LR frames does not exist. Thus, these algorithms effectively perform multi-frame deblurring/denoising but do not directly address undersampling. In our benchmark, SISR partly outperformed MFSR. Among the MFSR algorithms, VSRNET performed best. This can be explained by the external training data used for VSRNET. We found that in the absence of global motion this approach drops back to SISR and better recovers discontinuities and fine image details, see \fref{fig:staticBackgroundExample}.
\\[0.8ex]
\textbf{Photometric variations.}
We also studied SR under photometric variations over the input frames. This situation appears if input frames are collected over a longer period of time with environmental changes, \eg in remote sensing. An exact handling requires photometric registration schemes \cite{Capel2003}, which is omitted by most state-of-the-art algorithms. %We analyzed the impact of this issue on our photometric variation dataset under global motion.

\begin{figure}[!t]
	\scriptsize 
	\centering
	\setlength \figurewidth{0.185\textwidth}
	\setlength \figureheight{0.80\figurewidth}
	\subfloat{\includegraphics[width=0.45\textwidth]{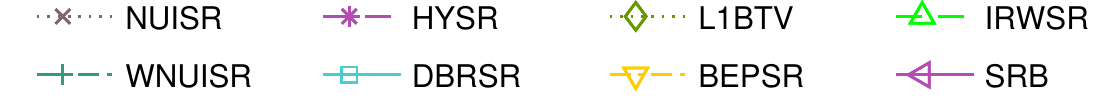}}\\[-1.5ex]
	\subfloat{% This file was created by matlab2tikz.
%
\definecolor{mycolor1}{rgb}{0.50000,0.40000,0.40000}%
\definecolor{mycolor2}{rgb}{0.20000,0.60000,0.50000}%
\definecolor{mycolor3}{rgb}{0.70000,0.30000,0.70000}%
\definecolor{mycolor4}{rgb}{0.30000,0.80000,0.80000}%
\definecolor{mycolor5}{rgb}{1.00000,0.80000,0.00000}%
\begin{tikzpicture}

\begin{axis}[%
width=0.951\figurewidth,
height=\figureheight,
at={(0\figurewidth,0\figureheight)},
scale only axis,
xmin=0,
xmax=5,
xlabel={Number of outlier frames},
ymin=-0.4,
ymax=0.2,
ylabel={Norm. PSNR},
axis background/.style={fill=white},
xmajorgrids,
ymajorgrids,
xlabel near ticks,ylabel near ticks,scaled y ticks=false,yticklabel style={/pgf/number format/fixed, /pgf/number format/precision=2},
]
\addplot [color=mycolor1,dotted,line width=0.8pt,mark=x,mark options={solid},forget plot]
  table[row sep=crcr]{%
0	0.15109423690482\\
1	-0.0399393606841242\\
2	-0.106475078129163\\
3	-0.155387440099399\\
4	-0.200448287827326\\
5	-0.241216154475588\\
};
\addplot [color=mycolor2,dashdotted,line width=0.8pt,mark=+,mark options={solid},forget plot]
  table[row sep=crcr]{%
0	0.133286855287587\\
1	0.089035341975155\\
2	0.0299460014939986\\
3	-0.0307925011041782\\
4	-0.0868237111109418\\
5	-0.128565852109317\\
};
\addplot [color=mycolor3,dashed,line width=0.8pt,mark=asterisk,mark options={solid},forget plot]
  table[row sep=crcr]{%
0	0.153103748072748\\
1	-0.00151587582282846\\
2	-0.0591950554327084\\
3	-0.101822175643497\\
4	-0.14167440598481\\
5	-0.176923616623051\\
};
\addplot [color=mycolor4,line width=0.8pt,mark=square,mark options={solid},forget plot]
  table[row sep=crcr]{%
0	0.137967470029238\\
1	0.0926225939789484\\
2	0.0359876968633064\\
3	-0.0238411427186085\\
4	-0.0781802056979427\\
5	-0.118500927625139\\
};
\addplot [color=red!40!green,dotted,line width=0.8pt,mark=diamond,mark options={solid},forget plot]
  table[row sep=crcr]{%
0	0.127516679822097\\
1	0.124363776062917\\
2	0.11785195465732\\
3	0.106313429312329\\
4	0.0841058313143529\\
5	0.0292610752107396\\
};
\addplot [color=mycolor5,dashdotted,line width=0.8pt,mark=triangle,mark options={solid,rotate=180},forget plot]
  table[row sep=crcr]{%
0	0.145574307098237\\
1	0.118506853019624\\
2	0.121382658133101\\
3	0.122959831274123\\
4	0.122029482110344\\
5	0.124009051912793\\
};
\addplot [color=green,dashed,line width=0.8pt,mark=triangle,mark options={solid},forget plot]
  table[row sep=crcr]{%
0	0.0986688969619943\\
1	0.101405464445265\\
2	0.100190552309769\\
3	0.1039147381751\\
4	0.10263366330035\\
5	0.101820365976338\\
};
\addplot [color=mycolor3,line width=0.8pt,mark=triangle,mark options={solid,rotate=90},forget plot]
  table[row sep=crcr]{%
0	0.0117839952054577\\
1	-0.183910932312258\\
2	-0.173359735009984\\
3	-0.201058363011996\\
4	-0.275877800267851\\
5	-0.312568229850198\\
};
\end{axis}
\end{tikzpicture}%
}~
	\subfloat{% This file was created by matlab2tikz.
%
\definecolor{mycolor1}{rgb}{0.50000,0.40000,0.40000}%
\definecolor{mycolor2}{rgb}{0.20000,0.60000,0.50000}%
\definecolor{mycolor3}{rgb}{0.70000,0.30000,0.70000}%
\definecolor{mycolor4}{rgb}{0.30000,0.80000,0.80000}%
\definecolor{mycolor5}{rgb}{1.00000,0.80000,0.00000}%
\begin{tikzpicture}

\begin{axis}[%
width=0.951\figurewidth,
height=\figureheight,
at={(0\figurewidth,0\figureheight)},
scale only axis,
xmin=0,
xmax=5,
xlabel={Number of outlier frames},
ymin=-0.6,
ymax=0.2,
ylabel={Norm. SSIM},
axis background/.style={fill=white},
xmajorgrids,
ymajorgrids,
xlabel near ticks,ylabel near ticks,scaled y ticks=false,yticklabel style={/pgf/number format/fixed, /pgf/number format/precision=2},
]
\addplot [color=mycolor1,dotted,line width=0.8pt,mark=x,mark options={solid},forget plot]
  table[row sep=crcr]{%
0	0.0938413352046996\\
1	-0.124767079040191\\
2	-0.21709377906444\\
3	-0.279207220545889\\
4	-0.332491738926177\\
5	-0.380883809879642\\
};
\addplot [color=mycolor2,dashdotted,line width=0.8pt,mark=+,mark options={solid},forget plot]
  table[row sep=crcr]{%
0	0.0883196257731889\\
1	0.0403145081701623\\
2	-0.0174493195855733\\
3	-0.0876468382865683\\
4	-0.157852792456236\\
5	-0.219375381223034\\
};
\addplot [color=mycolor3,dashed,line width=0.8pt,mark=asterisk,mark options={solid},forget plot]
  table[row sep=crcr]{%
0	0.0946002292907898\\
1	-0.0807290006982337\\
2	-0.155228517672191\\
3	-0.203747224320547\\
4	-0.243497132078369\\
5	-0.276796075721848\\
};
\addplot [color=mycolor4,line width=0.8pt,mark=square,mark options={solid},forget plot]
  table[row sep=crcr]{%
0	0.0917506203862049\\
1	0.0447366878990766\\
2	-0.0087069588230146\\
3	-0.0765917040102366\\
4	-0.143811867869745\\
5	-0.20281892851951\\
};
\addplot [color=red!40!green,dotted,line width=0.8pt,mark=diamond,mark options={solid},forget plot]
  table[row sep=crcr]{%
0	0.085301576773699\\
1	0.0824795176411761\\
2	0.0778824643235152\\
3	0.0692569135197032\\
4	0.0571888013939342\\
5	0.0271645029823364\\
};
\addplot [color=mycolor5,dashdotted,line width=0.8pt,mark=triangle,mark options={solid,rotate=180},forget plot]
  table[row sep=crcr]{%
0	0.10581471715789\\
1	0.101420246391084\\
2	0.101433408763723\\
3	0.100114222537017\\
4	0.0986741799687535\\
5	0.096045430552462\\
};
\addplot [color=green,dashed,line width=0.8pt,mark=triangle,mark options={solid},forget plot]
  table[row sep=crcr]{%
0	0.087562029661523\\
1	0.0900180555291024\\
2	0.0862628492255005\\
3	0.0874215189980136\\
4	0.0843679317649318\\
5	0.0815646509332461\\
};
\addplot [color=mycolor3,line width=0.8pt,mark=triangle,mark options={solid,rotate=90},forget plot]
  table[row sep=crcr]{%
0	0.0548603678153369\\
1	-0.382536889976532\\
2	-0.344351011384681\\
3	-0.389476478726961\\
4	-0.495409837856242\\
5	-0.534045060719225\\
};
\end{axis}
\end{tikzpicture}%
}\\
	\subfloat{% This file was created by matlab2tikz.
%
\definecolor{mycolor1}{rgb}{0.50000,0.40000,0.40000}%
\definecolor{mycolor2}{rgb}{0.20000,0.60000,0.50000}%
\definecolor{mycolor3}{rgb}{0.70000,0.30000,0.70000}%
\definecolor{mycolor4}{rgb}{0.30000,0.80000,0.80000}%
\definecolor{mycolor5}{rgb}{1.00000,0.80000,0.00000}%
\begin{tikzpicture}

\begin{axis}[%
width=0.951\figurewidth,
height=\figureheight,
at={(0\figurewidth,0\figureheight)},
scale only axis,
xmin=0,
xmax=5,
xlabel={Number of outlier frames},
ymin=-0.18,
ymax=0.02,
ylabel={Norm. MS-SSIM},
axis background/.style={fill=white},
xmajorgrids,
ymajorgrids,
xlabel near ticks,ylabel near ticks,scaled y ticks=false,yticklabel style={/pgf/number format/fixed, /pgf/number format/precision=2},
]
\addplot [color=mycolor1,dotted,line width=0.8pt,mark=x,mark options={solid},forget plot]
  table[row sep=crcr]{%
0	0.0199789764111893\\
1	-0.0184435658046907\\
2	-0.0401759367925041\\
3	-0.0583577788638119\\
4	-0.0752348854801629\\
5	-0.0927599984464531\\
};
\addplot [color=mycolor2,dashdotted,line width=0.8pt,mark=+,mark options={solid},forget plot]
  table[row sep=crcr]{%
0	0.0189746767026099\\
1	0.0117568777133504\\
2	0.000882192209876161\\
3	-0.0140830200382798\\
4	-0.0322611731883631\\
5	-0.0494617287773426\\
};
\addplot [color=mycolor3,dashed,line width=0.8pt,mark=asterisk,mark options={solid},forget plot]
  table[row sep=crcr]{%
0	0.0199174033531306\\
1	-0.00855237704328308\\
2	-0.0241319892654042\\
3	-0.0367065689288864\\
4	-0.04781132193091\\
5	-0.0581119803816443\\
};
\addplot [color=mycolor4,line width=0.8pt,mark=square,mark options={solid},forget plot]
  table[row sep=crcr]{%
0	0.0193061649855758\\
1	0.0123287011672139\\
2	0.00195626714022028\\
3	-0.0126786704234382\\
4	-0.0303884450236226\\
5	-0.0471578403905295\\
};
\addplot [color=red!40!green,dotted,line width=0.8pt,mark=diamond,mark options={solid},forget plot]
  table[row sep=crcr]{%
0	0.0160299617822544\\
1	0.0155213919381734\\
2	0.0148350603982797\\
3	0.0134248018759433\\
4	0.0111813756006443\\
5	0.00407125266966686\\
};
\addplot [color=mycolor5,dashdotted,line width=0.8pt,mark=triangle,mark options={solid,rotate=180},forget plot]
  table[row sep=crcr]{%
0	0.0195125574374503\\
1	0.018579865686036\\
2	0.0185708824829842\\
3	0.0183501990836721\\
4	0.0181350958707723\\
5	0.0177481582897237\\
};
\addplot [color=green,dashed,line width=0.8pt,mark=triangle,mark options={solid},forget plot]
  table[row sep=crcr]{%
0	0.016477247589538\\
1	0.0166356246507973\\
2	0.0160844013593261\\
3	0.0161848737200282\\
4	0.0155764397363479\\
5	0.0149696736265714\\
};
\addplot [color=mycolor3,line width=0.8pt,mark=triangle,mark options={solid,rotate=90},forget plot]
  table[row sep=crcr]{%
0	0.000376007042515938\\
1	-0.09229308505683\\
2	-0.0794155098093345\\
3	-0.0880508483468486\\
4	-0.148020986656238\\
5	-0.163355538298734\\
};
\end{axis}
\end{tikzpicture}%
}~
	\subfloat{% This file was created by matlab2tikz.
%
\definecolor{mycolor1}{rgb}{0.50000,0.40000,0.40000}%
\definecolor{mycolor2}{rgb}{0.20000,0.60000,0.50000}%
\definecolor{mycolor3}{rgb}{0.70000,0.30000,0.70000}%
\definecolor{mycolor4}{rgb}{0.30000,0.80000,0.80000}%
\definecolor{mycolor5}{rgb}{1.00000,0.80000,0.00000}%
\begin{tikzpicture}

\begin{axis}[%
width=0.951\figurewidth,
height=\figureheight,
at={(0\figurewidth,0\figureheight)},
scale only axis,
xmin=0,
xmax=5,
xlabel={Number of outlier frames},
ymin=-0.4,
ymax=0.8,
ylabel={Norm. IFC},
axis background/.style={fill=white},
xmajorgrids,
ymajorgrids,
xlabel near ticks,ylabel near ticks,scaled y ticks=false,yticklabel style={/pgf/number format/fixed, /pgf/number format/precision=2},
]
\addplot [color=mycolor1,dotted,line width=0.8pt,mark=x,mark options={solid},forget plot]
  table[row sep=crcr]{%
0	0.680611504720233\\
1	0.206665663927983\\
2	0.0557481560181852\\
3	-0.0363415638691471\\
4	-0.123100486957702\\
5	-0.206281328808001\\
};
\addplot [color=mycolor2,dashdotted,line width=0.8pt,mark=+,mark options={solid},forget plot]
  table[row sep=crcr]{%
0	0.568273405581151\\
1	0.419070223683882\\
2	0.262904062548548\\
3	0.125549049849619\\
4	-0.0117394365191173\\
5	-0.119216816686432\\
};
\addplot [color=mycolor3,dashed,line width=0.8pt,mark=asterisk,mark options={solid},forget plot]
  table[row sep=crcr]{%
0	0.680715476429996\\
1	0.259963673764173\\
2	0.122732368061377\\
3	0.0421564597592284\\
4	-0.0312946435335718\\
5	-0.101604116343369\\
};
\addplot [color=mycolor4,line width=0.8pt,mark=square,mark options={solid},forget plot]
  table[row sep=crcr]{%
0	0.56888752569857\\
1	0.421381047966027\\
2	0.256451329049681\\
3	0.110993354082728\\
4	-0.0365279199373962\\
5	-0.148503082713915\\
};
\addplot [color=red!40!green,dotted,line width=0.8pt,mark=diamond,mark options={solid},forget plot]
  table[row sep=crcr]{%
0	0.573824130840248\\
1	0.552582576996278\\
2	0.510559896876152\\
3	0.450289000076777\\
4	0.360761298527627\\
5	0.168929251929548\\
};
\addplot [color=mycolor5,dashdotted,line width=0.8pt,mark=triangle,mark options={solid,rotate=180},forget plot]
  table[row sep=crcr]{%
0	0.719861889344969\\
1	0.710956993506556\\
2	0.691707908319797\\
3	0.66793932673302\\
4	0.64047958544157\\
5	0.598059828066008\\
};
\addplot [color=green,dashed,line width=0.8pt,mark=triangle,mark options={solid},forget plot]
  table[row sep=crcr]{%
0	0.685964680010534\\
1	0.689861566932909\\
2	0.664752904146582\\
3	0.646200068373533\\
4	0.616023351092214\\
5	0.573484658303963\\
};
\addplot [color=mycolor3,line width=0.8pt,mark=triangle,mark options={solid,rotate=90},forget plot]
  table[row sep=crcr]{%
0	0.278462140912457\\
1	-0.0738318439004435\\
2	-0.0583408872628142\\
3	-0.114652946025534\\
4	-0.213804184577421\\
5	-0.283006266039484\\
};
\end{axis}
\end{tikzpicture}%
}
	\caption{Robustness analysis of multi-frame SR \wrt photometric variations. The $x$-axis denote the number of photometric outliers within a set of $K = 11$ LR frames. The $y$-axis denote the average normalized quality measures on our photometric outlier datasets.}
	\label{fig:photometricOutlierDatasets}
\end{figure}
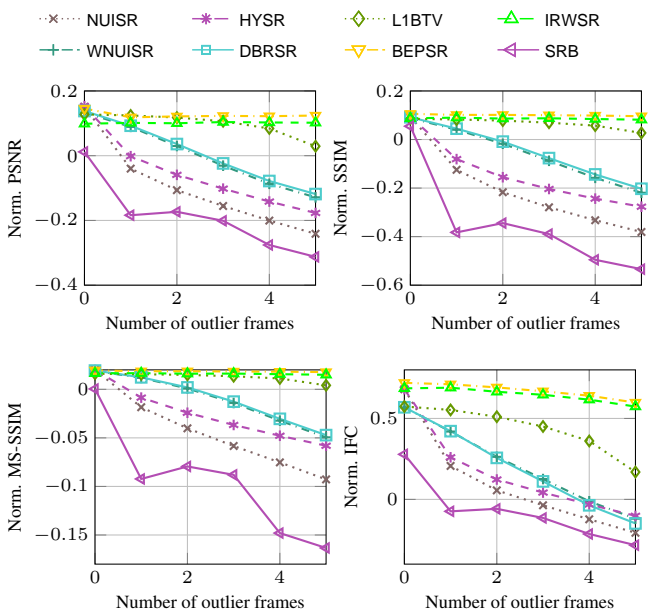

\begin{figure}[!t]
	\centering
	\mbox{
	\hspace{-1.0em}
	\subfloat[NUISR \cite{Park2003}]{
		\begin{tikzpicture}[spy using outlines={rectangle,red,magnification=4.0,height=1.5cm, width=2.7cm, connect spies, every spy on node/.append style={thick}}] 
			\node {\pgfimage[width=0.329\linewidth]{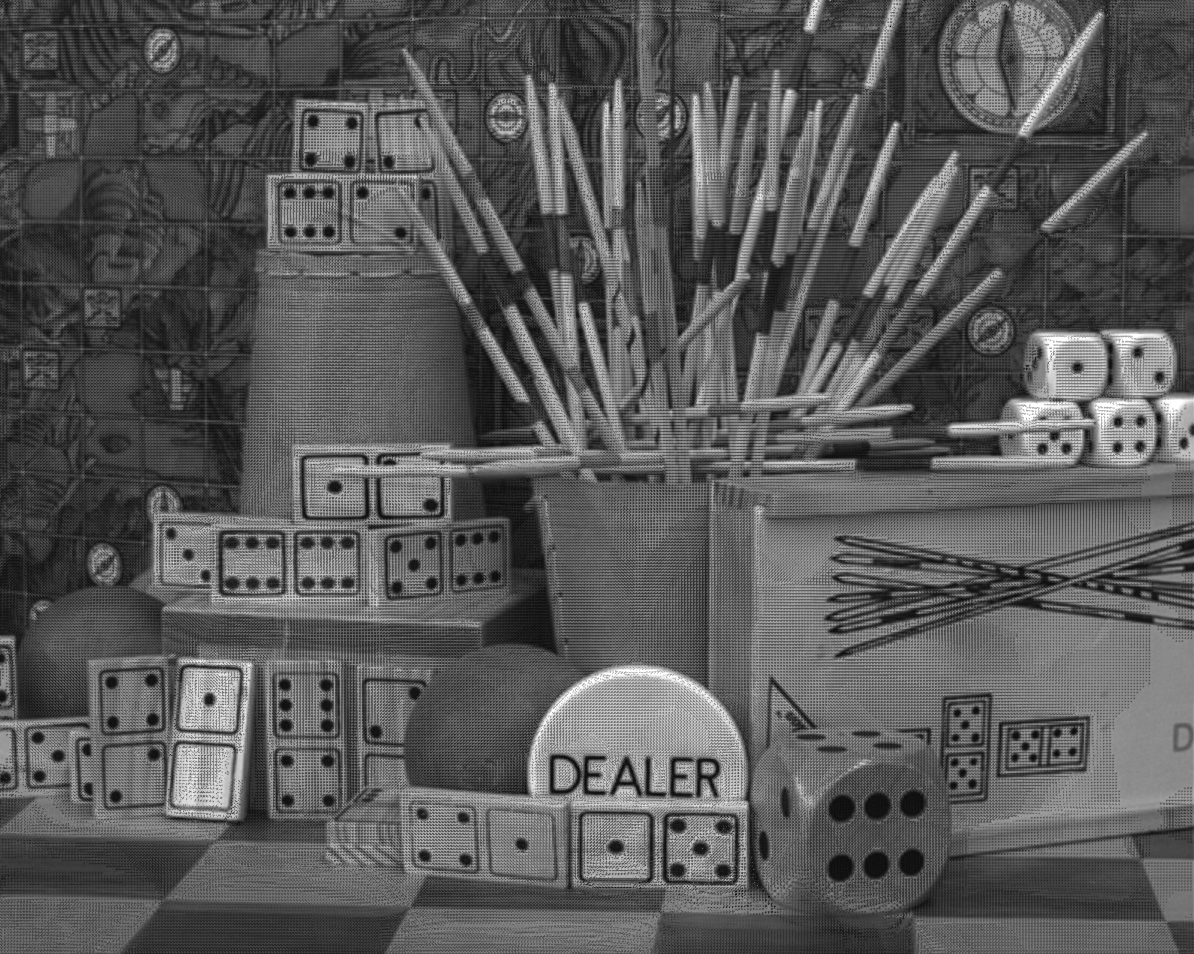}}; 
      \spy on (0.75, -0.1) in node [left] at (1.35, -1.88); 
    \end{tikzpicture}
	}\hspace{-1.2em}
	\subfloat[BEPSR \cite{Zeng2013}]{
		\begin{tikzpicture}[spy using outlines={rectangle,red,magnification=4.0,height=1.5cm, width=2.7cm, connect spies, every spy on node/.append style={thick}}] 
			\node {\pgfimage[width=0.329\linewidth]{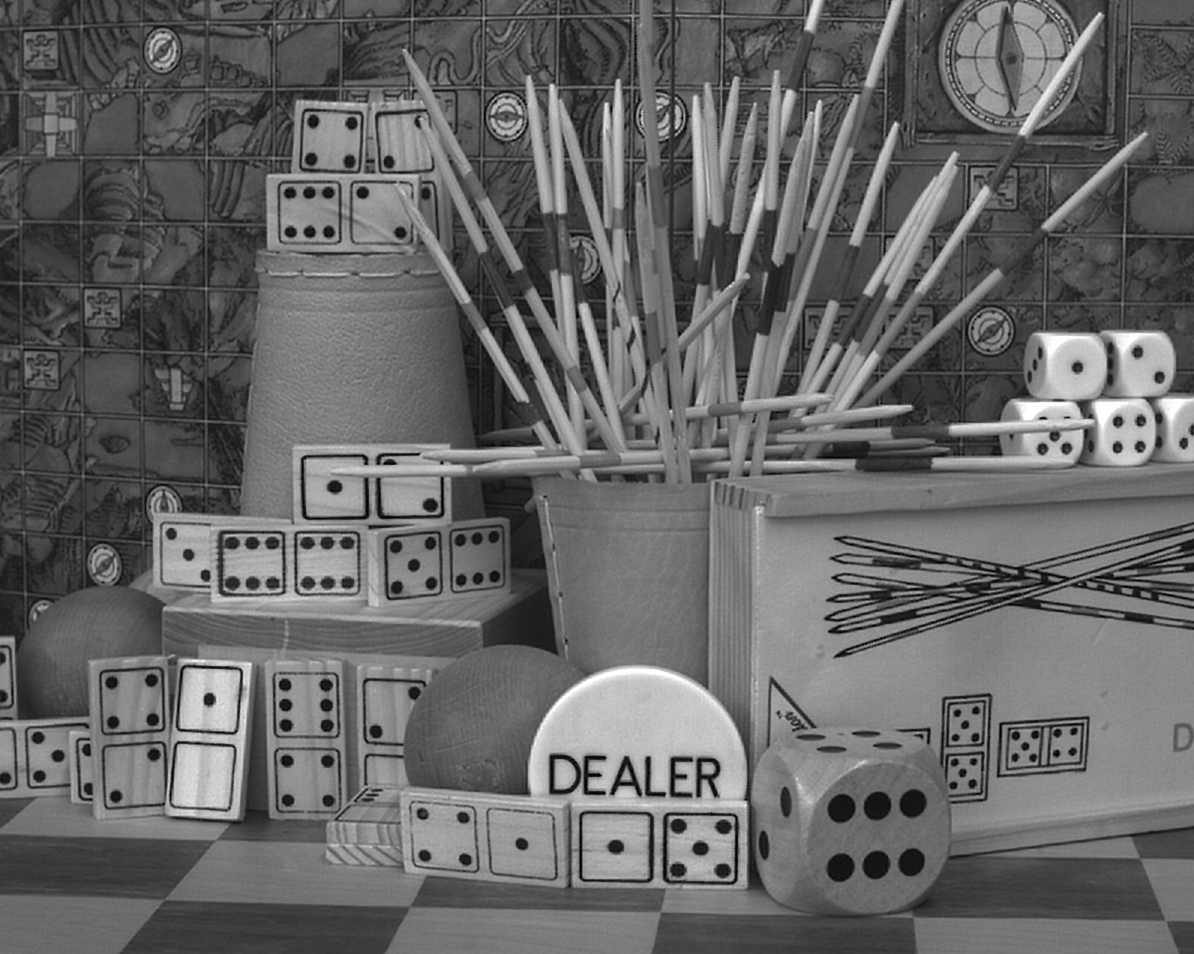}}; 
      \spy on (0.75, -0.1) in node [left] at (1.35, -1.88); 
    \end{tikzpicture}
	}\hspace{-1.2em}
	\subfloat[Ground truth]{
		\begin{tikzpicture}[spy using outlines={rectangle,red,magnification=4.0,height=1.5cm, width=2.7cm, connect spies, every spy on node/.append style={thick}}] 
			\node {\pgfimage[width=0.329\linewidth]{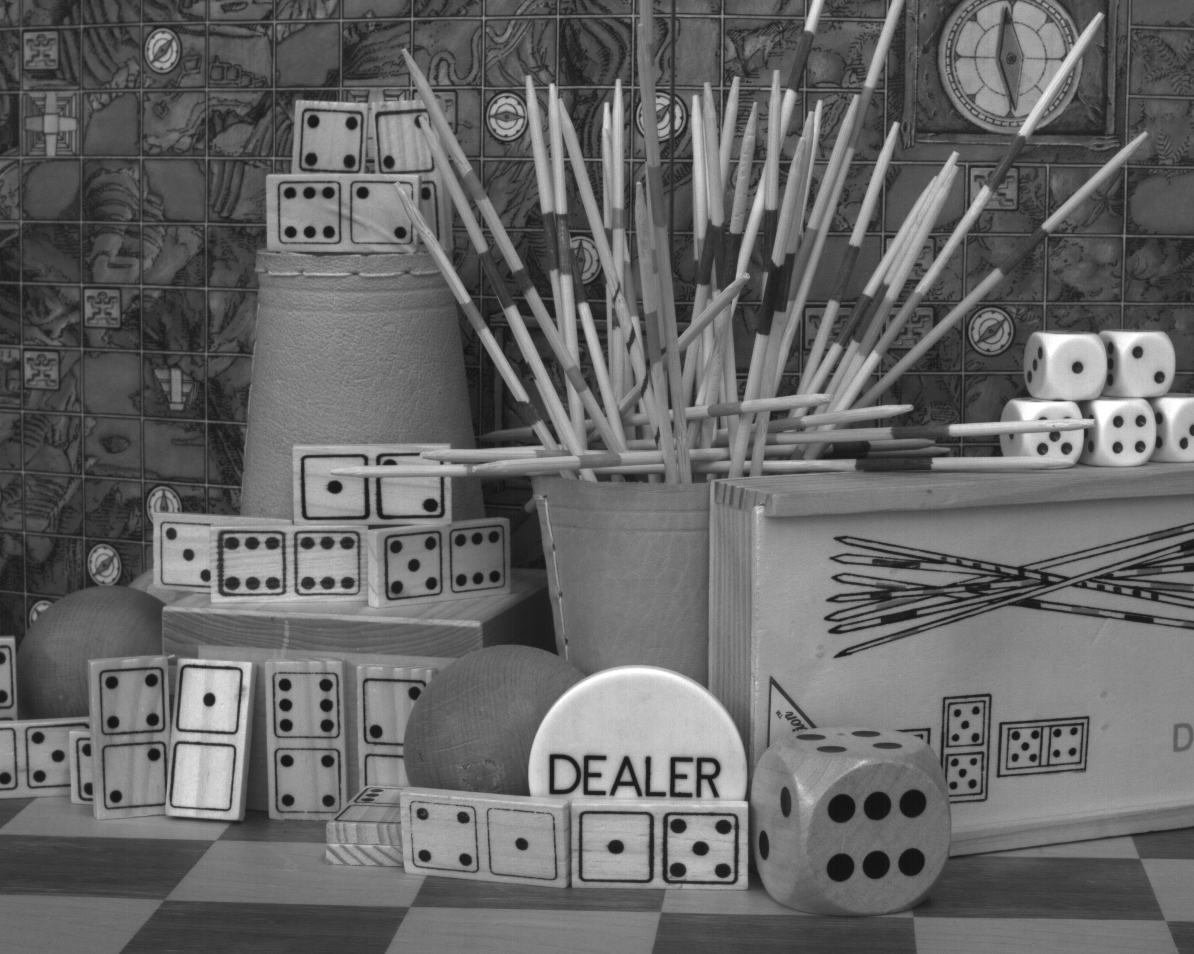}}; 
      \spy on (0.75, -0.1) in node [left] at (1.35, -1.88); 
    \end{tikzpicture}
	}}
	\caption{Multi-frame SR in the presence of photometric variations on the \textit{games} dataset ($3\times$ magnification).}
	\label{fig:photometricVariationsExample}
\end{figure}

\Fref{fig:photometricOutlierDatasets} compares various MFSR algorithms for an increasing number of photometric outlier frames within a sequence of $K = 11$ consecutive frames. We found that even for a single outlier most methods performed worse than LR data as photometric variations are neither considered implicitly by generative models nor explicitly by proper correction methods. Reconstruction-based algorithms with robust and adaptive models (IRWSR, BEPSR) were less sensitive and adaptively handled photometric variations compared to interpolation-based SR. \Fref{fig:photometricVariationsExample} depicts this behavior on the dataset shown in \fref{fig:hardwareSetup:photo}. The photometric variations resulted in intensity distortions and noise in interpolation-based SR (NUISR) while adaptive reconstruction-based SR (BEPSR) was unaffected.

\section{Conclusion}
\label{sec:Conclusion}

This paper presented the SupER database -- a new image database to benchmark SR algorithms. Unlike related studies, our database comprises real LR acquisitions and ground truth data to facilitate quantitative evaluations. We conducted comprehensive experiments of 15 SISR and MFSR algorithms to gain insights of their behavior in real applications. The main conclusions observed from our benchmark are as follows.
\\[0.8ex]
\textbf{Influence of the magnification factor.} In general, SR becomes more difficult for larger magnification factors. However, we found that the relative improvement over the LR images increases with the target magnification. Thus, SR becomes more effective for larger magnification factors.
\\[0.8ex]
\textbf{Single-image super-resolution.} For small magnification factors, most advanced SISR techniques are only slightly better or even inferior to simple bicubic interpolation. Unlike MFSR algorithms, SISR is unaffected by challenging motion or environmental conditions. Among the SISR algorithms, we observed that external methods \cite{Dong2014,Salvador2015} outperform self-exemplar methods \cite{Huang2015a} on most of our datasets.
\\[0.8ex]
\textbf{Multi-frame super-resolution.} For global camera motion, MFSR tend to outperform SISR while in case of mixed motion, MFSR is affected by inaccurate motion estimation. We found that in both situations robust reconstruction algorithms \cite{Farsiu2004a,Kohler2015c,Zeng2013} are more reliable than interpolation-based algorithms \cite{Batz2016,Batz2016b,Park2003}, particularly for longer input sequences and large magnifications. For pure local motion, interpolation-based and reconstruction-based methods are inherently limited. We observed that deep learning approaches \cite{Kappeler2016} perform well even without camera motion. Furthermore, we found that except robust reconstruction methods, all MFSR algorithms are sensitive to photometric variations. 

We provide our database, evaluation protocols, and all results on our webpage. We encourage other authors to evaluate their algorithms on our database to broaden our benchmark. In our future work, we will provide datasets that consider additional use-cases of practical relevance, \eg color images, compressed images, or motion blurred acquisitions. We also aim at analyzing image quality according to human visual perception in human subject studies to complement our objective benchmark.
	
{\small
\bibliographystyle{ieee}
\bibliography{egbib}
}

\end{document}